\documentclass[letterpaper]{article} 
\usepackage{aaai2027}
\usepackage[hyphens]{url}  
\usepackage{graphicx} 
\usepackage{natbib}  
\usepackage{caption} 
\usepackage{algorithm}
\usepackage{algorithmic}

\usepackage{amsmath}
\usepackage{amssymb}
\usepackage{cleveref}
\usepackage{newfloat}
\usepackage{listings}
\DeclareCaptionStyle{ruled}{labelfont=normalfont,labelsep=colon,strut=off} 
\floatstyle{ruled}
\newfloat{listing}{tb}{lst}{}
\floatname{listing}{Listing}

\usepackage{booktabs}

\title{Brain4FMs: A Benchmark of Foundation Models on Electrical Brain Signals}
\author{
    Fanqi Shen\textsuperscript{\rm 1},
    Junru Chen\textsuperscript{\rm 1},
    Enhong Yang\textsuperscript{\rm 1},
    Jiahe Li\textsuperscript{\rm 1},
    Xiaoran Pan\textsuperscript{\rm 1},
    Zhizhang Yuan\textsuperscript{\rm 1},\\
    Li Meng\textsuperscript{\rm 2},
    Yang Yang\textsuperscript{\rm 1}\corresponding
}

\affiliations{
    \textsuperscript{\rm 1} Zhejiang University, Hangzhou, Zhejiang, China\\
    \textsuperscript{\rm 2} Shanghai Institute of Microsystem and Information Technology,
    Chinese Academy of Sciences, Shanghai, China\\
}

\begin{document}

\maketitle

\begin{abstract}
Brain foundation models (BFMs) are advancing neurotechnology by learning transferable representations from neural signals, with broad potential in clinical diagnosis and neuroscience research. Their development relies on large-scale pretraining corpora of electrical brain signals, including scalp electroencephalography (EEG) and intracranial EEG (iEEG). However, existing BFM benchmarks primarily focus on EEG, cover only a limited subset of models, and provide limited analysis beyond downstream performance. We introduce Brain4FMs, the first unified benchmark, to our knowledge, for jointly evaluating BFMs on EEG and iEEG. It integrates 17 representative models and 21 public datasets across clinical diagnosis, sleep staging, communication, and affective computing. Brain4FMs is open and plug-and-play, with dataset-aware preprocessing, cross-subject evaluation, heterogeneous multichannel handling, and standardized downstream adaptation workflows. The benchmark reveals performance variation across tasks, signal modalities, and adaptation protocols, with no single BFM consistently dominating all evaluation scenarios.
To better understand these behaviors, we further conduct exploratory analyses of model-specific properties of spatial, spectral, and discrete representations.
\end{abstract}

\begin{links}
\link{Code}{https://github.com/wajtsq/Brain4FMs}
\end{links}
\section{Introduction}
\label{sec:introduction}
Neuroscience is fundamental to understanding the brain and translating insights into societal impact.
Electrical brain signals, including scalp electroencephalography (EEG) and intracranial EEG (iEEG), provide millisecond-level measurements of neural activity. EEG supports non-invasive, scalable recording, whereas iEEG offers higher-fidelity and more spatially precise measurements of cortical or deep-brain activity but is restricted to clinically indicated patients due to its invasive nature~\cite{parvizi2018human}. Together, these complementary modalities support a wide range of applications, including neurological disease detection, sleep staging, neural communication, and affective computing.

Deep learning has improved neural signal decoding~\cite{li2021end,jing2023development}, but its generalization is often limited by costly annotations, inter-subject variability, and distribution shifts across devices or tasks. 
To overcome these limitations, self-supervised learning (SSL) leverages unlabeled neural data for representation learning~\cite{henaff2020data}.
This shift parallels advances in Natural Language Processing and Computer Vision, where large-scale pretraining enabled foundation models. A similar transition is emerging in neuroscience, with Brain Foundation Models (BFMs) serving as universal encoders that learn robust representations and support adaptation to downstream tasks.

Despite rapid progress, existing BFM benchmarks remain largely restricted to EEG and evaluate only a limited subset of available BFMs, leaving their performance on iEEG recordings unclear~\cite{xiong2025eeg,wu2025adabrain,lu2026omnieeg}. \citet{duan2026omni} advance standardized iEEG evaluation through epilepsy, but focus on task-specific baselines rather than existing BFMs. Figure~\ref{fig:background}(a) summarizes the complementary benchmark scopes. Thus, these limitations motivate a unified cross-subject benchmark for comparing BFMs across EEG and iEEG.
To bridge these gaps, we introduce Brain4FMs, a standardized benchmark for evaluating BFMs on EEG/iEEG classification. The evaluation centers around three questions: 1) How do current BFMs perform across tasks and modalities? 2) How does adaptation change their transferability? 3) What structural patterns can be observed in their representations? Our main contributions are:
\begin{itemize}
\item[1)] \textbf{A Unified EEG/iEEG Benchmark.} We introduce Brain4FMs, an open benchmark that integrates 17 representative BFMs and 21 public EEG/iEEG datasets covering 26 classification settings in clinical diagnosis, sleep staging, communication, and affective computing.
\item[2)] \textbf{Standardized Evaluation Protocol.} We establish an extensible cross-subject evaluation framework that standardizes downstream procedures while accommodating model- and modality-specific requirements.
\item[3)] \textbf{Extensive Empirical Evaluation.} We report substantial variation in BFM performance and modality-associated behavior under standardized evaluation, complemented by exploratory analyses of model representations.
\end{itemize}

\section{Benchmark Design}
To support standardized and reproducible evaluation, we introduce Brain4FMs, a comprehensive benchmark for EEG/iEEG classification with BFMs. It unifies data processing, model integration, and cross-subject evaluation, with plug-and-play interfaces for adding new models and datasets.

\subsection{Dataset coverage}
We select well-established datasets that support cross-subject evaluation and are widely used in brain-signal research. Dataset selection balances task diversity, recording modality, and practical accessibility, while avoiding overlap with the reported pretraining corpora of evaluated BFMs whenever possible. Brain4FMs covers 21 public datasets and 26 classification settings, with five datasets contributing two settings each. Table~\ref{tab:benchmark_d} summarizes the benchmark datasets, with complete statistics provided in the supplement.

\paragraph{Disease Diagnosis.}
Neurological disorders are among the leading causes of disability and mortality worldwide ~\cite{li2025deep}.
To support clinically relevant evaluation, Brain4FMs covers epilepsy, drug-resistant epilepsy (DRE), depression, Major Depressive Disorder (MDD), Schizophrenia (SZ), Alzheimer’s Disease (AD), Parkinson’s disease (PD) and Attention Deficit Hyperactivity Disorder (ADHD). These tasks use clinically grounded labels for realistic evaluation.

\paragraph{Sleep Staging.}
Sleep staging supports sleep disorder diagnosis, but manual overnight scoring is labor-intensive, motivating automated models that capture long-range temporal dependencies and rich time-frequency patterns.

\paragraph{Communication.}
Communication tasks map neural activity to external outputs for intent expression. Typical tasks include Motor Imagery (MI), Motor Execution (ME), and multimodal neural decoding, which aims to infer visual, auditory, or semantic information from brain signals.

\begin{table}[!hb]
\setlength{\tabcolsep}{1.8pt}
\centering
{\small
\begin{tabular}{lccc}
\toprule
Name & Signal & Task & Cat. \\ \midrule
CHBMIT~\cite{guttag2010chb} & EEG & Epilepsy  & 2 \\
MAYO~\cite{nejedly2020multicenter} & iEEG & DRE & 2 \\
FNUSA~\cite{nejedly2020multicenter} & iEEG & DRE & 2 \\
HUP~\cite{bernabei2023quantitative} & iEEG  & DRE & 2/2 \\
SWEC~\cite{carzaniga2025foundation} & iEEG  & DRE  & 2 \\
Dep~\cite{ds003478:1.1.0} & EEG & Depression & 2/3 \\
MDD-64~\cite{Mumtaz2016} & EEG & MDD & 2 \\
SZ-28~\cite{repod.0107441_2017} & EEG & SZ & 2 \\
UCSD~\cite{ds002778:1.0.5} & EEG & PD & 2/2 \\
ADFD~\cite{ds004504:1.0.2} & EEG & AD & 2 \\
ADHD-Adult~\cite{bajestani2023dataset} & EEG & ADHD & 2 \\
ADHD-Child~\cite{rzfh-zn36-20} & EEG & ADHD & 2 \\ \midrule
ISRUC-G1~\cite{khalighi2016isruc} & EEG & Sleep Stage & 5 \\
SleepEDF~\cite{kemp2000analysis} & EEG & Sleep Stage & 5 \\ \midrule
EEGMMIDB~\cite{goldberger2000physiobank} & EEG & ME & 4/4 \\
BCI-2a~\cite{brunner2008bci} & EEG & MI & 4 \\
Chisco~\cite{zhang2024chisco} & EEG & Concept & 39/39 \\
Cogitate-CF~\cite{seedat2025open} & iEEG & Visual & 2 \\ \midrule
DEAP~\cite{koelstra2011deap} & EEG & Emotion & 4 \\
SEED-IV~\cite{zheng2018emotionmeter} & EEG & Emotion & 4 \\
EEGMat~\cite{zyma2019electroencephalograms} & EEG & MW & 2 \\ 
  \bottomrule
\end{tabular}
}
\caption{Benchmark datasets by signal type, task, and number of classes (Cat.); "/" separates two evaluation settings.}
\label{tab:benchmark_d}
\end{table}


\begin{figure*}[!ht]
  \centering
  \includegraphics[width = 0.95\textwidth]{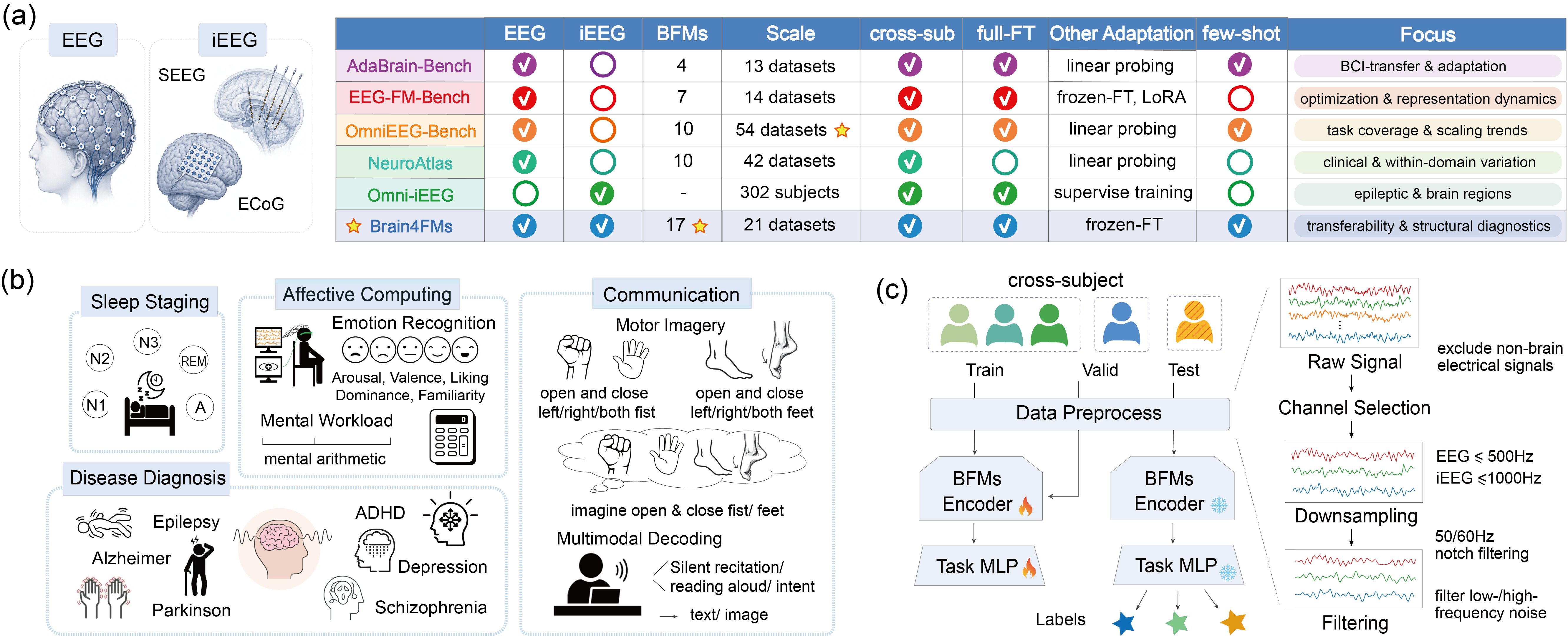}
  \caption{Overview of Brain4FMs. 
  (a) Scope of related work by signal modality, benchmark scale, evaluation protocol, adaptation setting, and analysis focus. (b) EEG/iEEG acquisition across diverse clinical and human-centered scenarios. (c) Dataset-aware preprocessing and subject-disjoint cross-validation under a standardized evaluation framework. }
  \label{fig:background}
\end{figure*}

\paragraph{Affective Computing.}
Affective computing infers internal emotional and cognitive states from EEG/iEEG. Core tasks include emotion recognition and mental workload estimation (MW), which involve temporal, spectral, and spatial patterns of affective and cognitive demand across diverse settings.

\subsection{Model Interfaces}

As summarized in Table~\ref{tab:benchmark_m}, Brain4FMs includes 17 representative BFMs spanning diverse architectures and pretraining objectives. We include BFMs from peer-reviewed studies or widely cited preprints with publicly available code and pretrained checkpoints. All models use the same dataset-specific preprocessing pipeline and standardized downstream evaluation procedures while retaining each model’s original architecture, thereby assessing system-level transfer under controlled adaptation. Given an EEG or iEEG sample $X\in\mathbb{R}^{C\times N}$, where $C$ and $N$ denote the numbers of channels and temporal samples, respectively, each encoder produces a latent representation for downstream prediction. Plug-and-play interfaces further support new models. We group the included BFMs by their primary pretraining mechanisms.

\begin{table}[!hb]
\setlength{\tabcolsep}{1pt}
\centering
{\footnotesize
\begin{tabular}{lllll}
\toprule
Name  & Strategy  & Param & Modality & Domain \\
\midrule
\multicolumn{4}{l}{\textbf{Contrastive-based Method}} \\
\midrule
SppEEGNet & Aug.  & 138 K  & EEG   & time   \\
BIOT & Aug.    & 3.19 M   & EEG   & time, frequency   \\
Bendr & CPC   & 3.97 M  & EEG  & time, frequency   \\
MBrain  & CPC  & 8.34 M & EEG/iEEG & time, frequency, space   \\
\midrule
\multicolumn{4}{l}{\textbf{Generative-based Method}} \\
\midrule
Brant   & AE  & 196.10 M  & iEEG & time, frequency, space    \\
BFM  & AE   & 708.96 M  & EEG   & time     \\
BrainBERT & AE & 43.18 M  & iEEG   & time, frequency     \\
CBraMod  & AE  & 4.88 M  & EEG   & time, frequency, space  \\
NeuroGPT & AR  & 79.62 M   & EEG  & time, frequency        \\
LaBraM  & AE  & 5.80 M  & EEG  & time, frequency, space   \\
BrainWave & AE & 102.13 M & EEG+iEEG & time, frequency, space    \\
REVE   & AE   & 69.19 M  & EEG   & time, space   \\
BrainOmni & AE & 32.71 M   & EEG   & time, frequency, space   \\
CodeBrain & AE & 15.17 M   & EEG   & time, frequency, space   \\
\midrule
\multicolumn{4}{l}{\textbf{Other Advanced Methods}} \\
\midrule
EEGPT   & G \& C  & 51.04 M   & EEG   & time, space    \\
NeuroLM  & G \& Adv.  & 169.60 M   & EEG   & time, frequency  \\
MVPFormer  & G \& C  & 73.59 M   & iEEG   & time, frequency, space    \\
\bottomrule
\end{tabular}
}
\caption{Overview of models by SSL strategy, encoder size, pretraining modality, and feature domain. C/G/O/Adv. indicate contrastive/generative/other/adversarial methods.}
\label{tab:benchmark_m}
\end{table}

\paragraph{Contrastive-based Methods.}
Contrastive methods learn representations by aligning related signal views or temporal segments while separating mismatched examples. SppEEGNet~\cite{li2022spp} and BIOT~\cite{yang2023biot} construct positive pairs through signal augmentations, whereas Bendr~\cite{kostas2021bendr} and MBrain~\cite{cai2023mbrain} adopt predictive or spatiotemporal contrastive objectives.

\paragraph{Generative-based Methods.}
These methods learn representations by reconstructing masked signals, predicting latent targets, or modeling neural signals as token sequences. BrainBERT, Brant, BrainWave, CBraMod, REVE, and CodeBrain primarily employ masked or structured reconstruction~\cite{wang2023brainbert, zhang2023brant, yuan2024brainwave, wang2024cbramod, ouahidi2025reve, ma2025codebrain}.
NeuroGPT~\cite{cui2024neurogpt} adopts autoregressive prediction, whereas LaBraM, BFM, and BrainOmni incorporate discrete tokenization or codebook-based modeling~\cite{jiang2024large, bayazi2024general, xiao2025brainomni}.

\paragraph{Other Advanced Methods.}
Beyond contrastive and generative SSL, BFMs also explore predictive, hybrid, adversarial, and instruction-tuning strategies. EEGPT and MVPFormer combine generative and contrastive objectives, whereas NeuroLM combines generative and adversarial learning~\cite{wang2024eegpt,carzaniga2025foundation,jiang2024neurolm}.

\subsection{Evaluation Protocol}
Brain4FMs focuses on classification tasks, which aim to divide the input EEG/iEEG samples into pre-defined categories. The pipeline consists of two components: dataset preprocessing and downstream evaluation (Figure~\ref{fig:background}(c)).

\paragraph{Dataset preprocessing.}
We standardize data organization and preprocessing principles rather than enforcing identical signal-processing parameters across datasets. The pipeline applies dataset-appropriate bandpass and notch filtering, downsampling, event-aligned window segmentation, channel selection, and per-channel z-score normalization. We use group-wise cross-validation with approximately $3{:}1{:}1$ cross-subject training, validation, and test splits, ensuring that recordings from the same subject never appear across splits. For iEEG with patient-specific channel counts, compatible models use patient-wise batching to preserve native channel structures without cross-subject alignment.

\paragraph{Downstream Evaluation.}
Each pretrained BFM serves as the backbone with its model-compatible classification interface. Full-parameter fine-tuning is used as the primary setting, while frozen-encoder and prototype-based low-label evaluations are included as complementary analyses. For each model-dataset pair, we search the encoder learning rate (LR) over $\{10^{-4}, 5\times10^{-5}, 10^{-5}, 5\times10^{-6}, 10^{-6}\}$, with the classification-head LR set to ten times the encoder rate. Within each fold, checkpoints are selected by validation loss, and the learning rate is selected using the primary validation metric: AUROC for binary tasks and Cohen's $\kappa$ for multiclass tasks. 
Final performance is averaged across test folds, with selected LR, complete metrics, and training details such as the optimizer, batch size, and schedule reported in the supplement.

\section{Empirical Results}
\label{sec:empirical}
\textbf{Q1: How do current BFMs perform across tasks and modalities?} To answer this question, we report the main results under full-FT, followed by analyses of task performance, EEG and iEEG modalities, and pretraining paradigms.

\subsection{Task Performance}

\begin{figure*}[!ht]
  \centering
  \includegraphics[width = 0.99\textwidth]{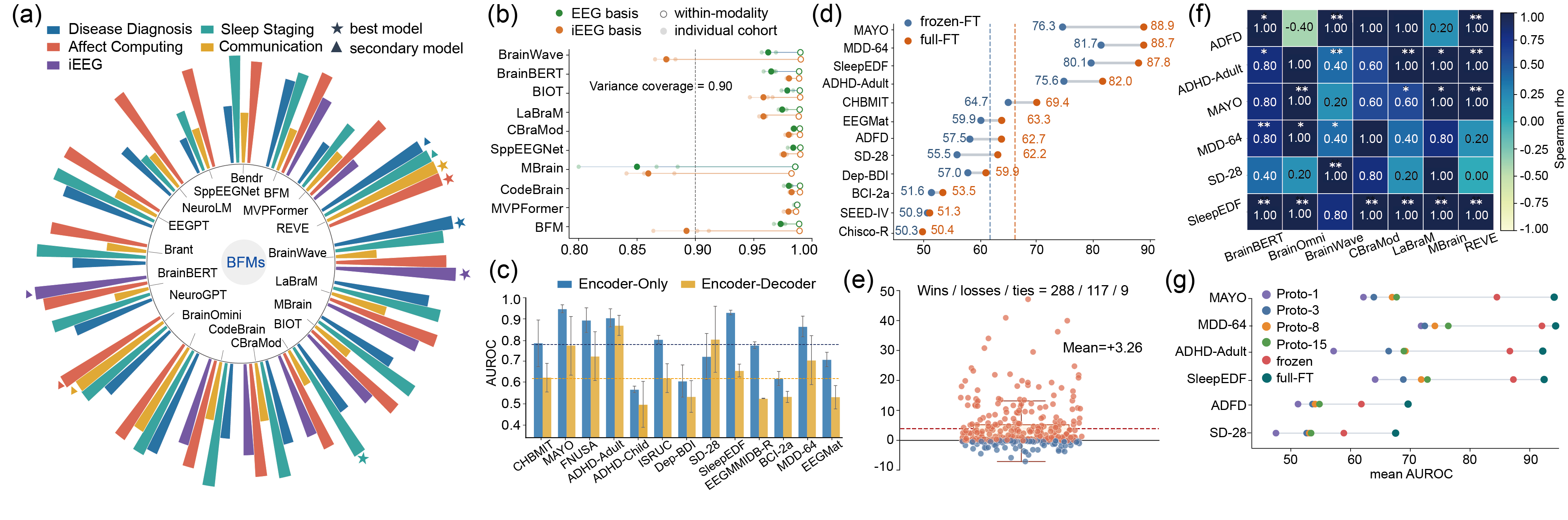}
\caption{
Empirical results and downstream adaptation analyses. 
(a) Overall BFM performance, averaged by task using AUROC.
(b) Bidirectional EEG/iEEG subspace coverage. 
(c) AUROC comparison of encoder-only and encoder-decoder NeuroGPT variants.
(d-e) AUROC gains of full-FT over frozen-FT across BFMs. 
(f) Pair-level trends across Proto-1/3/8/15, * q<0.05, ** q<0.01 by the BH-corrected Page test.
(g) Mean AUROC under prototype transfer, frozen-FT, and full-FT.
}
  \label{fig:total}
\end{figure*}

We compare BFMs with supervised baselines selected for each task under fully matched experimental settings. The reference models include SPaRCNet~\cite{jing2023development} for epilepsy and ADHD, DeprNet for depression~\cite{seal2021deprnet}, LocCNN~\cite{Choi2026DetectionEarlyStageParkinsons} for AD and PD, CCNSE for sleep staging~\cite{li2021end}, and MSCARNet for communication and affective computing tasks~\cite{avola2024multi}. As shown in Tables~\ref{tab:tab1_epilepsy} and ~\ref{tab:tab3_other}, BFMs exhibit stronger cross-subject performance on epilepsy and sleep staging. Supervised baselines remain competitive but typically outperform only a subset of BFMs. On depression datasets, DeprNet achieves moderate AUROC but low accuracy, indicating limited robustness. On PD, LocCNN outperforms most BFMs, although cross-subject performance remains unstable. Overall, BFMs surpass the selected supervised references on most clinical tasks, showing promise for cross-subject transfer.

Performance within affective computing and communication is highly task dependent. Cross-subject concept decoding, visual classification and emotion recognition remain particularly challenging because of pronounced non-stationarity and inter-subject variability~\cite{lotte2018review}. On these tasks, the BCI-tailored supervised baseline MSCARNet outperforms most BFMs, although models such as NeuroGPT remain competitive on selected datasets. In contrast, MI, ME, and MW yield stronger results for several BFMs: REVE and NeuroGPT achieve competitive accuracy and AUROC, while BrainOmni performs strongly on EEGMMIDB. These observations suggest that current BFMs transfer more consistently to motor-related classification than to higher-level semantic or affective decoding under the cross-subject setting. NeuroLM, despite pretraining on emotion-related data, does not consistently generalize across subjects.

\subsection{EEG and iEEG modalities}
The benchmark reveals a modality-matched transfer trend. On the epilepsy datasets, iEEG-pretrained models generally perform better on intracranial datasets, whereas EEG-pretrained models are more competitive on scalp EEG. For example, BrainBERT performs strongly on FNUSA and HUP-ECoG but its relative advantage is smaller on CHBMIT. LaBraM performs well on EEG and single-channel iEEG datasets but falls behind supervised baselines on the patient-specific HUP datasets. BrainWave, jointly pretrained on EEG and iEEG, remains competitive across both modalities, consistent with its controlled finding that joint EEG-iEEG pretraining outperforms EEG-only and iEEG-only variants.


Motivated by this trend, we analyzed cross-modal subspace overlap and mixed-space expansion in ten pretrained BFMs using matched 2-s epochs from six EEG and six multichannel iEEG cohorts. From each cohort, we sampled 2,000 label-balanced epochs and centered representations within the cohort before pooling, which yielded 12,000 samples per modality. To assess robustness to EEG cohort composition, we repeated the analysis with three random EEG-cohort selections. For each modality, we extracted the PCA subspace explaining 99\% of variance. Directional coverage $C_{S\rightarrow T}$ is the fraction of target variance captured by the source subspace. We estimated coverage on complementary sample halves and computed within-modality coverage in the same way. Modality-specific residual structure is defined as $G=C_{\mathrm{within}}-C_{\mathrm{cross}}$, with larger values indicating more modality-specific variance.

\begin{table}[!hb]
\setlength{\tabcolsep}{1.8pt}
\centering
{\footnotesize
\begin{tabular}{lcccccc}
\toprule
\multicolumn{1}{l}{}     & \multicolumn{2}{c}{HUP-ECoG}                           & \multicolumn{2}{c}{FNUSA}                     & \multicolumn{2}{c}{CHBMIT}                         \\ \midrule
Model       & \multicolumn{1}{c}{AUROC} & \multicolumn{1}{c}{Acc} & \multicolumn{1}{c}{AUROC} & \multicolumn{1}{c}{Acc}  & \multicolumn{1}{c}{AUROC} & \multicolumn{1}{c}{Acc} \\ \midrule
MBrain     & .57$\pm$.09                   & .61$\pm$.12                & .91$\pm$.08                   & .87$\pm$.08           & .75$\pm$.08                   & .71$\pm$.12                 \\
BIOT        & .54$\pm$.08                   & .61$\pm$.12                & .87$\pm$.07                   & .83$\pm$.08           & .56$\pm$.08                   & .29$\pm$.05                 \\ 
SppEEGNet   & .44$\pm$.07                   & .52$\pm$.09                & .82$\pm$.09                   & .84$\pm$.09           & .43$\pm$.05                   & .60$\pm$.06                 \\ 
  \midrule
BrainWave  & \textbf{.88$\pm$.07}                   & \textbf{.80$\pm$.11}               & \textbf{.92$\pm$.05}                   & .89$\pm$.06           & \textbf{.90$\pm$.03}                   & \textbf{.80$\pm$.12}                 \\
BrainBERT   & .81$\pm$.12                   & .73$\pm$.09             & .90$\pm$.04                   & \textbf{.89$\pm$.04}           & .78$\pm$.06                   & .75$\pm$.10                 \\
BFM         & .72$\pm$.11                   & .71$\pm$.10                & .78$\pm$.12                   & .69$\pm$.11           & .74$\pm$.05                   & .72$\pm$.03    \\
CodeBrain     & .67$\pm$.05                   & .71$\pm$.06                & .83$\pm$.10                   & .85$\pm$.06           & .60$\pm$.08                   & .74$\pm$.10     \\ 
CBraMod      & .64$\pm$.09                   & .58$\pm$.10                & .71$\pm$.13          & .78$\pm$.08           & .69$\pm$.04         & .70$\pm$.08         \\
LaBraM      & .51$\pm$.06                   & .64$\pm$.13                & .89$\pm$.08                   & .82$\pm$.10           & .75$\pm$.10                   & .73$\pm$.10                 \\ \midrule
SPaRCNet  & .56$\pm$.01                   & .59$\pm$.09                & .85$\pm$.10                   & .82$\pm$.11           & .61$\pm$.10                   & .66$\pm$.05                 \\ \bottomrule
\end{tabular}
}
\caption{Primary epilepsy metrics for representative models.}
\label{tab:tab1_epilepsy}
\end{table}

\begin{table*}[t]
\setlength{\tabcolsep}{1.6pt}
\centering
{\footnotesize
\begin{tabular}{lcccccccccccccc}
\toprule
\multicolumn{1}{l}{}          & \multicolumn{2}{c}{ADHD-Adult}                       & \multicolumn{2}{c}{ADHD-Child}                       & \multicolumn{2}{c}{ADFD}                            & \multicolumn{2}{c}{MDD-64}             & \multicolumn{2}{c}{SZ-28}            & \multicolumn{2}{c}{SleepEDF}       & \multicolumn{2}{c}{UCSD-ON}         \\ \midrule
Model       & \multicolumn{1}{c}{AUROC} & \multicolumn{1}{c}{Acc} & \multicolumn{1}{c}{AUROC} & \multicolumn{1}{c}{Acc} & \multicolumn{1}{c}{AUROC} & \multicolumn{1}{c}{Acc} & \multicolumn{1}{c}{AUROC} & \multicolumn{1}{c}{Acc} & \multicolumn{1}{c}{AUROC} & \multicolumn{1}{c}{Acc} & \multicolumn{1}{c}{AUROC} & \multicolumn{1}{c}{Acc}  & \multicolumn{1}{c}{AUROC} & \multicolumn{1}{c}{Acc} \\ \midrule
MBrain      & .95$\pm$.03                   & .92$\pm$.04                  & .69$\pm$.11                   & .63$\pm$.08                  & .53$\pm$.12                   & .50$\pm$.10         & .93$\pm$.08                   & .87$\pm$.11     & .58$\pm$.10                   & .58$\pm$.10          & \textbf{.95$\pm$.01}                   & \textbf{.78$\pm$.04}    & .49$\pm$.15                   & .46$\pm$.14           \\
Bendr       & .62$\pm$.03                   & .61$\pm$.04                  & .53$\pm$.02                   & .51$\pm$.01                  & .52$\pm$.02                   & .52$\pm$.01         & .85$\pm$.07                   & .78$\pm$.05     & .49$\pm$.05                   & .50$\pm$.03          & .91$\pm$.01                   & .69$\pm$.06    & .51$\pm$.05                   & .49$\pm$.05           \\ \midrule
BrainOmni   & .96$\pm$.02                   & .91$\pm$.02                  & .71$\pm$.09                   & .62$\pm$.07                  & .80$\pm$.07                   & .71$\pm$.04         & .94$\pm$.07                   & .85$\pm$.09     & .69$\pm$.15                   & .69$\pm$.12          & .87$\pm$.00                   & .63$\pm$.03    & .53$\pm$.14                   & .49$\pm$.12           \\
BrainWave   & .96$\pm$.03                   & .91$\pm$.04                  & .78$\pm$.04                   & .71$\pm$.04                  & .74$\pm$.05                   & .68$\pm$.05         & .94$\pm$.03                   & .85$\pm$.02     & \textbf{.88$\pm$.07}                   & \textbf{.81$\pm$.06}          & .89$\pm$.02                   & .67$\pm$.05    & \textbf{.69$\pm$.10}                   & .53$\pm$.08           \\
REVE        & .96$\pm$.02                   & .91$\pm$.02                  & \textbf{.79$\pm$.07}                   & \textbf{.71$\pm$.04}                  & \textbf{.84$\pm$.07}                   & \textbf{.77$\pm$.08}         & \textbf{.97$\pm$.04}                   & \textbf{.88$\pm$.09}     & .81$\pm$.13                   & .75$\pm$.13          & .94$\pm$.02                   & .75$\pm$.04    & .63$\pm$.19                   & \textbf{.56$\pm$.13}           \\
LaBraM      & .95$\pm$.04                   & .91$\pm$.04                  & .66$\pm$.12                   & .64$\pm$.08                  & .83$\pm$.10                   & .75$\pm$.07         & .96$\pm$.04                   & .87$\pm$.08     & .54$\pm$.07                   & .54$\pm$.04          & .93$\pm$.01                   & .76$\pm$.04    & .51$\pm$.22                   & .45$\pm$.11           \\
CBraMod     & .95$\pm$.02                   & \textbf{.92$\pm$.04}                  & .64$\pm$.05                   & .64$\pm$.06                  & .55$\pm$.06                   & .59$\pm$.04         & .90$\pm$.13                   & .84$\pm$.12     & .54$\pm$.14                   & .58$\pm$.05          & .93$\pm$.01                   & .73$\pm$.04    & .60$\pm$.12                   & .49$\pm$.08           \\
BrainBERT   & .74$\pm$.05                   & .69$\pm$.05                  & .55$\pm$.03                   & .56$\pm$.03                  & .59$\pm$.05                   & .58$\pm$.05         & .92$\pm$.07                   & .85$\pm$.07     & .73$\pm$.11                   & .66$\pm$.15          & .91$\pm$.02                   & .68$\pm$.07    & .51$\pm$.05                   & .51$\pm$.06           \\ \midrule
EEGPT       & \textbf{.96$\pm$.03}                   & .91$\pm$.04                  & .67$\pm$.13                   & .64$\pm$.10                  & .81$\pm$.05                   & .72$\pm$.05         & .94$\pm$.05                   & .87$\pm$.08     & .56$\pm$.21                   & .56$\pm$.11          & .91$\pm$.00                   & .68$\pm$.03    & .51$\pm$.21                   & .48$\pm$.17           \\
  \midrule
$SL^{\dagger}$  & .94$\pm$.02  & .89$\pm$.03  & .66$\pm$.05  & .62$\pm$.03  & .76$\pm$.05  & .70$\pm$.05  & .91$\pm$.06  & .78$\pm$.04  & .78$\pm$.15  & .74$\pm$.09  & .84$\pm$.02  & .33$\pm$.04  & .55$\pm$.23  & .52$\pm$.18    \\ \bottomrule
\end{tabular}
}
\caption{Primary performance metrics on clinical tasks. We show a subset of BFMs and $SL^{\dagger}$ denotes supervised baselines.}
\label{tab:tab3_other}
\end{table*}

The effective dimensionality of representations is primarily model dependent, with broadly consistent distributions across datasets within each model. Across all ten BFMs, the dominant EEG and iEEG subspaces are largely shared: mean bidirectional coverage exceeds 0.85 for every model, indicating that each modality’s principal variance is largely captured by the other modality’s subspace. This pattern is robust to EEG cohort selection, with highly consistent model rankings across the three draws.
However, the overlap can be asymmetric. For BrainWave, $C_{EEG\to iEEG} = 0.96$, compared with 0.88 in the reverse direction, indicating modality-specific residuals.
Moreover, the residual gap \(G\) is positively correlated with macro-averaged AUROC across the 12 cohorts in each draw (\(\rho=0.86\text{-}0.90\), all $p<0.01$). This association is stronger for iEEG tasks (\(\rho=0.81\text{-}0.87\)) than for EEG tasks (\(\rho=0.66\text{-}0.75\)). Together, these results suggest that BFMs learn a largely shared EEG/iEEG representational backbone while retaining modality-specific residual structure associated with higher frozen-transfer performance.

\subsection{Pretraining Paradigms}
Beyond data modality, we also observe that SSL objectives are associated with different transfer behaviors. 
Among contrastive BFMs, MBrain shows the strongest overall performance on clinically relevant tasks. Compared with other models, MBrain uniquely combines multichannel CPC with graph-based modeling. 
Within generative BFMs, most models adopt autoencoding (AE) rather than autoregressive (AR) pretraining. 
AR models predict tokens sequentially under a causal context~\cite{liu2021self}, whereas AE can aggregate information across the full input window.
The latter may be better aligned with downstream classification, which typically relies on window-level discrimination rather than sequence generation. The comparison between the encoder-only NeuroGPT-E and the encoder-decoder NeuroGPT-D further supports this observation. As shown in Figure~\ref{fig:total}(c), NeuroGPT-E consistently outperforms NeuroGPT-D across epilepsy, sleep staging, MW, and MI tasks, suggesting that downstream classification relies primarily on encoder representations, with limited additional benefit from the decoder.
\section{Downstream Adaptation}
\label{sec:adaptation}
\textbf{Q2: How does adaptation change their transferability?} To address this question, we consider two complementary settings: frozen-encoder transfer, which isolates the benefit of pretrained representations, and prototype-based transfer, which evaluates representations under limited supervision.
\subsection{Frozen-Encoder Transfer}
We compare frozen-encoder fine-tuning (frozen-FT) with full fine-tuning (full-FT). 
Frozen-FT fixes the pretrained encoder and optimizes only the classifier, directly evaluating the transferability of frozen representations. Full-FT instead updates both the encoder and classifier, allowing us to quantify the additional gain from adapting the pretrained encoder to the target task.
To ensure a controlled comparison, frozen-FT uses the same classifier-head LR selected for the corresponding full-FT setting, with detailed results reported in the supplementary material. After averaging AUROC across cross-validation folds for each model–dataset pair, we obtain paired observations (n=414). Full-FT improves over frozen-FT by 3.26 AUROC points on average, with significant gains under both the Wilcoxon signed-rank test ($p=3.49\times10^{-25}$) and the sign test ($p=5.96\times10^{-17}$). These results show that transfer performance depends not only on pretrained representations but also on their downstream adaptation.

The gain from full-FT is not uniform: most model–dataset pairs show moderate improvements, while a few benefit substantially from encoder updates.
Representative examples include Bendr on MAYO (+47.2 AUROC) and BIOT on MDD-64 (+36.4), as shown in Figure~\ref{fig:total}(d). In these cases, frozen representations provide a useful initialization, but full-FT is needed to reshape the feature space and form more effective task-specific decision boundaries. 
Full-FT is not always better than frozen-FT. 
In a few cases, frozen-FT achieves slightly higher AUROC. However, these cases generally correspond to weak or unstable transfer performance rather than strong frozen representations. This suggests that the benefit of full-FT may be limited or unstable on difficult tasks. Despite these differences, frozen-FT and full-FT show strong ranking consistency across model-dataset pairs (Spearman $\rho=0.82$, $p=2.29\times{10}^{-91}$). Thus, frozen-FT remains an important indicator of BFM transferability: it largely preserves the relative ranking among models, while full-FT further raises the achievable performance ceiling through encoder adaptation.

\subsection{Prototype-Based Few-Shot Transfer}
Although most BFMs are not explicitly trained for few-shot inference, limited labeled brain-signal data make this setting practically important. We therefore adopt a parameter-free prototype-based protocol that classifies each query by comparing its representation with class prototypes. For each class, we sample a small support set and define its prototype as the mean frozen representation:
$\mathbf{p}_c = \frac{1}{|\mathcal{S}_c|}\sum_{i\in \mathcal{S}_c} f_\theta(x_i)$,
where $\mathcal{S}_c$ denotes the sampled support set for class $c$, and $f_\theta(\cdot)$ is the frozen BFM encoder. A test sample $x$ is assigned to the class whose prototype has the highest cosine similarity averaged across channels:
$\hat{y} = \arg\max_c \ \mathrm{sim}\bigl(f_\theta(x), \mathbf{p}_c\bigr)
$.
This protocol evaluates whether the frozen representation supports class discrimination under limited supervision without updating model parameters.

Figure~\ref{fig:total}(g) reports results for seven high-performing BFMs on six datasets under 1-, 3-, 8-, and 15-shot settings, with each fold repeated five times. Prototype-based transfer remains below full-FT in absolute performance, but generally improves as the support set grows. Proto-15 outperforms Proto-1 in 95\% of the 42 model-dataset pairs, with an average gain of 6.68 AUROC points. 
Across the ordered 1/3/8/15-shot settings, most model-dataset pairs show positive trend correlations, with a mean Spearman correlation of $\rho=0.73$. We assess these trends using the Page test and control the false discovery rate with the Benjamini-Hochberg procedure, as shown in Figure~\ref{fig:total}(f). The pair-level results show that the gains are heterogeneous: only a subset of model-dataset pairs retain statistically significant positive trends after correction.

Prototype performance also moderately agrees with the model ordering under full-FT. Across Proto-1, Proto-3, Proto-8, and Proto-15, the Spearman correlations with full-FT range from $0.64$ to $0.68$. The agreement is stronger on MAYO and SleepEDF, where the correlations range from $0.75\text{-}0.79$ and $0.79\text{-}0.86$, respectively. Frozen-FT generally remains stronger than prototype-based transfer, including at 15 shots. However, prototype-based transfer outperforms frozen-FT for seven of the 42 model–dataset pairs in at least two shot settings, including LaBraM, BrainWave, and CBraMod on SZ-28; MBrain, CBraMod, and BrainBERT on ADFD; and BrainWave on ADHD-Adult. These recurring gains across shot settings are unlikely to arise from a single support-set sample, suggesting that prototype transfer can effectively exploit the geometry in frozen representations for some model-dataset pairs.
\section{Exploratory Structural Analyses}
\label{sec:structural}
\textbf{Q3: What structural patterns can be observed in BFM representations?} To address this question, we examine spatial sensitivity, frequency-domain structure, and discrete representations. These analyses characterize structural tendencies rather than establishing causal learning mechanisms.

\begin{figure}[!ht]
  \centering
  \includegraphics[width = 0.43\textwidth]{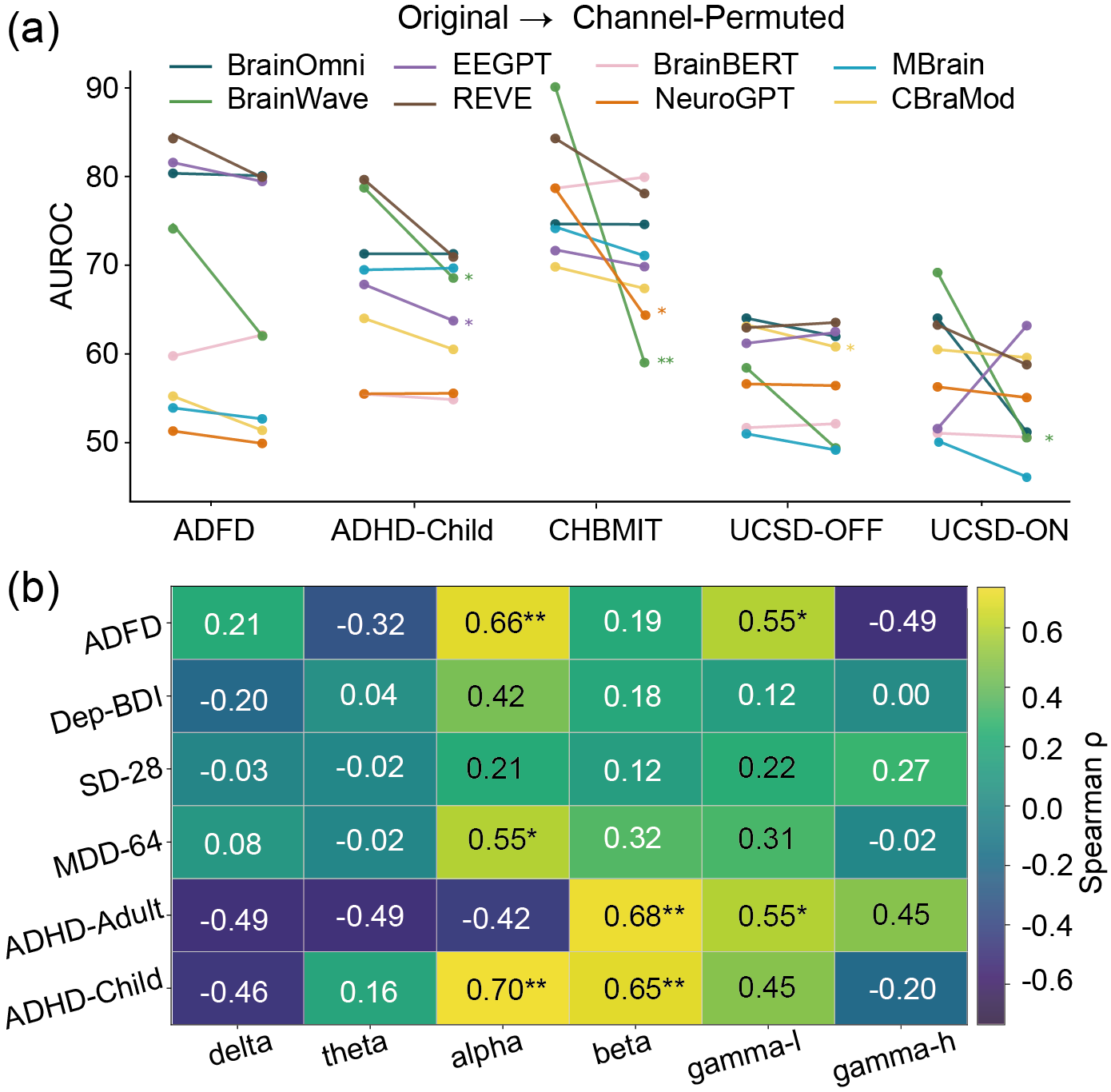}
\caption{Structural Analyses.  
    (a) AUROC changes under channel permutation. 
    (b) Spearman correlation between band-wise predictability and model performance rankings. Asterisks indicate * ($p<0.05$) and ** ($p<0.01$).
    }
  \label{fig:exp_ans}
\end{figure}

\subsection{Spatial Sensitivity}
Several high-performing BFMs in our benchmark, including BrainWave, BrainOmni, and REVE, explicitly model channel identities or cross-channel interactions, motivating an analysis of their sensitivity to spatial organization. We therefore jointly permute signal channels and their corresponding channel identifiers during training, preserving signal-channel correspondence, while evaluating on the canonical test split. 
We compare models with explicit spatial modeling (BrainWave, BrainOmni, CBraMod, EEGPT, MBrain, and REVE) with models that use less explicit spatial information (BrainBERT and NeuroGPT-E) by measuring changes in AUROC and applying paired significance tests (Figure~\ref{fig:exp_ans}(a)); complete results are reported in the supplementary material.

Performance changes under channel permutation vary across tasks, indicating different dependencies on spatial organization. We use spatial sensitivity to encompass both spatial heterogeneity in region-dependent channel statistics and topological sensitivity to relative channel organization. The perturbation preserves per-channel temporal statistics while changing the canonical input order and, for order-dependent spatial modules, the induced channel topology, thereby measuring the combined effects of these two factors. 
Epilepsy tasks show the largest AUROC drops, consistent with the importance of spatially distributed seizure onset and propagation patterns~\cite{VANMIERLO20202600}. PD exhibits intermediate sensitivity, in line with previously reported region-dependent spectral abnormalities~\cite{barone2021understanding}.
ADHD shows smaller changes, suggesting weaker dependence on the presented channel organization under our evaluation setting, despite reported frontal-network alterations~\cite{murias2007functional}. Most models are also relatively stable on AD, where commonly used EEG biomarkers are dominated by broad spectral changes rather than narrowly localized activity patterns~\cite{babiloni2004abnormal}.

Robustness to channel permutation also differs across model architectures. BrainBERT, which does not explicitly model cross-channel interactions, is largely insensitive to the perturbation, whereas NeuroGPT-E degrades on several datasets. Among models with explicit spatial components, BrainWave, which uses CNNs to capture spatial information, shows notable drops on CHBMIT, ADHD-Child, and ADFD, indicating sensitivity to the channel organization presented during training. EEGPT also remains sensitive on several spatially heterogeneous tasks despite its channel-alignment mechanism. 
By contrast, BrainOmni, CBraMod, MBrain, and REVE are generally more stable across datasets. These models incorporate graph structures, spatial embeddings, or semantic channel modeling to represent multichannel signals. 
However, robustness to channel reconfiguration is not sufficient: spatial modules that rely on fixed channel layouts remain difficult to transfer across iEEG patients with heterogeneous implantations. Learning structure-aware yet layout-flexible representations remains an open challenge.

\subsection{Frequency-Domain Associations}
Several BFMs that explicitly reconstruct frequency-domain features, including BrainWave, BrainBERT, and LaBraM, perform strongly
on multiple tasks. This motivates us to examine whether their task-adapted representations retain task-relevant spectral information, given that neural oscillations across frequency bands are associated with distinct cognitive and pathological states~\cite{buzsaki2004neuronal}.
We freeze each fine-tuned encoder and train a lightweight PSD probe to predict the spectral profile of six frequency bands (Figure~\ref{fig:exp_ans}(b)). Separate linear heads map the frozen representation to the PSD of each band. Let $P_b(x)$ and $\hat{P}_b(x)$ denote the ground-truth and predicted PSD for band $b$. We quantify band predictability as $r_b=\operatorname{corr}(P_b(x),\hat{P}_b(x))$ and normalize it across bands to obtain the relative predictive strength $r_b^{n}$. We then compute Spearman correlations between $r_b^{n}$ and downstream model rankings, primarily based on AUROC. Complete results are reported in the supplementary material.

The associations are strongly task dependent. On ADFD, the relative predictability of $\alpha$ and low-$\gamma$ is significantly associated with downstream ranking, whereas no band shows a significant association on the depression datasets. On ADHD-Adult, $\beta$ and low-$\gamma$ show significant positive correlations; on
ADHD-Child, significant associations are observed for $\alpha$ and $\beta$. These patterns are broadly consistent with prior reports of frequency-specific biomarkers, including altered $\delta/\alpha$ and $\theta/\alpha$ ratios in AD ~\cite{newson2019eeg} and $\theta/\beta$-related markers in ADHD~\cite{loo2013characterization}. 
These results indicate that task-adapted BFM representations emphasize different frequency bands across tasks, some of which are associated with downstream performance.

\subsection{Codebook Representations}
Discrete codebooks have recently been adopted in several BFMs to compress continuous signals or latent features into reusable tokens. We analyze three models with explicit discretization (BFM, LaBraM, and BrainOmni), which differ in where and how quantization is applied. BFM uses the Chronos tokenizer to map raw signals into token IDs. LaBraM learns a VQ-VAE codebook over encoder features during pretraining but removes it during downstream fine-tuning. BrainOmni adopts a related design with multi-level residual vector quantization (RVQ). On the test set, we characterize token usage through entropy (Ent), and evaluate token-embedding geometry using class-center similarity ($\text{Inter}$) and the inter-to-intra-class distance ratio (DR).

\begin{table}[]
\setlength{\tabcolsep}{1.6pt}
\centering
{\footnotesize
\begin{tabular}{lccccccccc}
\toprule
              & \multicolumn{3}{c}{ADFD}    & \multicolumn{3}{c}{CHBMIT}  & \multicolumn{3}{c}{SZ-28}  \\ \midrule
Model    & Ent & Inter & DR  & Ent & Inter & DR  & Ent & Inter & DR   \\ \midrule
BFM      & .868 & .998  & .002  & .838 & .999  & .001  & .853 & .999  & .001 \\
BrainOmni$_1$  & \textbf{.947} & .996  & .007  & .948 & .999  & .001  & .941 & .999  & .002 \\
BrainOmni$_2$  & .891 & .928  & .077  & .950 & .966  & .037 & \textbf{.982} & .994  & .007 \\
BrainOmni$_3$ & .929 & .927  & .076  & .970 & .959  & .043 & .973 & .954  & .049 \\
BrainOmni$_4$ & .946 & .947  & .055 & \textbf{.988} & .930  & .072  & .972 & .909  & .094 \\
LaBraM$_{no}$ & .755 & .995  & .006 & .721 & .973  & .038 & .607 & .986  & .022 \\
LaBraM    & .694 & \textbf{.267}  & \textbf{1.155} & .518 & \textbf{-.954} & \textbf{6.755} & .586 & \textbf{.800}  & \textbf{.294} \\ \bottomrule
\end{tabular}
}
\caption{Codebook usage and class geometry across datasets. Ent, Inter, and DR denote entropy, mean inter-class similarity, and inter-/intra-class distance ratio. $\mathrm{LaBraM}_{\mathrm{no}}$ uses the frozen codebook, and $\mathrm{BrainOmni}_{1\text{-}4}$ denote 4 RVQ levels.
}
\label{tab:codebook}
\end{table}

The three tokenization schemes show distinct relationships between codebook utilization and class geometry (Table \ref{tab:codebook}).
BFM uses a moderate fraction of its vocabulary but distributes assignments relatively evenly among active tokens; however, its high $\text{Inter}$ and low DR indicate weak class separation. LaBraM uses fewer and less evenly distributed tokens and shows stronger class separation than BFM, although its frozen codebook remains weakly discriminative without downstream adaptation. BrainOmni maintains near-complete utilization and high entropy across RVQ levels, with deeper levels generally showing larger DR, particularly on CHBMIT and SD-28.
Retaining and updating LaBraM's codebook during full fine-tuning further reduces coverage and entropy while increasing class separation. This variant outperforms frozen transfer but remains below standard full fine-tuning without quantization. By contrast, retaining BrainOmni's codebook yields a modest but statistically non-significant improvement over its no-codebook variant.
Overall, discrete tokenization is not uniformly beneficial for electrical brain signal decoding. Its effect varies across models and tasks and should be interpreted jointly with codebook utilization, class geometry, and downstream adaptation.
\section{Conclusion}
Brain4FMs provides a unified and extensible framework for evaluating BFMs on EEG/iEEG classification. We benchmark 17 models across 21 public datasets under standardized cross-subject protocols. 
Our goal is to assess how well existing BFMs transfer to EEG/iEEG analysis. Beyond task performance, we examine modality-dependent behavior, the effects of adaptation, and the representation structure. Together, these analyses characterize the current capabilities and limitations of BFMs and highlight the need for stronger generalization across subjects, devices, and tasks.
The supplementary material provides detailed benchmark settings and complete experimental results.
We currently focus on classification because it remains central to brain-signal analysis and is the primary downstream setting supported by existing BFMs. Generative tasks are not yet covered and will be incorporated as the field and available models mature.

\bibliography{bfms}

\clearpage
\appendix

\begin{center}
    {\LARGE\bfseries Appendix}
\end{center}

\section{Experimental Details}
This section details the dataset preprocessing and experimental protocols used in our benchmark. We describe the unified evaluation setup adopted across models and tasks, together with the learning-rate search and model-compatible optimization settings used to ensure consistent and reproducible comparisons.
\subsection{Dataset Details}
\begin{table*}[!hbt]
\renewcommand{\arraystretch}{1}
\setlength{\tabcolsep}{2pt}
\centering
\footnotesize
{\footnotesize
\begin{tabular}{lcccccccccc}
\toprule
Name & Signal & Task & Subject & \#Ch & SeqLen & sfreq (Hz) & Passband (Hz) & Total Time & Fold & Cat. \\ \midrule
CHBMIT & EEG & Epilepsy & 23 sub. & 23 & 10s & 256 & 0.01-256/3  & 686 h & 5 & 2 \\
MAYO & iEEG & DRE & 25 sub. & 1 & 3s & $5000\to1000$ & 0.01-1000/3  & 94 h & 6 & 2 \\
FNUSA & iEEG & DRE & 14 sub. & 1 & 3s & $5000\to1000$ & 0.01-1000/3  & 150 h & 5 & 2 \\
HUP-ECoG & iEEG & DRE  & 20 sub. & 54-125 & 2s & $1024\to512$ & 0.5-120 & 1020 min & 5 & 2 \\
HUP-SEEG & iEEG  & DRE  & 38 sub. & 36-195 & 2s & $1024\to512$ & 0.5-120 & 38 h & 5 & 2 \\
SWEC & iEEG  & DRE  & 68 sub. & 22-128 & 5s & $512/1024\to512$ & 0.5-150 & 9328 h & 5 & 2 \\
Dep-BDI & EEG & Depression & 122 sub. & 64 & 10s & $500\to250$ & 0.01-250/3  & 32 h & 5 & 2 \\
Dep-STAI & EEG & Anxiety & 122 sub. & 64 & 10s & $500\to250$ & 0.01-250/3  & 32 h & 5 & 3 \\
MDD-64 & EEG & MDD & 30H, 43MDD & 19 & 10s & 256 & 0.5-70  & 21 h & 5 & 2 \\
SZ-28 & EEG & SZ & 28 sub. & 19 & 5s & 250 & 0.2-250/3  & 958 min & 5 & 2 \\
UCSD-ON & EEG & PD & 31H, 15PD & 32 & 10s & $512\to256$ & 0.01-256/3  & 220 min & 5 & 2 \\
UCSD-OFF & EEG & PD & 31H, 15PD & 32 & 10s & $512\to256$ & 0.01-256/3  & 220 min & 5 & 2 \\
ADFD & EEG & AD & 88 sub. & 19 & 10s & $500\to250$ & 0.01-250/3  & 1164 min & 5 & 2 \\
ADHD-Adult & EEG & ADHD & 42H, 37ADHD & 2 & 5s & 256 & 0.01-256/3  & 423 min & 5 & 2 \\
ADHD-Child & EEG & ADHD & 60H, 61ADHD & 19 & 5s & 128 & 0.5-128/3  & 275 min & 5 & 2 \\ \midrule
ISRUC-G1 & EEG & Sleep Stage & 100 sub. & 6 & 30s & 200 & 0.3-45  & 745 h & 5 & 5 \\
SleepEDF & EEG & Sleep Stage & 44 sub. & 1 & 30s & 100 & 0.3-45  & 378.7 h & 5 & 5 \\ \midrule
EEGMMIDB-R & EEG & ME & 109 sub. & 64 & 4s & 160 & 0.01-70  & 2834 min & 6 & 4 \\
EEGMMIDB-I & EEG & MI & 109 sub. & 64 & 4s & 160 & 0.01-70  & 2834 min & 6 & 4 \\
BCI-2a & EEG & MI & 9 sub. & 22 & 3s & 250 & 0.5-100 & 130 min & 4 & 4 \\
Chisco-R & EEG & Concept Classification & 5 sub. & 125 & 3.3s & $1000 \to 500$ & 1–Nyquist (<250) & 59 h & 5 & 39 \\
Chisco-I & EEG & Concept Classification & 5 sub. & 125 & 3.3s & $1000 \to 500$ & 1–Nyquist (<250) & 59 h & 5 & 39 \\ 
Cogitate-CF & iEEG & Visual Classification &  19 sub. & 6-142 & 2s & $2048 \to 512$ & 0.1-150 & 120 min & 5 & 2 \\ \midrule
DEAP & EEG & Emotion Recognition & 32 sub. & 32 & 10s & 128 & 4-45  & 2133 min & 5 & 4 \\
SEED-IV & EEG & Emotion Recognition & 15 sub. & 62 & 4s & 200 & 1-75  & 54 h & 6 & 4 \\
EEGMat & EEG & MW & 36 sub. & 32 & 10s & $500\to250$ & 0.1-75  & 144 min & 4 & 2 \\ \bottomrule
\end{tabular}}
\caption{Public EEG and iEEG datasets used in the benchmark, including signal type, task, subjects (Sub.), number of channels (\#Ch), sequence length (SeqLen), sampling frequency (sfreq), effective preprocessing passband (Passband), total recording duration, cross-validation folds (Fold) and number of categories (Cat.); H denotes healthy subjects.}
\label{tab:dataset-total}
\end{table*}
Table~\ref{tab:dataset-total} summarizes dataset characteristics omitted from the main text for brevity. We select datasets by balancing diversity, practical availability, and minimal dataset-level overlap with the pretraining corpora of the evaluated BFMs. These details reflect the dataset acquisition protocols and facilitate benchmark reproduction.

All datasets are processed under consistent cross-subject cross-validation protocols. Depending on the dataset and task protocol, preprocessing may include band-pass filtering, notch filtering, downsampling, segmentation into fixed-length non-overlapping samples or trials, removal of non-EEG/iEEG channels (e.g., EMG and EOG), and per-channel z-score normalization. Preprocessing is defined at the dataset level and shared by all models to avoid model-specific signal-processing bias. Chisco and DEAP use the officially preprocessed data. For the remaining datasets, we apply notch filtering at 50 or 60 Hz to remove power-line noise when applicable. The complete dataset-specific preprocessing configurations, including band-pass ranges, target sampling rates, and segment lengths, are reported in Table~\ref{tab:dataset-total}.

We do not impose an additional unified bipolar montage or re-referencing scheme. Instead, we preserve the channel montage and reference configuration released with each public dataset. For example, ISRUC-G1 retains its original contralateral-mastoid reference setting, while CHBMIT retains its original bipolar montage. Dataset-specific preprocessing procedures are implemented in our open-source code and documented in the \texttt{data\_preprocess} directory. Since preprocessing and storage are performed independently within each subject or patient, no cross-subject information leakage is introduced. All datasets used in Brain4FMs are publicly released; ethics and fairness considerations are therefore limited, as no new data collection is involved.

Potential overlap between downstream datasets and pretraining corpora may introduce uncontrolled bias, which is a common challenge in evaluating released foundation models. We therefore minimized such overlap during downstream dataset selection whenever possible. Selected datasets are disjoint from the dominant reported pretraining sources, such as TUH~\cite{obeid2016temple} and TUEG~\cite{harati2014tuh}. 
The treatment of known pretraining-downstream overlap is transparent. The manuscript identifies the potentially overlapping NeuroLM/LaBraM on SEED-IV and MVPFormer on SWEC.

To preserve benchmark completeness and provide a transparent account of all available evaluations, we report the performance of every evaluated model-dataset pair, including pairs with potential pretraining overlap. These potentially affected entries are, however, excluded from all primary analyses and conclusions in the main text. 
Specifically, we exclude NeuroLM and LaBraM on SEED-IV, for which exposure to related emotion data cannot be ruled out, as well as MVPFormer on SWEC, since SWEC was used as part of its pretraining data. 
These results are retained in the appendix solely for coverage and transparency. Removing the potentially overlapping pairs may change individual scores and some aggregate summary values, but it does not materially alter the qualitative conclusions drawn from the main-text analyses. 

\subsection{Experimental Setup}
\label{app:setup}

\begin{table}[!hb]
\centering
\setlength{\tabcolsep}{0.5pt}
\resizebox{1\columnwidth}{!}{
\begin{tabular}{lccc}
\toprule
Model & Downstream Head & Head Params & Optimizer \\
\midrule
BrainBERT  & ReLU+Linear & $295296+385C$ & AdamW \\
Brant      & ReLU+Linear & $524800+513C$ & AdamW \\
SppEEGNet  & ReLU+Linear & $5050+51C$ & Adam \\
BFM        & ReLU+Linear & $524800+513C$ & AdamW \\
CBraMod    & ReLU+Linear & $20000S^2+100S(C+1)+C$ & AdamW \\
NeuroLM    & ReLU+Linear & $295296+385C$ & AdamW \\
LaBraM     & ReLU+Linear & $201C$ & AdamW \\
Bendr      & Norm+Linear & $513C$ & Adam \\
MBrain     & Linear & $229760+129C$ & Adam \\
NeuroGPT   & Linear & $1320224+33C$ & AdamW \\
EEGPT      & Linear & $1320224+33C$ & AdamW \\
BrainOmni  & Linear & $262144N_{\mathrm{ch}}+512+513C$ & AdamW \\
BrainWave  & CNN & $295296+385C$ & AdamW \\
BIOT       & ELU+Linear & $257C$ & Adam \\
REVE       & Norm+Linear & $512(\lfloor \frac{200(S-1)}{180} \rfloor+1)N_{\mathrm{ch}}(C+1)+C$ & AdamW \\
MVPFormer  & ReLU+Linear	& $295296+385C$	& AdamW \\
CodeBrain  & ReLU+Linear    & $20100+101C$  & AdamW \\
\bottomrule
\end{tabular}
}
\caption{Downstream classification interfaces, trainable head-parameter formulas, and optimizers.
The reported formulas count trainable parameters in the downstream classification head only, excluding the pretrained encoder. 
$C$ denotes the number of target classes, $S$ the effective sequence length, and $N_{\mathrm{ch}}$ the number of input channels.}
\label{tab:model-head}
\end{table}

We evaluate all models under a unified cross-subject cross-validation protocol. For each fold, subjects in the training, validation, and test sets are kept disjoint to prevent subject-level information leakage. Models are fine-tuned for up to 50 epochs with early stopping using a patience of 5. We use Adam-family optimizers, i.e., Adam or AdamW following the official implementations (\ref{tab:model-head}), with a batch size of 128 and no learning rate scheduler.

For downstream adaptation, we retain each BFM’s released input transformation, representation aggregation, classification interface, and compatible Adam-family optimizer, while applying a standardized training and hyperparameter-selection protocol. Specifically, data splits, preprocessing, learning-rate search, training budget, checkpoint selection, and evaluation metrics are standardized across models. This design preserves model-compatible readout mechanisms without reproducing model-specific fine-tuning recipes.

For transparency, all classification head types, head-parameter formulas, and optimizer choices are reported in Table~\ref{tab:model-head}. The number of trainable head parameters is computed after instantiating each model under the corresponding downstream configuration. It is task- and configuration-dependent, varying with the number of target classes $C$, the final representation dimension, and, for models whose flattened feature dimension depends on temporal tokenization, the sequence length $S$ and the number of channels $N_{\mathrm{ch}}$. Accordingly, Table~\ref{tab:model-head} reports the symbolic parameter count of the downstream classification head.

To facilitate exact reproduction, we release fixed subject-level splits, dataset-specific preprocessing scripts, model wrappers, pretrained weight version records, and all training and evaluation configuration files in the code repository. These artifacts ensure that all reported results can be reproduced under the same data splits, preprocessing pipeline, and fine-tuning protocol.

\subsection{Learning Rate Search}
\label{app:lr}
\begin{figure*}[!ht]
  \centering
  \includegraphics[width = 1\textwidth]{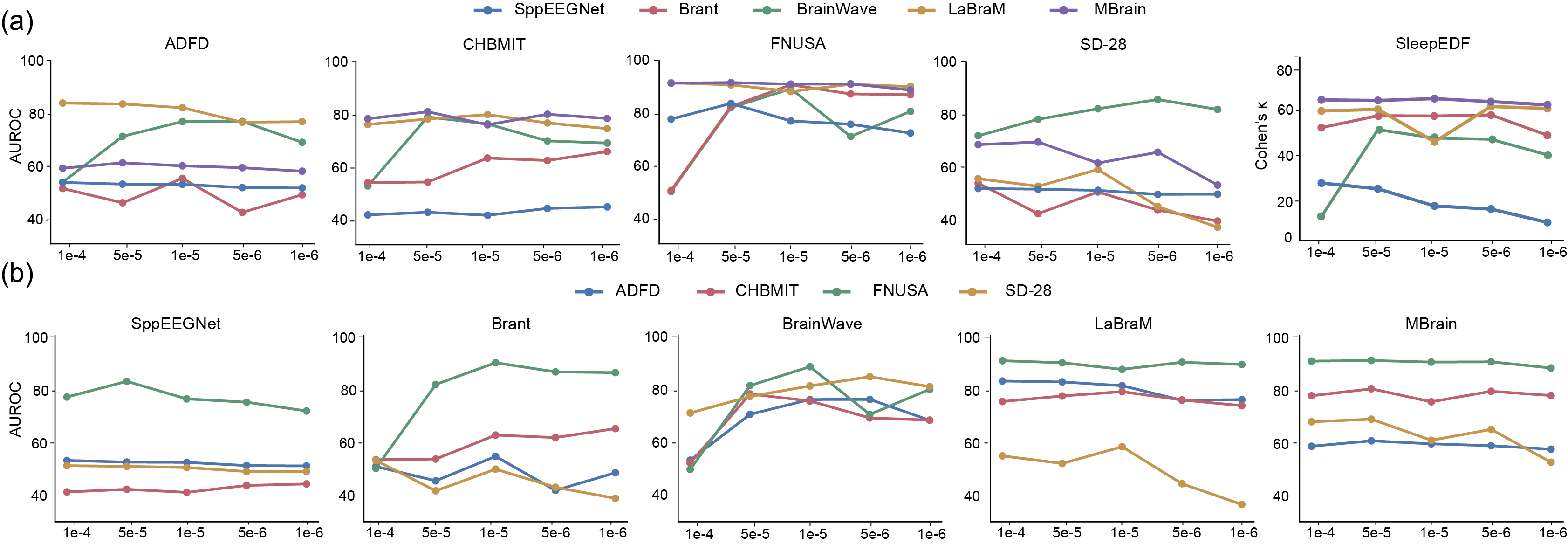}
\caption{
    Learning rate search for full-FT across datasets and models.
    (a) Results grouped by dataset, enabling direct comparison of different models on each dataset. 
    (b) Results grouped by model, highlighting dataset-specific LR sensitivity for each architecture.
    }
  \label{fig:lr_search}
\end{figure*}

\begin{table*}[!h]
\centering
\setlength{\tabcolsep}{0.8pt}
\resizebox{1.05\textwidth}{!}{
\begin{tabular}{lccccccccccccccccc}
\toprule
Dataset & BIOT & Bendr & BrainBERT & BrainOmni & BrainWave & Brant & CBraMod & EEGPT & LaBraM & MBrain & NeuroGPT & NeuroLM & REVE & SppEEGNet & BFM & MVPFormer & CodeBrain \\
\midrule
MAYO & $10^{-4}$ & $10^{-4}$ & $5\times10^{-5}$ & $5\times10^{-5}$ & $5\times10^{-5}$ & $5\times10^{-5}$ & $10^{-4}$ & $5\times10^{-5}$ & $5\times10^{-5}$ & $5\times10^{-5}$ & $5\times10^{-5}$ & $5\times10^{-5}$ & $10^{-5}$ & $10^{-4}$ & $10^{-5}$ & $5\times10^{-5}$ & $10^{-5}$ \\
SEED-IV & $5\times10^{-5}$ & $5\times10^{-5}$ & $10^{-5}$ & $5\times10^{-5}$ & $10^{-5}$ & $10^{-5}$ & $5\times10^{-5}$ & $10^{-5}$ & $5\times10^{-5}$ & $5\times10^{-5}$ & $10^{-5}$ & $10^{-5}$ & $10^{-5}$ & $5\times10^{-5}$ & $10^{-5}$ & $10^{-5}$ & $5\times10^{-5}$\\
EEGMMIDB-I & $5\times10^{-5}$ & $5\times10^{-5}$ & $10^{-5}$ & $5\times10^{-5}$ & $10^{-5}$ & $10^{-5}$ & $5\times10^{-5}$ & $10^{-5}$ & $5\times10^{-5}$ & $5\times10^{-5}$ & $10^{-5}$ & $10^{-5}$ & $10^{-5}$ & $5\times10^{-5}$ & $10^{-5}$ & $10^{-5}$ & $5\times10^{-5}$ \\
EEGMMIDB-R & $5\times10^{-5}$ & $5\times10^{-5}$ & $10^{-5}$ & $5\times10^{-5}$ & $10^{-5}$ & $10^{-5}$ & $5\times10^{-5}$ & $10^{-5}$ & $5\times10^{-5}$ & $5\times10^{-5}$ & $10^{-5}$ & $10^{-5}$ & $10^{-5}$ & $5\times10^{-5}$ & $10^{-5}$ & $10^{-5}$ & $5\times10^{-5}$ \\
UCSD-ON & $5\times10^{-5}$ & $5\times10^{-5}$ & $10^{-5}$ & $10^{-5}$ & $10^{-5}$ & $10^{-5}$ & $5\times10^{-5}$ & $10^{-5}$ & $10^{-5}$ & $10^{-5}$ & $10^{-5}$ & $10^{-5}$ & $10^{-5}$ & $5\times10^{-5}$ & $10^{-5}$ & $10^{-5}$ & $10^{-5}$\\
Dep-STAI & $5\times10^{-5}$ & $5\times10^{-5}$ & $10^{-5}$ & $5\times10^{-5}$ & $10^{-5}$ & $10^{-5}$ & $5\times10^{-5}$ & $10^{-5}$ & $5\times10^{-5}$ & $5\times10^{-5}$ & $10^{-5}$ & $10^{-5}$ & $10^{-5}$ & $5\times10^{-5}$ & $10^{-5}$ & $10^{-5}$ & $5\times10^{-5}$\\
FNUSA & $10^{-4}$ & $10^{-4}$ & $5\times10^{-5}$ & $5\times10^{-5}$ & $10^{-5}$ & $10^{-5}$ & $10^{-4}$ & $5\times10^{-5}$ & $10^{-4}$ & $5\times10^{-5}$ & $10^{-4}$ & $5\times10^{-5}$ & $5\times10^{-6}$ & $10^{-4}$ & $10^{-5}$ & $5\times10^{-5}$ & $5\times10^{-5}$ \\
UCSD-OFF & $5\times10^{-5}$ & $5\times10^{-5}$ & $10^{-5}$ & $10^{-5}$ & $10^{-5}$ & $10^{-5}$ & $5\times10^{-5}$ & $10^{-5}$ & $10^{-5}$ & $10^{-5}$ & $10^{-5}$ & $10^{-5}$ & $10^{-5}$ & $5\times10^{-5}$ & $10^{-5}$ & $10^{-5}$ & $10^{-5}$\\
Dep-BDI & $10^{-4}$ & $10^{-4}$ & $5\times10^{-5}$ & $5\times10^{-5}$ & $5\times10^{-5}$ & $5\times10^{-5}$ & $10^{-4}$ & $5\times10^{-5}$ & $5\times10^{-5}$ & $5\times10^{-5}$ & $5\times10^{-5}$ & $5\times10^{-5}$ & $10^{-5}$ & $10^{-4}$ & $10^{-5}$ & $5\times10^{-5}$ & $5\times10^{-5}$\\
DEAP & $5\times10^{-5}$ & $5\times10^{-5}$ & $10^{-5}$ & $5\times10^{-5}$ & $10^{-5}$ & $10^{-5}$ & $5\times10^{-5}$ & $10^{-5}$ & $5\times10^{-5}$ & $5\times10^{-5}$ & $10^{-5}$ & $10^{-5}$ & $10^{-5}$ & $5\times10^{-5}$ & $5\times10^{-5}$ & $10^{-5}$ & $5\times10^{-5}$\\
Chisco-R & $5\times10^{-5}$ & $5\times10^{-5}$ & $10^{-5}$ & $10^{-5}$ & $10^{-5}$ & $10^{-5}$ & $5\times10^{-5}$ & $10^{-5}$ & $10^{-5}$ & $10^{-5}$ & $10^{-5}$ & $10^{-5}$ & $10^{-5}$ & $5\times10^{-5}$ & $10^{-5}$ & $10^{-5}$ & $10^{-5}$\\
Chisco-I & $5\times10^{-5}$ & $5\times10^{-5}$ & $10^{-5}$ & $10^{-5}$ & $10^{-5}$ & $10^{-5}$ & $5\times10^{-5}$ & $10^{-5}$ & $10^{-5}$ & $10^{-5}$ & $10^{-5}$ & $10^{-5}$ & $10^{-5}$ & $5\times10^{-5}$ & $10^{-5}$ & $10^{-5}$ & $10^{-5}$\\
CHBMIT & $5\times10^{-5}$ & $5\times10^{-5}$ & $10^{-5}$ & $10^{-5}$ & $5\times10^{-5}$ & $10^{-6}$ & $10^{-4}$ & $10^{-5}$ & $10^{-5}$ & $5\times10^{-5}$ & $10^{-5}$ & $10^{-5}$ & $5\times10^{-5}$ & $10^{-4}$ & $10^{-6}$ & $10^{-5}$ & $5\times10^{-5}$\\
ADHD-Child & $5\times10^{-5}$ & $5\times10^{-5}$ & $10^{-5}$ & $5\times10^{-5}$ & $10^{-5}$ & $10^{-5}$ & $5\times10^{-5}$ & $10^{-5}$ & $5\times10^{-5}$ & $5\times10^{-5}$ & $10^{-5}$ & $10^{-5}$ & $10^{-5}$ & $5\times10^{-5}$ & $10^{-5}$ & $10^{-5}$ & $5\times10^{-5}$\\
ADFD & $10^{-4}$ & $10^{-4}$ & $5\times10^{-5}$ & $5\times10^{-5}$ & $5\times10^{-6}$ & $5\times10^{-5}$ & $10^{-4}$ & $5\times10^{-5}$ & $10^{-4}$ & $5\times10^{-5}$ & $5\times10^{-5}$ & $5\times10^{-5}$ & $10^{-5}$ & $10^{-4}$ & $10^{-5}$ & $10^{-5}$ & $10^{-5}$ \\
ADHD-Adult & $5\times10^{-5}$ & $5\times10^{-5}$ & $10^{-5}$ & $5\times10^{-5}$ & $10^{-5}$ & $10^{-5}$ & $5\times10^{-5}$ & $10^{-5}$ & $5\times10^{-5}$ & $5\times10^{-5}$ & $10^{-5}$ & $10^{-5}$ & $10^{-5}$ & $5\times10^{-5}$ & $10^{-5}$ & $10^{-5}$ & $5\times10^{-5}$ \\
MDD-64 & $10^{-4}$ & $10^{-4}$ & $10^{-5}$ & $5\times10^{-5}$ & $10^{-5}$ & $10^{-5}$ & $5\times10^{-6}$ & $10^{-5}$ & $10^{-5}$ & $10^{-5}$ & $10^{-5}$ & $10^{-5}$ & $5\times10^{-6}$ & $10^{-4}$ & $10^{-5}$ & $10^{-4}$ & $5\times10^{-5}$\\
SZ-28 & $5\times10^{-5}$ & $5\times10^{-5}$ & $10^{-5}$ & $5\times10^{-5}$ & $5\times10^{-6}$ & $10^{-5}$ & $5\times10^{-5}$ & $10^{-5}$ & $10^{-5}$ & $5\times10^{-5}$ & $10^{-5}$ & $10^{-5}$ & $10^{-5}$ & $10^{-4}$ & $5\times10^{-6}$ & $5\times10^{-5}$ & $5\times10^{-5}$\\
SleepEDF & $5\times10^{-5}$ & $5\times10^{-5}$ & $10^{-5}$ & $5\times10^{-5}$ & $5\times10^{-5}$ & $10^{-5}$ & $5\times10^{-5}$ & $10^{-5}$ & $5\times10^{-5}$ & $10^{-4}$ & $10^{-5}$ & $10^{-5}$ & $10^{-5}$ & $10^{-4}$ & $10^{-5}$ & $10^{-4}$ & $10^{-4}$\\
ISRUC & $5\times10^{-5}$ & $5\times10^{-5}$ & $10^{-5}$ & $5\times10^{-5}$ & $10^{-5}$ & $10^{-5}$ & $5\times10^{-5}$ & $10^{-5}$ & $5\times10^{-5}$ & $5\times10^{-5}$ & $10^{-5}$ & $10^{-5}$ & $10^{-5}$ & $5\times10^{-5}$ & $10^{-5}$ & $5\times10^{-5}$ & $10^{-4}$\\
BCI-2a & $5\times10^{-5}$ & $5\times10^{-5}$ & $10^{-5}$ & $10^{-5}$ & $10^{-5}$ & $10^{-5}$ & $5\times10^{-5}$ & $10^{-5}$ & $10^{-5}$ & $10^{-5}$ & $10^{-5}$ & $10^{-5}$ & $10^{-5}$ & $5\times10^{-5}$ & $10^{-5}$ & $10^{-5}$ & $5\times10^{-5}$ \\
EEGMat & $5\times10^{-5}$ & $5\times10^{-5}$ & $10^{-5}$ & $10^{-5}$ & $10^{-5}$ & $10^{-5}$ & $5\times10^{-5}$ & $10^{-5}$ & $10^{-5}$ & $10^{-5}$ & $10^{-5}$ & $10^{-5}$ & $10^{-5}$ & $5\times10^{-5}$ & $10^{-5}$ & $10^{-5}$ & $5\times10^{-5}$ \\
HUP-ECoG & $10^{-4}$ & - & $5\times10^{-5}$ & - & $5\times10^{-5}$ & - & $10^{-4}$ & - & $5\times10^{-5}$ & $5\times10^{-5}$ & - & - & - & $10^{-4}$ & $10^{-5}$ & $10^{-5}$ & $10^{-5}$ \\
HUP-SEEG & $10^{-4}$ & - & $5\times10^{-5}$ & - & $5\times10^{-5}$ & - & $10^{-4}$ & - & $5\times10^{-5}$ & $5\times10^{-5}$ & - & - & - & $10^{-4}$ & $10^{-5}$ & $5\times10^{-5}$ & $10^{-5}$ \\
Cogitate-CF & $5\times10^{-5}$ & - & $5\times10^{-5}$ & - & $10^{-5}$ & - & $10^{-4}$ & - & $5\times10^{-5}$ & $5\times10^{-5}$ & - & - & - & $10^{-4}$ & $5\times10^{-5}$ & $10^{-5}$ & $5\times10^{-5}$ \\
SWEC & $10^{-4}$ & - & $5\times10^{-5}$ & - & $5\times10^{-5}$ & - & $10^{-4}$ & - & $5\times10^{-5}$ & $5\times10^{-5}$ & - & - & - & $10^{-4}$ & $10^{-5}$ & $10^{-5}$ & $10^{-5}$ \\
\bottomrule
\end{tabular}
}
\caption{Best learning rates selected for each dataset-model pair in our benchmark.}
\label{tab:best_lr}
\end{table*}

To examine the sensitivity of full-FT to learning rate (LR), we conduct a focused LR search on 5 representative BFMs spanning different model scales. These include the lightweight SppEEGNet (138K), medium-scale models LaBraM (5.80M) and MBrain (8.34M), and large-scale models BrainWave (102.13M) and Brant (196.10M). Figure~\ref{fig:lr_search} reports their AUROC under five candidate encoder lrs, i.e., $\{1\times10^{-4}, 5\times10^{-5}, 1\times10^{-5}, 5\times10^{-6}, 1\times10^{-6}\}$, on 5 downstream datasets: ADFD, CHBMIT, FNUSA, SZ-28, and SleepEDF. For all experiments, the downstream head LR is set to $10\times$ the encoder LR.

Figure~\ref{fig:lr_search} reports the task-specific primary validation metric under different encoder lrs. Following the main evaluation protocol, AUROC is used for binary classification datasets, while Cohen's $\kappa$ is used for the multi-class SleepEDF dataset. The results show that model performance is highly sensitive to the choice of LR. First, the optimal LR differs across models even on the same dataset. For example, on CHBMIT, Brant tends to favor smaller lrs, whereas SppEEGNet achieves better performance with relatively larger lrs. This indicates that different model architectures can exhibit distinct fine-tuning dynamics, even when evaluated on the same downstream task. The optimal LR for the same model also varies across datasets. For instance, Brant achieves its best performance with an encoder LR of $1\times10^{-5}$ on FNUSA, while its optimal encoder LR shifts to $1\times10^{-6}$ on CHBMIT. This dataset-dependent behavior suggests that the fine-tuning stability of a given model is affected not only by the model architecture, but also by the characteristics of the downstream dataset.

These observations demonstrate that using a fixed LR for all models and datasets would introduce model- and dataset-dependent bias. Therefore, we perform LR search independently for each model-dataset pair. For each fold, we select the checkpoint with the lowest validation loss for evaluation. The best encoder LR is chosen according to the average task-specific primary validation metric across folds, i.e., AUROC for binary tasks and Cohen's $\kappa$ for multi-class tasks. Final performance is reported as the average test performance across folds using the selected encoder LR and its corresponding head LR. The selected encoder LR for each model-dataset pair used in full-FT is summarized in Table~\ref{tab:best_lr}.


\section{Additional Experimental Results}
This section reports the complete results underlying the main findings reported in the paper. We use full-parameter fine-tuning as the primary setting because Brain4FMs evaluates the downstream performance attainable after adapting each pretrained model to the target task, whereas frozen-encoder transfer serves as a complementary measure of fixed-representation transferability. Empirically, encoder adaptation improves AUROC by 3.26 points on average, with particularly large gains for some model–dataset pairs. It includes the full benchmark results under full-parameter fine-tuning, detailed frozen-encoder and prototype-based transfer results, and complete exploratory analyses of spatial sensitivity, frequency-domain associations, and codebook representations.
\subsection{Complete Benchmark Results}
\label{app:res}
We report the complete evaluation results of all 17 BFMs across 26 downstream classification settings. All 17 BFMs are evaluated on 22 settings, while four heterogeneous multichannel iEEG settings support 10 compatible models (BIOT, LaBraM, CBraMod, BrainBERT, SppEEGNet, BrainWave, MBrain, BFM, CodeBrain and MVPFormer), yielding 414 model–setting pairs. For NeuroGPT, we additionally report both encoder-only (NeuroGPT-E) and encoder–decoder (NeuroGPT-D) variants on the 22 applicable settings, resulting in 436 reported pairs overall. 
For binary tasks, we report AUROC, Accuracy, F1, and F2 to capture threshold-free discrimination and positive-class detection under imbalance. For multi-class tasks, we report Accuracy, AUROC (one-vs-rest; OvR), macro-F1 (MF1), and Cohen’s $\kappa$ (Kappa) to measure overall discrimination, class-wise performance, and agreement beyond chance. We report results as mean ± standard deviation over n-fold cross-validation, and bold indicates the highest mean performance based on unrounded values. All metrics use two decimal places, except Cohen’s $\kappa$, which uses three decimals due to its small magnitude on some datasets. In \Cref{tab:app_ADFD,tab:app_ADHD_Adult,tab:app_FNUSA,tab:app_ADHD_Child,tab:app_BCI_2a,tab:app_CHBMIT,tab:app_Chisco_I,tab:app_Chisco_R,tab:app_DEAP,tab:app_Dep_BDI,tab:app_Dep_STAI,tab:app_EEGMat,tab:app_EEGMMIDB_I,tab:app_EEGMMIDB_R,tab:app_ISRUC,tab:app_MAYO,tab:app_MDD_64,tab:app_SD_28,tab:app_SEED_IV,tab:app_SleepEDF,tab:app_UCSD_OFF,tab:app_UCSD_ON,tab:app_Cogitate,tab:app_HUP_ECoG,tab:app_HUP_SEEG,tab:app_SWEC}, models are grouped by SSL paradigm and ranked by AUROC within each group.
\FloatBarrier
\begin{table}[!h]
\setlength{\tabcolsep}{2.2pt}
{\small
\begin{tabular}{llcccc}
\toprule
Type & Model & AUROC & Acc & F1 & F2 \\ \midrule
C & MBrain & .75$\pm$.08 & .71$\pm$.12 & .54$\pm$.10 & .60$\pm$.09 \\
 & BIOT & .56$\pm$.08 & .29$\pm$.05 & .41$\pm$.03 & .61$\pm$.02 \\
 & Bendr & .55$\pm$.03 & .59$\pm$.03 & .32$\pm$.02 & .36$\pm$.04 \\
 & SppEEGNet & .43$\pm$.05 & .60$\pm$.06 & .31$\pm$.04 & .33$\pm$.05 \\
\midrule
G & BrainWave & \textbf{.90$\pm$.03} & \textbf{.80$\pm$.12} & \textbf{.71$\pm$.09} & \textbf{.78$\pm$.04} \\
 & REVE & .84$\pm$.10 & .78$\pm$.12 & .63$\pm$.14 & .66$\pm$.13 \\
 & NeuroGPT-E & .78$\pm$.12 & .75$\pm$.07 & .50$\pm$.12 & .63$\pm$.16 \\
 & BrainBERT & .78$\pm$.06 & .75$\pm$.10 & .58$\pm$.08 & .61$\pm$.09 \\
 & BrainOmni & .75$\pm$.11 & .71$\pm$.12 & .51$\pm$.12 & .54$\pm$.12 \\
 & LaBraM & .75$\pm$.10 & .73$\pm$.10 & .53$\pm$.14 & .57$\pm$.17 \\
 & BFM & .74$\pm$.05 & .72$\pm$.03 & .51$\pm$.09 & .51$\pm$.14 \\
 & CBraMod & .69$\pm$.04 & .70$\pm$.08 & .46$\pm$.08 & .49$\pm$.14 \\
 & NeuroGPT-D & .61$\pm$.07 & .73$\pm$.03 & .18$\pm$.23 & .19$\pm$.25 \\
 & CodeBrain & .60$\pm$.08 & .74$\pm$.10 & .52$\pm$.09 & .52$\pm$.11 \\
 & Brant & .60$\pm$.09 & .72$\pm$.06 & .08$\pm$.12 & .07$\pm$.11 \\
\midrule
O & EEGPT & .71$\pm$.09 & .67$\pm$.11 & .46$\pm$.06 & .51$\pm$.15 \\
 & NeuroLM & .67$\pm$.06 & .54$\pm$.20 & .27$\pm$.23 & .38$\pm$.34 \\
 & MVPFormer & .55$\pm$.02 & .70$\pm$.02 & .29$\pm$.03 & .26$\pm$.04 \\
\bottomrule
\end{tabular}
}
\caption{Performance of BFMs on the CHBMIT.}
\label{tab:app_CHBMIT}
\end{table}

\FloatBarrier
\begin{table}[!h]
\setlength{\tabcolsep}{2.2pt}
{\small
\begin{tabular}{llcccc}
\toprule
Type & Model & AUROC & Acc & F1 & F2 \\ \midrule
C & Bendr & .93$\pm$.03 & .90$\pm$.02 & .69$\pm$.13 & .75$\pm$.11 \\
 & MBrain & .92$\pm$.04 & .92$\pm$.02 & .70$\pm$.11 & .73$\pm$.09 \\
 & BIOT & .90$\pm$.07 & .88$\pm$.05 & .63$\pm$.14 & .72$\pm$.13 \\
 & SppEEGNet & .80$\pm$.03 & .86$\pm$.05 & .57$\pm$.14 & .60$\pm$.10 \\
\midrule
G & BrainWave & \textbf{.98$\pm$.01} & .93$\pm$.02 & .81$\pm$.09 & \textbf{.86$\pm$.03} \\
 & BrainBERT & .97$\pm$.01 & \textbf{.95$\pm$.01} & \textbf{.81$\pm$.07} & .83$\pm$.07 \\
 & LaBraM & .96$\pm$.02 & .93$\pm$.02 & .76$\pm$.12 & .80$\pm$.09 \\
 & NeuroGPT-E & .96$\pm$.02 & .93$\pm$.02 & .71$\pm$.07 & .67$\pm$.08 \\
 & REVE & .92$\pm$.04 & .89$\pm$.04 & .66$\pm$.13 & .72$\pm$.11 \\
 & Brant & .92$\pm$.03 & .82$\pm$.12 & .58$\pm$.19 & .69$\pm$.14 \\
 & BrainOmni & .91$\pm$.06 & .91$\pm$.02 & .67$\pm$.13 & .69$\pm$.12 \\
 & CBraMod & .89$\pm$.04 & .87$\pm$.02 & .59$\pm$.12 & .64$\pm$.10 \\
 & CodeBrain & .89$\pm$.02 & .91$\pm$.03 & .74$\pm$.11 & .76$\pm$.11 \\
 & BFM & .81$\pm$.08 & .68$\pm$.08 & .43$\pm$.15 & .59$\pm$.14 \\
 & NeuroGPT-D & .77$\pm$.15 & .88$\pm$.03 & .23$\pm$.26 & .19$\pm$.21 \\
\midrule
O & EEGPT & .92$\pm$.03 & .90$\pm$.03 & .67$\pm$.10 & .70$\pm$.07 \\
 & MVPFormer & .68$\pm$.08 & .72$\pm$.06 & .34$\pm$.07 & .39$\pm$.10 \\
 & NeuroLM & .64$\pm$.09 & .72$\pm$.16 & .30$\pm$.20 & .37$\pm$.26 \\
\bottomrule
\end{tabular}
}
\caption{Performance of BFMs on the MAYO.}
\label{tab:app_MAYO}
\end{table}

\FloatBarrier
\begin{table}[!h]
\setlength{\tabcolsep}{2.2pt}
{\small
\begin{tabular}{llcccc}
\toprule
Type & Model & AUROC & Acc & F1 & F2 \\ \midrule
C & MBrain & .91$\pm$.08 & .87$\pm$.08 & .75$\pm$.13 & .76$\pm$.16 \\
 & Bendr & .88$\pm$.06 & .84$\pm$.06 & .73$\pm$.11 & .77$\pm$.10 \\
 & BIOT & .87$\pm$.07 & .83$\pm$.08 & .73$\pm$.14 & .77$\pm$.10 \\
 & SppEEGNet & .82$\pm$.09 & .84$\pm$.09 & .73$\pm$.09 & .73$\pm$.09 \\
\midrule
G & BrainWave & \textbf{.92$\pm$.05} & .89$\pm$.06 & \textbf{.83$\pm$.06} & \textbf{.83$\pm$.04} \\
 & NeuroGPT-E & .90$\pm$.06 & .86$\pm$.05 & .74$\pm$.04 & .72$\pm$.11 \\
 & BrainBERT & .90$\pm$.04 & \textbf{.89$\pm$.04} & .77$\pm$.09 & .79$\pm$.08 \\
 & LaBraM & .89$\pm$.08 & .82$\pm$.10 & .72$\pm$.15 & .76$\pm$.11 \\
 & BrainOmni & .88$\pm$.07 & .84$\pm$.05 & .73$\pm$.08 & .75$\pm$.09 \\
 & REVE & .87$\pm$.08 & .82$\pm$.07 & .71$\pm$.10 & .74$\pm$.08 \\
 & Brant & .87$\pm$.12 & .84$\pm$.07 & .75$\pm$.06 & .79$\pm$.08 \\
 & CodeBrain & .83$\pm$.10 & .85$\pm$.06 & .78$\pm$.06 & .76$\pm$.09 \\
 & BFM & .78$\pm$.12 & .69$\pm$.11 & .62$\pm$.05 & .69$\pm$.08 \\
 & NeuroGPT-D & .72$\pm$.12 & .77$\pm$.12 & .30$\pm$.38 & .27$\pm$.35 \\
 & CBraMod & .71$\pm$.13 & .78$\pm$.08 & .62$\pm$.13 & .64$\pm$.12 \\
\midrule
O & EEGPT & .90$\pm$.04 & .84$\pm$.06 & .75$\pm$.11 & .78$\pm$.08 \\
 & NeuroLM & .64$\pm$.14 & .72$\pm$.17 & .37$\pm$.28 & .38$\pm$.35 \\
 & MVPFormer & .59$\pm$.07 & .63$\pm$.06 & .37$\pm$.07 & .38$\pm$.11 \\
\bottomrule
\end{tabular}
}
\caption{Performance of BFMs on the FNUSA.}
\label{tab:app_FNUSA}
\end{table}

\FloatBarrier
\begin{table}[!h]
\setlength{\tabcolsep}{2.2pt}
{\small
\begin{tabular}{llcccc}
\toprule
Type & Model & AUROC & Acc & F1 & F2 \\ \midrule
C & MBrain & .57$\pm$.09 & .61$\pm$.12 & .02$\pm$.05 & .02$\pm$.04 \\
 & BIOT & .54$\pm$.08 & .61$\pm$.12 & .19$\pm$.13 & .40$\pm$.06 \\
 & SppEEGNet & .44$\pm$.07 & .52$\pm$.09 & .35$\pm$.10 & .36$\pm$.10 \\
\midrule
G & BrainWave & \textbf{.88$\pm$.07} & \textbf{.80$\pm$.11} & \textbf{.72$\pm$.13} & \textbf{.65$\pm$.18} \\
 & BrainBERT & .81$\pm$.12 & .73$\pm$.09 & .63$\pm$.11 & .60$\pm$.23 \\
 & BFM & .72$\pm$.11 & .71$\pm$.10 & .47$\pm$.13 & .41$\pm$.18 \\
 & CodeBrain & .67$\pm$.05 & .71$\pm$.06 & .33$\pm$.16 & .30$\pm$.20 \\
 & CBraMod & .64$\pm$.09 & .58$\pm$.10 & .39$\pm$.24 & .45$\pm$.31 \\
 & LaBraM & .51$\pm$.06 & .64$\pm$.13 & .00$\pm$.00 & .00$\pm$.00 \\
\midrule
O & MVPFormer & .59$\pm$.08 & .62$\pm$.07 & .46$\pm$.12 & .44$\pm$.22 \\
\bottomrule
\end{tabular}
}
\caption{Performance of BFMs on the HUP-ECoG.}
\label{tab:app_HUP_ECoG}
\end{table}

\FloatBarrier
\begin{table}[!h]
\setlength{\tabcolsep}{2.2pt}
{\small
\begin{tabular}{llcccc}
\toprule
Type & Model & AUROC & Acc & F1 & F2 \\ \midrule
C & SppEEGNet & .52$\pm$.02 & .52$\pm$.19 & .22$\pm$.16 & .25$\pm$.24 \\
 & MBrain & .50$\pm$.08 & .62$\pm$.15 & .08$\pm$.18 & .13$\pm$.28 \\
 & BIOT & .45$\pm$.08 & .68$\pm$.21 & .28$\pm$.39 & .30$\pm$.41 \\
\midrule
G & BrainWave & \textbf{.79$\pm$.11} & .74$\pm$.14 & \textbf{.50$\pm$.20} & \textbf{.45$\pm$.25} \\
 & BFM & .61$\pm$.02 & \textbf{.74$\pm$.01} & .19$\pm$.08 & .13$\pm$.06 \\
 & CodeBrain & .61$\pm$.04 & .69$\pm$.03 & .29$\pm$.10 & .25$\pm$.12 \\
 & BrainBERT & .57$\pm$.09 & .67$\pm$.14 & .00$\pm$.00 & .00$\pm$.00 \\
 & CBraMod & .55$\pm$.08 & .63$\pm$.17 & .38$\pm$.15 & .32$\pm$.13 \\
 & LaBraM & .53$\pm$.09 & .63$\pm$.19 & .34$\pm$.19 & .30$\pm$.17 \\
\midrule
O & MVPFormer & .54$\pm$.04 & .60$\pm$.18 & .27$\pm$.09 & .26$\pm$.20 \\
\bottomrule
\end{tabular}
}
\caption{Performance of BFMs on the HUP-SEEG.}
\label{tab:app_HUP_SEEG}
\end{table}

\FloatBarrier
\begin{table}[!h]
\setlength{\tabcolsep}{2.2pt}
{\small
\begin{tabular}{llcccc}
\toprule
Type & Model & AUROC & Acc & F1 & F2 \\ \midrule
C & BIOT & .66$\pm$.06 & .45$\pm$.19 & .41$\pm$.01 & .56$\pm$.10 \\
 & MBrain & .63$\pm$.05 & .75$\pm$.00 & .00$\pm$.00 & .00$\pm$.00 \\
 & SppEEGNet & .44$\pm$.05 & .57$\pm$.10 & .34$\pm$.14 & .42$\pm$.21 \\
\midrule
G & BrainWave & \textbf{.84$\pm$.08} & \textbf{.82$\pm$.08} & .61$\pm$.17 & .58$\pm$.16 \\
 & LaBraM & .83$\pm$.10 & .80$\pm$.07 & .59$\pm$.17 & .60$\pm$.21 \\
 & CBraMod & .78$\pm$.11 & .75$\pm$.09 & \textbf{.62$\pm$.05} & \textbf{.70$\pm$.06} \\
 & BrainBERT & .77$\pm$.08 & .61$\pm$.18 & .48$\pm$.07 & .54$\pm$.16 \\
 & BFM & .65$\pm$.10 & .73$\pm$.04 & .36$\pm$.21 & .37$\pm$.24 \\
 & CodeBrain & .59$\pm$.07 & .60$\pm$.17 & .44$\pm$.07 & .51$\pm$.11 \\
\midrule
O & MVPFormer & .62$\pm$.05 & .64$\pm$.10 & .43$\pm$.06 & .49$\pm$.11 \\
\bottomrule
\end{tabular}
}
\caption{Performance of BFMs on the SWEC.}
\label{tab:app_SWEC}
\end{table}

\FloatBarrier
\begin{table}[!h]
\setlength{\tabcolsep}{2.2pt}
{\small
\begin{tabular}{llcccc}
\toprule
Type & Model & AUROC & Acc & F1 & F2 \\ \midrule
C & BIOT & .56$\pm$.17 & .49$\pm$.12 & .58$\pm$.14 & .70$\pm$.24 \\
 & MBrain & .53$\pm$.12 & .50$\pm$.10 & .53$\pm$.12 & .52$\pm$.14 \\
 & Bendr & .52$\pm$.02 & .52$\pm$.01 & .54$\pm$.05 & .52$\pm$.07 \\
 & SppEEGNet & .52$\pm$.02 & .51$\pm$.02 & .51$\pm$.05 & .49$\pm$.07 \\
\midrule
G & REVE & \textbf{.84$\pm$.07} & \textbf{.77$\pm$.08} & \textbf{.79$\pm$.07} & .79$\pm$.09 \\
 & LaBraM & .83$\pm$.10 & .75$\pm$.07 & .77$\pm$.08 & .76$\pm$.09 \\
 & BrainOmni & .80$\pm$.07 & .71$\pm$.04 & .73$\pm$.05 & .72$\pm$.09 \\
 & BrainWave & .74$\pm$.05 & .68$\pm$.05 & .68$\pm$.10 & .73$\pm$.10 \\
 & BrainBERT & .59$\pm$.05 & .58$\pm$.05 & .60$\pm$.11 & .64$\pm$.15 \\
 & BFM & .59$\pm$.08 & .58$\pm$.09 & .59$\pm$.12 & .61$\pm$.09 \\
 & CBraMod & .55$\pm$.06 & .59$\pm$.04 & .41$\pm$.20 & .37$\pm$.20 \\
 & NeuroGPT-E & .51$\pm$.03 & .55$\pm$.01 & .71$\pm$.01 & .85$\pm$.01 \\
 & Brant & .50$\pm$.07 & .54$\pm$.04 & .70$\pm$.03 & .84$\pm$.04 \\
 & CodeBrain & .50$\pm$.09 & .69$\pm$.05 & .72$\pm$.04 & .72$\pm$.04 \\
 & NeuroGPT-D & .50$\pm$.03 & .55$\pm$.01 & .71$\pm$.01 & \textbf{.86$\pm$.01} \\
\midrule
O & EEGPT & .81$\pm$.05 & .72$\pm$.05 & .73$\pm$.05 & .70$\pm$.07 \\
 & MVPFormer & .52$\pm$.01 & .53$\pm$.03 & .56$\pm$.05 & .55$\pm$.06 \\
 & NeuroLM & .51$\pm$.04 & .53$\pm$.03 & .63$\pm$.09 & .71$\pm$.16 \\
\bottomrule
\end{tabular}
}
\caption{Performance of BFMs on the ADFD.}
\label{tab:app_ADFD}
\end{table}

\FloatBarrier
\begin{table}[!h]
\setlength{\tabcolsep}{2.2pt}
{\small
\begin{tabular}{llcccc}
\toprule
Type & Model & AUROC & Acc & F1 & F2 \\ \midrule
C & BIOT & .96$\pm$.02 & .90$\pm$.04 & .89$\pm$.04 & .90$\pm$.05 \\
 & MBrain & .95$\pm$.03 & .92$\pm$.04 & .91$\pm$.04 & .91$\pm$.06 \\
 & Bendr & .62$\pm$.03 & .61$\pm$.04 & .60$\pm$.05 & .61$\pm$.06 \\
 & SppEEGNet & .51$\pm$.04 & .54$\pm$.06 & .36$\pm$.30 & .36$\pm$.35 \\
\midrule
G & BrainOmni & .96$\pm$.02 & .91$\pm$.02 & .90$\pm$.03 & .89$\pm$.04 \\
 & BrainWave & .96$\pm$.03 & .91$\pm$.04 & .91$\pm$.04 & .91$\pm$.05 \\
 & REVE & .96$\pm$.02 & .91$\pm$.02 & .91$\pm$.02 & .90$\pm$.05 \\
 & Brant & .95$\pm$.03 & .89$\pm$.04 & .89$\pm$.04 & .90$\pm$.05 \\
 & LaBraM & .95$\pm$.04 & .91$\pm$.04 & \textbf{.92$\pm$.04} & .91$\pm$.04 \\
 & CBraMod & .95$\pm$.02 & .92$\pm$.04 & .91$\pm$.04 & .91$\pm$.07 \\
 & NeuroGPT-E & .91$\pm$.05 & .84$\pm$.06 & .83$\pm$.05 & .82$\pm$.03 \\
 & CodeBrain & .88$\pm$.08 & \textbf{.92$\pm$.03} & .92$\pm$.03 & \textbf{.92$\pm$.04} \\
 & BFM & .88$\pm$.04 & .82$\pm$.04 & .80$\pm$.04 & .81$\pm$.05 \\
 & NeuroGPT-D & .88$\pm$.05 & .82$\pm$.05 & .82$\pm$.04 & .85$\pm$.03 \\
 & BrainBERT & .74$\pm$.05 & .69$\pm$.05 & .61$\pm$.11 & .57$\pm$.14 \\
\midrule
O & EEGPT & \textbf{.96$\pm$.03} & .91$\pm$.04 & .90$\pm$.05 & .89$\pm$.08 \\
 & NeuroLM & .82$\pm$.14 & .75$\pm$.07 & .74$\pm$.06 & .77$\pm$.11 \\
 & MVPFormer & .50$\pm$.01 & .51$\pm$.05 & .51$\pm$.03 & .53$\pm$.02 \\
\bottomrule
\end{tabular}
}
\caption{Performance of BFMs on the ADHD-Adult.}
\label{tab:app_ADHD_Adult}
\end{table}

\FloatBarrier
\begin{table}[!h]
\setlength{\tabcolsep}{2.2pt}
{\small
\begin{tabular}{llcccc}
\toprule
Type & Model & AUROC & Acc & F1 & F2 \\ \midrule
C & BIOT & .75$\pm$.07 & .67$\pm$.04 & \textbf{.76$\pm$.03} & .86$\pm$.04 \\
 & MBrain & .69$\pm$.11 & .63$\pm$.08 & .66$\pm$.06 & .66$\pm$.04 \\
 & SppEEGNet & .53$\pm$.05 & .49$\pm$.04 & .48$\pm$.07 & .45$\pm$.08 \\
 & Bendr & .53$\pm$.02 & .51$\pm$.01 & .52$\pm$.02 & .49$\pm$.02 \\
\midrule
G & REVE & \textbf{.79$\pm$.07} & \textbf{.71$\pm$.04} & .74$\pm$.06 & .74$\pm$.08 \\
 & BrainWave & .78$\pm$.04 & .71$\pm$.04 & .74$\pm$.06 & .74$\pm$.12 \\
 & BrainOmni & .71$\pm$.09 & .62$\pm$.07 & .63$\pm$.09 & .61$\pm$.14 \\
 & BFM & .69$\pm$.11 & .69$\pm$.09 & .75$\pm$.07 & .75$\pm$.04 \\
 & Brant & .69$\pm$.10 & .58$\pm$.09 & .51$\pm$.14 & .44$\pm$.15 \\
 & LaBraM & .66$\pm$.12 & .64$\pm$.08 & .74$\pm$.06 & .70$\pm$.07 \\
 & CodeBrain & .64$\pm$.09 & .63$\pm$.06 & .69$\pm$.07 & .74$\pm$.11 \\
 & CBraMod & .64$\pm$.05 & .64$\pm$.06 & .73$\pm$.03 & .81$\pm$.05 \\
 & NeuroGPT-E & .55$\pm$.02 & .56$\pm$.03 & .72$\pm$.02 & \textbf{.86$\pm$.01} \\
 & BrainBERT & .55$\pm$.03 & .56$\pm$.03 & .56$\pm$.07 & .53$\pm$.09 \\
 & NeuroGPT-D & .47$\pm$.12 & .56$\pm$.06 & .64$\pm$.11 & .69$\pm$.18 \\
\midrule
O & EEGPT & .67$\pm$.13 & .64$\pm$.10 & .68$\pm$.10 & .68$\pm$.12 \\
 & NeuroLM & .64$\pm$.06 & .60$\pm$.05 & .71$\pm$.04 & .80$\pm$.05 \\
 & MVPFormer & .54$\pm$.02 & .57$\pm$.03 & .61$\pm$.02 & .60$\pm$.02 \\
\bottomrule
\end{tabular}
}
\caption{Performance of BFMs on the ADHD-Child.}
\label{tab:app_ADHD_Child}
\end{table}

\FloatBarrier
\begin{table}[!h]
\setlength{\tabcolsep}{2.2pt}
{\small
\begin{tabular}{llcccc}
\toprule
Type & Model & AUROC & Acc & F1 & F2 \\ \midrule
C & SppEEGNet & .53$\pm$.07 & .51$\pm$.06 & .59$\pm$.15 & \textbf{.71$\pm$.25} \\
 & Bendr & .52$\pm$.04 & .51$\pm$.03 & .48$\pm$.05 & .47$\pm$.05 \\
 & BIOT & .51$\pm$.19 & .49$\pm$.05 & .56$\pm$.27 & .69$\pm$.35 \\
 & MBrain & .50$\pm$.04 & .44$\pm$.04 & .32$\pm$.23 & .34$\pm$.29 \\
\midrule
G & BrainOmni & \textbf{.64$\pm$.09} & .57$\pm$.06 & .54$\pm$.16 & .57$\pm$.26 \\
 & CBraMod & .63$\pm$.05 & .58$\pm$.04 & .46$\pm$.09 & .40$\pm$.12 \\
 & REVE & .62$\pm$.19 & \textbf{.58$\pm$.17} & .58$\pm$.19 & .60$\pm$.24 \\
 & LaBraM & .59$\pm$.05 & .56$\pm$.03 & .57$\pm$.08 & .60$\pm$.15 \\
 & BrainWave & .58$\pm$.06 & .53$\pm$.10 & \textbf{.62$\pm$.12} & .71$\pm$.19 \\
 & NeuroGPT-E & .56$\pm$.06 & .54$\pm$.05 & .57$\pm$.12 & .62$\pm$.22 \\
 & BFM & .53$\pm$.03 & .49$\pm$.12 & .36$\pm$.09 & .47$\pm$.17 \\
 & Brant & .52$\pm$.10 & .45$\pm$.22 & .42$\pm$.29 & .58$\pm$.35 \\
 & BrainBERT & .51$\pm$.06 & .54$\pm$.06 & .49$\pm$.17 & .49$\pm$.20 \\
 & CodeBrain & .51$\pm$.08 & .54$\pm$.07 & .55$\pm$.14 & .57$\pm$.21 \\
 & NeuroGPT-D & .45$\pm$.07 & .51$\pm$.05 & .01$\pm$.01 & .00$\pm$.01 \\
\midrule
O & EEGPT & .61$\pm$.13 & .55$\pm$.08 & .56$\pm$.10 & .58$\pm$.17 \\
 & NeuroLM & .51$\pm$.07 & .50$\pm$.05 & .53$\pm$.15 & .61$\pm$.23 \\
 & MVPFormer & .49$\pm$.04 & .53$\pm$.07 & .53$\pm$.09 & .54$\pm$.10 \\
\bottomrule
\end{tabular}
}
\caption{Performance of BFMs on the UCSD-OFF.}
\label{tab:app_UCSD_OFF}
\end{table}

\FloatBarrier
\begin{table}[!h]
\setlength{\tabcolsep}{2.2pt}
{\small
\begin{tabular}{llcccc}
\toprule
Type & Model & AUROC & Acc & F1 & F2 \\ \midrule
C & Bendr & .51$\pm$.05 & .49$\pm$.05 & .43$\pm$.05 & .41$\pm$.06 \\
 & SppEEGNet & .49$\pm$.08 & .50$\pm$.11 & .46$\pm$.23 & .48$\pm$.27 \\
 & MBrain & .49$\pm$.15 & .46$\pm$.14 & .37$\pm$.17 & .34$\pm$.18 \\
 & BIOT & .46$\pm$.18 & .49$\pm$.05 & \textbf{.65$\pm$.05} & \textbf{.81$\pm$.04} \\
\midrule
G & BrainWave & \textbf{.69$\pm$.10} & .53$\pm$.08 & .60$\pm$.06 & .69$\pm$.14 \\
 & REVE & .63$\pm$.19 & .56$\pm$.13 & .45$\pm$.29 & .46$\pm$.33 \\
 & CodeBrain & .62$\pm$.11 & .56$\pm$.10 & .60$\pm$.11 & .63$\pm$.14 \\
 & CBraMod & .60$\pm$.12 & .49$\pm$.08 & .18$\pm$.08 & .13$\pm$.07 \\
 & NeuroGPT-E & .56$\pm$.10 & .53$\pm$.07 & .58$\pm$.10 & .64$\pm$.13 \\
 & Brant & .56$\pm$.11 & .52$\pm$.11 & .29$\pm$.28 & .31$\pm$.34 \\
 & BrainOmni & .53$\pm$.14 & .49$\pm$.12 & .49$\pm$.16 & .51$\pm$.20 \\
 & LaBraM & .51$\pm$.22 & .45$\pm$.11 & .33$\pm$.19 & .31$\pm$.19 \\
 & BrainBERT & .51$\pm$.05 & .51$\pm$.06 & .51$\pm$.11 & .54$\pm$.17 \\
 & NeuroGPT-D & .46$\pm$.10 & .47$\pm$.06 & .29$\pm$.25 & .31$\pm$.28 \\
 & BFM & .44$\pm$.08 & .48$\pm$.07 & .42$\pm$.17 & .42$\pm$.20 \\
\midrule
O & MVPFormer & .58$\pm$.07 & \textbf{.57$\pm$.05} & .50$\pm$.05 & .46$\pm$.05 \\
 & EEGPT & .51$\pm$.21 & .48$\pm$.17 & .49$\pm$.15 & .49$\pm$.14 \\
 & NeuroLM & .48$\pm$.13 & .51$\pm$.18 & .47$\pm$.24 & .48$\pm$.29 \\
\bottomrule
\end{tabular}
}
\caption{Performance of BFMs on the UCSD-ON.}
\label{tab:app_UCSD_ON}
\end{table}

\FloatBarrier
\begin{table}[!h]
\setlength{\tabcolsep}{2.2pt}
{\small
\begin{tabular}{llcccc}
\toprule
Type & Model & AUROC & Acc & F1 & F2 \\ \midrule
C & MBrain & .68$\pm$.06 & .75$\pm$.02 & .13$\pm$.15 & .10$\pm$.11 \\
 & BIOT & .59$\pm$.08 & .27$\pm$.01 & .42$\pm$.01 & .65$\pm$.01 \\
 & Bendr & .54$\pm$.02 & .66$\pm$.05 & .34$\pm$.07 & .34$\pm$.06 \\
 & SppEEGNet & .52$\pm$.03 & .50$\pm$.07 & .37$\pm$.04 & .47$\pm$.09 \\
\midrule
G & REVE & \textbf{.78$\pm$.09} & \textbf{.75$\pm$.08} & \textbf{.57$\pm$.10} & \textbf{.66$\pm$.10} \\
 & LaBraM & .71$\pm$.03 & .68$\pm$.04 & .48$\pm$.07 & .53$\pm$.13 \\
 & NeuroGPT-E & .70$\pm$.04 & .73$\pm$.04 & .25$\pm$.20 & .22$\pm$.20 \\
 & BrainOmni & .68$\pm$.07 & .60$\pm$.07 & .48$\pm$.04 & .60$\pm$.06 \\
 & BrainWave & .66$\pm$.06 & .51$\pm$.20 & .44$\pm$.07 & .55$\pm$.08 \\
 & CBraMod & .63$\pm$.06 & .72$\pm$.04 & .19$\pm$.18 & .16$\pm$.17 \\
 & BrainBERT & .61$\pm$.07 & .67$\pm$.06 & .45$\pm$.09 & .50$\pm$.13 \\
 & BFM & .60$\pm$.03 & .66$\pm$.04 & .35$\pm$.06 & .36$\pm$.10 \\
 & CodeBrain & .59$\pm$.05 & .62$\pm$.09 & .49$\pm$.04 & .58$\pm$.07 \\
 & Brant & .57$\pm$.12 & .75$\pm$.01 & .00$\pm$.00 & .00$\pm$.00 \\
 & NeuroGPT-D & .51$\pm$.06 & .73$\pm$.01 & .00$\pm$.00 & .00$\pm$.00 \\
\midrule
O & EEGPT & .67$\pm$.04 & .64$\pm$.06 & .49$\pm$.02 & .59$\pm$.04 \\
 & NeuroLM & .63$\pm$.08 & .31$\pm$.10 & .42$\pm$.04 & .64$\pm$.04 \\
 & MVPFormer & .56$\pm$.03 & .56$\pm$.03 & .37$\pm$.04 & .43$\pm$.04 \\
\bottomrule
\end{tabular}
}
\caption{Performance of BFMs on the EEGMat.}
\label{tab:app_EEGMat}
\end{table}

\FloatBarrier
\begin{table}[!h]
\setlength{\tabcolsep}{2.2pt}
{\small
\begin{tabular}{llcccc}
\toprule
Type & Model & AUROC & Acc & F1 & F2 \\ \midrule
C & MBrain & .58$\pm$.10 & .58$\pm$.10 & .64$\pm$.13 & .66$\pm$.15 \\
 & BIOT & .56$\pm$.09 & .62$\pm$.12 & .68$\pm$.18 & .65$\pm$.21 \\
 & Bendr & .49$\pm$.05 & .50$\pm$.03 & .48$\pm$.05 & .45$\pm$.04 \\
 & SppEEGNet & .49$\pm$.03 & .47$\pm$.04 & .33$\pm$.06 & .27$\pm$.05 \\
\midrule
G & BrainWave & \textbf{.88$\pm$.07} & \textbf{.81$\pm$.06} & \textbf{.83$\pm$.08} & .81$\pm$.11 \\
 & REVE & .81$\pm$.13 & .75$\pm$.13 & .80$\pm$.10 & \textbf{.86$\pm$.11} \\
 & NeuroGPT-D & .80$\pm$.17 & .77$\pm$.13 & .80$\pm$.11 & .84$\pm$.14 \\
 & BrainBERT & .73$\pm$.11 & .66$\pm$.15 & .63$\pm$.26 & .62$\pm$.29 \\
 & NeuroGPT-E & .72$\pm$.12 & .66$\pm$.14 & .73$\pm$.15 & .69$\pm$.19 \\
 & BrainOmni & .69$\pm$.15 & .69$\pm$.12 & .73$\pm$.14 & .77$\pm$.19 \\
 & BFM & .69$\pm$.15 & .62$\pm$.05 & .73$\pm$.04 & .81$\pm$.08 \\
 & CodeBrain & .65$\pm$.20 & .56$\pm$.08 & .61$\pm$.06 & .62$\pm$.15 \\
 & CBraMod & .54$\pm$.14 & .58$\pm$.05 & .58$\pm$.18 & .58$\pm$.24 \\
 & LaBraM & .54$\pm$.07 & .54$\pm$.04 & .59$\pm$.07 & .59$\pm$.13 \\
 & Brant & .51$\pm$.11 & .46$\pm$.08 & .47$\pm$.17 & .50$\pm$.30 \\
\midrule
O & EEGPT & .56$\pm$.21 & .56$\pm$.11 & .62$\pm$.12 & .66$\pm$.15 \\
 & NeuroLM & .49$\pm$.04 & .49$\pm$.05 & .53$\pm$.28 & .62$\pm$.35 \\
 & MVPFormer & .48$\pm$.05 & .51$\pm$.04 & .53$\pm$.09 & .54$\pm$.13 \\
\bottomrule
\end{tabular}
}
\caption{Performance of BFMs on the SZ-28.}
\label{tab:app_SD_28}
\end{table}

\FloatBarrier
\begin{table}[!h]
\setlength{\tabcolsep}{2.2pt}
{\small
\begin{tabular}{llcccc}
\toprule
Type & Model & AUROC & Acc & F1 & F2 \\ \midrule
C & BIOT & .93$\pm$.03 & .81$\pm$.07 & .85$\pm$.05 & \textbf{.93$\pm$.02} \\
 & MBrain & .93$\pm$.08 & .87$\pm$.11 & .89$\pm$.08 & .91$\pm$.05 \\
 & Bendr & .85$\pm$.07 & .78$\pm$.05 & .80$\pm$.02 & .81$\pm$.05 \\
 & SppEEGNet & .58$\pm$.02 & .60$\pm$.09 & .58$\pm$.12 & .54$\pm$.12 \\
\midrule
G & REVE & \textbf{.97$\pm$.04} & \textbf{.88$\pm$.09} & \textbf{.90$\pm$.06} & .92$\pm$.04 \\
 & LaBraM & .96$\pm$.04 & .87$\pm$.08 & .88$\pm$.07 & .89$\pm$.08 \\
 & BrainOmni & .94$\pm$.07 & .85$\pm$.09 & .87$\pm$.07 & .89$\pm$.05 \\
 & BrainWave & .94$\pm$.03 & .85$\pm$.02 & .84$\pm$.03 & .81$\pm$.06 \\
 & BFM & .93$\pm$.08 & .86$\pm$.11 & .88$\pm$.08 & .91$\pm$.05 \\
 & BrainBERT & .92$\pm$.07 & .85$\pm$.07 & .85$\pm$.06 & .85$\pm$.10 \\
 & Brant & .92$\pm$.09 & .80$\pm$.10 & .80$\pm$.14 & .80$\pm$.19 \\
 & CBraMod & .90$\pm$.13 & .84$\pm$.12 & .86$\pm$.09 & .87$\pm$.09 \\
 & NeuroGPT-E & .87$\pm$.05 & .80$\pm$.05 & .84$\pm$.03 & .90$\pm$.03 \\
 & CodeBrain & .77$\pm$.15 & .85$\pm$.09 & .86$\pm$.07 & .87$\pm$.09 \\
 & NeuroGPT-D & .70$\pm$.12 & .67$\pm$.13 & .77$\pm$.07 & .89$\pm$.04 \\
\midrule
O & EEGPT & .94$\pm$.05 & .87$\pm$.08 & .88$\pm$.06 & .90$\pm$.04 \\
 & NeuroLM & .82$\pm$.11 & .80$\pm$.06 & .80$\pm$.09 & .81$\pm$.18 \\
 & MVPFormer & .73$\pm$.09 & .83$\pm$.06 & .84$\pm$.04 & .86$\pm$.06 \\
\bottomrule
\end{tabular}
}
\caption{Performance of BFMs on the MDD-64.}
\label{tab:app_MDD_64}
\end{table}

\FloatBarrier
\begin{table}[!h]
\setlength{\tabcolsep}{2.2pt}
{\small
\begin{tabular}{llcccc}
\toprule
Type & Model & AUROC & Acc & F1 & F2 \\ \midrule
C & BIOT & .60$\pm$.11 & .30$\pm$.17 & .11$\pm$.02 & .22$\pm$.05 \\
 & MBrain & .59$\pm$.07 & .62$\pm$.03 & .24$\pm$.16 & .20$\pm$.16 \\
 & Bendr & .54$\pm$.03 & .54$\pm$.03 & .45$\pm$.03 & .47$\pm$.03 \\
 & SppEEGNet & .50$\pm$.02 & .49$\pm$.04 & .41$\pm$.04 & .45$\pm$.04 \\
\midrule
G & BrainWave & \textbf{.72$\pm$.03} & .66$\pm$.03 & \textbf{.62$\pm$.03} & .67$\pm$.06 \\
 & BrainOmni & .68$\pm$.12 & .63$\pm$.09 & .55$\pm$.10 & .58$\pm$.13 \\
 & REVE & .67$\pm$.12 & .65$\pm$.08 & .47$\pm$.14 & .45$\pm$.17 \\
 & LaBraM & .65$\pm$.10 & .59$\pm$.11 & .60$\pm$.09 & \textbf{.71$\pm$.08} \\
 & BrainBERT & .64$\pm$.09 & .61$\pm$.05 & .48$\pm$.11 & .48$\pm$.12 \\
 & BFM & .62$\pm$.10 & .59$\pm$.06 & .48$\pm$.10 & .51$\pm$.13 \\
 & NeuroGPT-E & .59$\pm$.08 & .62$\pm$.01 & .02$\pm$.02 & .01$\pm$.01 \\
 & CBraMod & .54$\pm$.05 & \textbf{.69$\pm$.14} & .13$\pm$.08 & .16$\pm$.10 \\
 & Brant & .53$\pm$.09 & .62$\pm$.01 & .00$\pm$.00 & .00$\pm$.00 \\
 & CodeBrain & .52$\pm$.11 & .60$\pm$.04 & .40$\pm$.13 & .41$\pm$.15 \\
 & NeuroGPT-D & .51$\pm$.08 & .62$\pm$.01 & .00$\pm$.00 & .00$\pm$.00 \\
\midrule
O & EEGPT & .63$\pm$.04 & .62$\pm$.05 & .50$\pm$.05 & .51$\pm$.11 \\
 & NeuroLM & .54$\pm$.05 & .61$\pm$.02 & .30$\pm$.26 & .31$\pm$.31 \\
 & MVPFormer & .51$\pm$.02 & .55$\pm$.02 & .39$\pm$.06 & .38$\pm$.09 \\
\bottomrule
\end{tabular}
}
\caption{Performance of BFMs on the Dep-BDI.}
\label{tab:app_Dep_BDI}
\end{table}

\FloatBarrier
\begin{table}[!h]
\setlength{\tabcolsep}{2.2pt}
{\small
\begin{tabular}{llcccc}
\toprule
Type & Model & Acc & AUROC & MF1 & Kappa \\ \midrule
C & BIOT & .37$\pm$.07 & .56$\pm$.09 & .33$\pm$.07 & .073$\pm$.088 \\
 & MBrain & .60$\pm$.03 & .55$\pm$.04 & .30$\pm$.05 & .054$\pm$.059 \\
 & Bendr & .48$\pm$.03 & .54$\pm$.03 & .36$\pm$.03 & .047$\pm$.041 \\
 & SppEEGNet & .46$\pm$.03 & .51$\pm$.02 & .33$\pm$.01 & .007$\pm$.024 \\
\midrule
G & BrainWave & .51$\pm$.05 & \textbf{.64$\pm$.06} & \textbf{.42$\pm$.05} & \textbf{.150$\pm$.053} \\
 & NeuroGPT-E & .55$\pm$.07 & .61$\pm$.09 & .38$\pm$.10 & .129$\pm$.155 \\
 & LaBraM & .36$\pm$.10 & .61$\pm$.09 & .26$\pm$.06 & .081$\pm$.093 \\
 & BrainBERT & .48$\pm$.11 & .60$\pm$.06 & .32$\pm$.08 & .064$\pm$.126 \\
 & BrainOmni & .53$\pm$.05 & .60$\pm$.06 & .36$\pm$.05 & .105$\pm$.062 \\
 & BFM & .45$\pm$.09 & .58$\pm$.08 & .37$\pm$.06 & .099$\pm$.117 \\
 & NeuroGPT-D & \textbf{.61$\pm$.01} & .57$\pm$.06 & .25$\pm$.00 & .000$\pm$.000 \\
 & REVE & .54$\pm$.10 & .54$\pm$.11 & .35$\pm$.08 & .079$\pm$.200 \\
 & CBraMod & .60$\pm$.02 & .54$\pm$.04 & .30$\pm$.04 & .049$\pm$.052 \\
 & Brant & \textbf{.61$\pm$.01} & .53$\pm$.03 & .25$\pm$.00 & .000$\pm$.000 \\
 & CodeBrain & .48$\pm$.08 & .52$\pm$.05 & .32$\pm$.04 & .034$\pm$.055 \\
\midrule
O & EEGPT & .51$\pm$.06 & .56$\pm$.06 & .34$\pm$.06 & .025$\pm$.080 \\
 & MVPFormer & .50$\pm$.04 & .54$\pm$.03 & .34$\pm$.01 & .028$\pm$.039 \\
 & NeuroLM & .58$\pm$.04 & .50$\pm$.04 & .26$\pm$.02 & .009$\pm$.018 \\
\bottomrule
\end{tabular}
}
\caption{Performance of BFMs on the Dep-STAI.}
\label{tab:app_Dep_STAI}
\end{table}

\FloatBarrier
\begin{table}[!h]
\setlength{\tabcolsep}{2.2pt}
{\small
\begin{tabular}{llcccc}
\toprule
Type & Model & Acc & AUROC & MF1 & Kappa \\ \midrule
C & BIOT & .68$\pm$.04 & .91$\pm$.02 & .65$\pm$.05 & .585$\pm$.049 \\
 & MBrain & .63$\pm$.07 & .87$\pm$.04 & .55$\pm$.07 & .514$\pm$.084 \\
 & Bendr & .52$\pm$.04 & .81$\pm$.02 & .49$\pm$.05 & .269$\pm$.076 \\
 & SppEEGNet & .26$\pm$.05 & .63$\pm$.03 & .21$\pm$.03 & .098$\pm$.038 \\
\midrule
G & REVE & \textbf{.69$\pm$.03} & \textbf{.92$\pm$.01} & \textbf{.65$\pm$.05} & \textbf{.598$\pm$.036} \\
 & BrainOmni & .68$\pm$.03 & .91$\pm$.02 & .64$\pm$.04 & .581$\pm$.030 \\
 & CodeBrain & .65$\pm$.03 & .90$\pm$.02 & .63$\pm$.05 & .557$\pm$.045 \\
 & BrainWave & .62$\pm$.05 & .89$\pm$.01 & .60$\pm$.02 & .522$\pm$.064 \\
 & CBraMod & .54$\pm$.08 & .86$\pm$.05 & .48$\pm$.10 & .419$\pm$.094 \\
 & LaBraM & .61$\pm$.02 & .85$\pm$.02 & .56$\pm$.05 & .502$\pm$.028 \\
 & Brant & .55$\pm$.01 & .84$\pm$.02 & .50$\pm$.04 & .427$\pm$.016 \\
 & NeuroGPT-E & .44$\pm$.11 & .80$\pm$.02 & .41$\pm$.11 & .319$\pm$.101 \\
 & BrainBERT & .47$\pm$.05 & .80$\pm$.03 & .42$\pm$.07 & .332$\pm$.058 \\
 & BFM & .47$\pm$.04 & .78$\pm$.03 & .37$\pm$.13 & .264$\pm$.148 \\
 & NeuroGPT-D & .31$\pm$.07 & .61$\pm$.07 & .17$\pm$.08 & .060$\pm$.102 \\
\midrule
O & EEGPT & .60$\pm$.05 & .87$\pm$.02 & .57$\pm$.08 & .486$\pm$.070 \\
 & MVPFormer & .38$\pm$.04 & .70$\pm$.02 & .33$\pm$.05 & .208$\pm$.044 \\
 & NeuroLM & .27$\pm$.11 & .67$\pm$.05 & .15$\pm$.10 & .074$\pm$.074 \\
\bottomrule
\end{tabular}
}
\caption{Performance of BFMs on the ISRUC-G1.}
\label{tab:app_ISRUC}
\end{table}

\FloatBarrier
\begin{table}[!h]
\setlength{\tabcolsep}{2.2pt}
{\small
\begin{tabular}{llcccc}
\toprule
Type & Model & Acc & AUROC & MF1 & Kappa \\ \midrule
C & BIOT & .78$\pm$.03 & \textbf{.95$\pm$.01} & \textbf{.74$\pm$.02} & .699$\pm$.043 \\
 & MBrain & \textbf{.78$\pm$.04} & .95$\pm$.01 & .74$\pm$.03 & \textbf{.709$\pm$.059} \\
 & Bendr & .69$\pm$.06 & .91$\pm$.01 & .65$\pm$.04 & .591$\pm$.062 \\
 & SppEEGNet & .43$\pm$.11 & .76$\pm$.03 & .39$\pm$.07 & .283$\pm$.075 \\
\midrule
G & CodeBrain & .73$\pm$.07 & .94$\pm$.02 & .71$\pm$.05 & .646$\pm$.087 \\
 & NeuroGPT-E & .78$\pm$.03 & .94$\pm$.01 & .70$\pm$.03 & .696$\pm$.045 \\
 & Brant & .73$\pm$.07 & .94$\pm$.01 & .70$\pm$.05 & .639$\pm$.083 \\
 & REVE & .75$\pm$.04 & .94$\pm$.02 & .71$\pm$.04 & .664$\pm$.048 \\
 & CBraMod & .73$\pm$.04 & .93$\pm$.01 & .69$\pm$.04 & .642$\pm$.054 \\
 & LaBraM & .76$\pm$.04 & .93$\pm$.01 & .68$\pm$.03 & .669$\pm$.050 \\
 & BrainBERT & .68$\pm$.07 & .91$\pm$.02 & .65$\pm$.05 & .578$\pm$.079 \\
 & BrainWave & .67$\pm$.05 & .89$\pm$.02 & .54$\pm$.08 & .516$\pm$.104 \\
 & BFM & .65$\pm$.04 & .89$\pm$.01 & .60$\pm$.03 & .558$\pm$.053 \\
 & BrainOmni & .63$\pm$.03 & .87$\pm$.00 & .60$\pm$.01 & .506$\pm$.025 \\
 & NeuroGPT-D & .48$\pm$.04 & .65$\pm$.03 & .21$\pm$.05 & .100$\pm$.068 \\
\midrule
O & EEGPT & .68$\pm$.03 & .91$\pm$.00 & .66$\pm$.01 & .578$\pm$.026 \\
 & MVPFormer & .49$\pm$.08 & .81$\pm$.02 & .45$\pm$.06 & .339$\pm$.068 \\
 & NeuroLM & .18$\pm$.04 & .61$\pm$.06 & .11$\pm$.03 & .189$\pm$.263 \\
\bottomrule
\end{tabular}
}
\caption{Performance of BFMs on the SleepEDF.}
\label{tab:app_SleepEDF}
\end{table}

\FloatBarrier
\begin{table}[!h]
\setlength{\tabcolsep}{2.2pt}
{\small
\begin{tabular}{llcccc}
\toprule
Type & Model & Acc & AUROC & MF1 & Kappa \\ \midrule
C & MBrain & .27$\pm$.02 & .54$\pm$.02 & .19$\pm$.03 & .027$\pm$.023 \\
 & SppEEGNet & .26$\pm$.01 & .51$\pm$.01 & .30$\pm$.02 & .115$\pm$.012 \\
 & BIOT & .26$\pm$.01 & .51$\pm$.01 & .24$\pm$.01 & .009$\pm$.012 \\
 & Bendr & .26$\pm$.01 & .51$\pm$.01 & .26$\pm$.01 & .013$\pm$.013 \\
\midrule
G & REVE & .34$\pm$.02 & \textbf{.66$\pm$.05} & .30$\pm$.02 & .115$\pm$.012 \\
 & NeuroGPT-E & \textbf{.35$\pm$.02} & .61$\pm$.04 & \textbf{.33$\pm$.01} & \textbf{.134$\pm$.032} \\
 & LaBraM & .29$\pm$.02 & .56$\pm$.04 & .22$\pm$.03 & .047$\pm$.021 \\
 & CodeBrain & .27$\pm$.01 & .55$\pm$.02 & .23$\pm$.04 & .032$\pm$.012 \\
 & BrainWave & .27$\pm$.02 & .54$\pm$.02 & .20$\pm$.03 & .031$\pm$.022 \\
 & BrainOmni & .27$\pm$.02 & .54$\pm$.01 & .22$\pm$.02 & .021$\pm$.029 \\
 & CBraMod & .27$\pm$.01 & .53$\pm$.01 & .22$\pm$.03 & .024$\pm$.020 \\
 & Brant & .25$\pm$.00 & .53$\pm$.01 & .10$\pm$.00 & .000$\pm$.000 \\
 & BFM & .26$\pm$.01 & .51$\pm$.01 & .25$\pm$.00 & .013$\pm$.012 \\
 & NeuroGPT-D & .26$\pm$.01 & .51$\pm$.03 & .15$\pm$.03 & .007$\pm$.019 \\
 & BrainBERT & .25$\pm$.01 & .50$\pm$.01 & .23$\pm$.02 & .002$\pm$.010 \\
\midrule
O & EEGPT & .28$\pm$.02 & .57$\pm$.03 & .22$\pm$.01 & .036$\pm$.024 \\
 & NeuroLM & .27$\pm$.01 & .52$\pm$.02 & .20$\pm$.02 & .030$\pm$.018 \\
 & MVPFormer & .27$\pm$.01 & .52$\pm$.01 & .25$\pm$.01 & .021$\pm$.011 \\
\bottomrule
\end{tabular}
}
\caption{Performance of BFMs on the BCI-2a.}
\label{tab:app_BCI_2a}
\end{table}

\FloatBarrier
\begin{table}[!h]
\setlength{\tabcolsep}{2.2pt}
{\small
\begin{tabular}{llcccc}
\toprule
Type & Model & Acc & AUROC & MF1 & Kappa \\ \midrule
C & Bendr & .29$\pm$.01 & .55$\pm$.01 & .29$\pm$.01 & .054$\pm$.010 \\
 & MBrain & .26$\pm$.01 & .51$\pm$.01 & .16$\pm$.05 & .009$\pm$.011 \\
 & SppEEGNet & .26$\pm$.01 & .51$\pm$.01 & .23$\pm$.01 & .008$\pm$.013 \\
 & BIOT & .25$\pm$.00 & .50$\pm$.01 & .20$\pm$.02 & -.005$\pm$.004 \\
\midrule
G & REVE & \textbf{.59$\pm$.02} & \textbf{.82$\pm$.01} & \textbf{.59$\pm$.02} & \textbf{.451$\pm$.031} \\
 & NeuroGPT-E & .54$\pm$.01 & .78$\pm$.01 & .54$\pm$.01 & .392$\pm$.019 \\
 & BrainOmni & .51$\pm$.01 & .76$\pm$.01 & .51$\pm$.01 & .351$\pm$.019 \\
 & LaBraM & .31$\pm$.02 & .59$\pm$.02 & .25$\pm$.03 & .078$\pm$.032 \\
 & CBraMod & .30$\pm$.02 & .58$\pm$.02 & .28$\pm$.04 & .073$\pm$.028 \\
 & CodeBrain & .30$\pm$.02 & .58$\pm$.02 & .28$\pm$.03 & .068$\pm$.020 \\
 & BFM & .27$\pm$.01 & .52$\pm$.01 & .26$\pm$.01 & .023$\pm$.009 \\
 & BrainWave & .25$\pm$.01 & .51$\pm$.01 & .18$\pm$.01 & .001$\pm$.014 \\
 & Brant & .25$\pm$.00 & .51$\pm$.01 & .10$\pm$.00 & .000$\pm$.000 \\
 & BrainBERT & .25$\pm$.00 & .49$\pm$.00 & .16$\pm$.02 & -.003$\pm$.005 \\
 & NeuroGPT-D & .25$\pm$.01 & .49$\pm$.01 & .16$\pm$.05 & -.007$\pm$.008 \\
\midrule
O & MVPFormer & .29$\pm$.02 & .56$\pm$.02 & .28$\pm$.01 & .060$\pm$.029 \\
 & EEGPT & .27$\pm$.02 & .52$\pm$.01 & .24$\pm$.04 & .023$\pm$.022 \\
 & NeuroLM & .25$\pm$.01 & .51$\pm$.01 & .15$\pm$.05 & .005$\pm$.008 \\
\bottomrule
\end{tabular}
}
\caption{Performance of BFMs on the EEGMMIDB-I.}
\label{tab:app_EEGMMIDB_I}
\end{table}

\FloatBarrier
\begin{table}[!h]
\setlength{\tabcolsep}{2.2pt}
{\small
\begin{tabular}{llcccc}
\toprule
Type & Model & Acc & AUROC & MF1 & Kappa \\ \midrule
C & Bendr & .29$\pm$.01 & .55$\pm$.01 & .29$\pm$.01 & .060$\pm$.014 \\
 & SppEEGNet & .26$\pm$.01 & .52$\pm$.00 & .24$\pm$.01 & .014$\pm$.008 \\
 & MBrain & .26$\pm$.01 & .52$\pm$.01 & .19$\pm$.05 & .008$\pm$.016 \\
 & BIOT & .25$\pm$.01 & .50$\pm$.01 & .21$\pm$.02 & .003$\pm$.011 \\
\midrule
G & REVE & \textbf{.59$\pm$.01} & \textbf{.82$\pm$.00} & \textbf{.58$\pm$.01} & \textbf{.448$\pm$.014} \\
 & NeuroGPT-E & .54$\pm$.02 & .77$\pm$.02 & .53$\pm$.02 & .381$\pm$.030 \\
 & BrainOmni & .49$\pm$.01 & .74$\pm$.01 & .49$\pm$.01 & .325$\pm$.014 \\
 & CBraMod & .30$\pm$.02 & .57$\pm$.01 & .27$\pm$.03 & .061$\pm$.022 \\
 & CodeBrain & .30$\pm$.02 & .57$\pm$.01 & .27$\pm$.02 & .067$\pm$.026 \\
 & LaBraM & .28$\pm$.02 & .55$\pm$.02 & .23$\pm$.03 & .041$\pm$.027 \\
 & BFM & .27$\pm$.01 & .52$\pm$.01 & .26$\pm$.01 & .024$\pm$.008 \\
 & NeuroGPT-D & .25$\pm$.00 & .50$\pm$.00 & .14$\pm$.01 & .003$\pm$.003 \\
 & Brant & .25$\pm$.00 & .50$\pm$.01 & .10$\pm$.00 & .000$\pm$.000 \\
 & BrainWave & .25$\pm$.01 & .50$\pm$.01 & .17$\pm$.02 & -.004$\pm$.011 \\
 & BrainBERT & .25$\pm$.00 & .50$\pm$.01 & .16$\pm$.03 & .001$\pm$.006 \\
\midrule
O & MVPFormer & .29$\pm$.01 & .54$\pm$.01 & .28$\pm$.01 & .047$\pm$.013 \\
 & NeuroLM & .26$\pm$.01 & .52$\pm$.02 & .14$\pm$.05 & .009$\pm$.017 \\
 & EEGPT & .25$\pm$.01 & .51$\pm$.01 & .23$\pm$.03 & .006$\pm$.017 \\
\bottomrule
\end{tabular}
}
\caption{Performance of BFMs on the EEGMMIDB-R.}
\label{tab:app_EEGMMIDB_R}
\end{table}

\FloatBarrier
\begin{table}[!h]
\setlength{\tabcolsep}{2.2pt}
{\small
\begin{tabular}{llcccc}
\toprule
Type & Model & Acc & AUROC & MF1 & Kappa \\ \midrule
C & SppEEGNet & .27$\pm$.06 & .51$\pm$.03 & .23$\pm$.03 & .007$\pm$.027 \\
 & MBrain & .42$\pm$.04 & .51$\pm$.05 & .15$\pm$.01 & .000$\pm$.000 \\
 & BIOT & .30$\pm$.06 & .50$\pm$.03 & .18$\pm$.02 & -.014$\pm$.030 \\
 & Bendr & .26$\pm$.02 & .50$\pm$.02 & \textbf{.24$\pm$.02} & -.004$\pm$.035 \\
\midrule
G & BrainBERT & .23$\pm$.02 & \textbf{.52$\pm$.02} & .18$\pm$.03 & .017$\pm$.020 \\
 & CodeBrain & .27$\pm$.12 & .52$\pm$.03 & .16$\pm$.06 & .004$\pm$.027 \\
 & NeuroGPT-D & \textbf{.44$\pm$.04} & .51$\pm$.03 & .15$\pm$.01 & .000$\pm$.000 \\
 & BrainWave & .27$\pm$.07 & .51$\pm$.03 & .23$\pm$.06 & .010$\pm$.077 \\
 & CBraMod & .19$\pm$.02 & .51$\pm$.01 & .14$\pm$.02 & -.001$\pm$.009 \\
 & LaBraM & .29$\pm$.04 & .51$\pm$.02 & .21$\pm$.03 & .016$\pm$.014 \\
 & REVE & .26$\pm$.05 & .50$\pm$.05 & .21$\pm$.03 & -.013$\pm$.035 \\
 & BFM & .28$\pm$.07 & .50$\pm$.02 & .22$\pm$.02 & .005$\pm$.015 \\
 & NeuroGPT-E & .42$\pm$.03 & .49$\pm$.01 & .17$\pm$.03 & -.004$\pm$.006 \\
 & Brant & .34$\pm$.17 & .48$\pm$.02 & .12$\pm$.06 & .000$\pm$.000 \\
 & BrainOmni & .22$\pm$.03 & .45$\pm$.03 & .19$\pm$.01 & -.046$\pm$.022 \\
\midrule
O & NeuroLM & .33$\pm$.13 & .51$\pm$.03 & .13$\pm$.03 & -.003$\pm$.007 \\
 & MVPFormer & .26$\pm$.05 & .50$\pm$.02 & .23$\pm$.04 & .011$\pm$.017 \\
 & EEGPT & .26$\pm$.05 & .50$\pm$.02 & .22$\pm$.04 & \textbf{.029$\pm$.027} \\
\bottomrule
\end{tabular}
}
\caption{Performance of BFMs on the DEAP.}
\label{tab:app_DEAP}
\end{table}

\FloatBarrier
\begin{table}[!h]
\setlength{\tabcolsep}{2.2pt}
{\small
\begin{tabular}{llcccc}
\toprule
Type & Model & Acc & AUROC & MF1 & Kappa \\ \midrule
C & MBrain & .26$\pm$.01 & .52$\pm$.01 & .21$\pm$.04 & .014$\pm$.020 \\
 & Bendr & .25$\pm$.00 & .50$\pm$.00 & .25$\pm$.00 & -.001$\pm$.005 \\
 & SppEEGNet & .25$\pm$.00 & .50$\pm$.00 & .24$\pm$.00 & -.001$\pm$.002 \\
 & BIOT & .24$\pm$.01 & .48$\pm$.00 & .19$\pm$.02 & -.018$\pm$.003 \\
\midrule
G & NeuroGPT-E & \textbf{.30$\pm$.01} & \textbf{.57$\pm$.01} & .26$\pm$.02 & .003$\pm$.005 \\
 & REVE & .28$\pm$.02 & .54$\pm$.01 & \textbf{.27$\pm$.01} & \textbf{.039$\pm$.019} \\
 & CodeBrain & .26$\pm$.01 & .53$\pm$.02 & .20$\pm$.03 & .020$\pm$.014 \\
 & BrainOmni & .28$\pm$.02 & .53$\pm$.01 & .26$\pm$.01 & .035$\pm$.019 \\
 & BrainWave & .27$\pm$.02 & .53$\pm$.02 & .22$\pm$.03 & .023$\pm$.026 \\
 & BFM & .27$\pm$.01 & .52$\pm$.01 & .24$\pm$.01 & .012$\pm$.004 \\
 & CBraMod & .26$\pm$.01 & .51$\pm$.01 & .21$\pm$.03 & .010$\pm$.010 \\
 & BrainBERT & .25$\pm$.02 & .51$\pm$.03 & .17$\pm$.05 & .003$\pm$.012 \\
 & LaBraM & .25$\pm$.01 & .51$\pm$.01 & .20$\pm$.02 & .008$\pm$.014 \\
 & NeuroGPT-D & .27$\pm$.00 & .50$\pm$.00 & .11$\pm$.00 & .000$\pm$.000 \\
 & Brant & .26$\pm$.01 & .50$\pm$.01 & .10$\pm$.00 & .000$\pm$.000 \\
\midrule
O & EEGPT & .25$\pm$.01 & .51$\pm$.01 & .22$\pm$.02 & .006$\pm$.016 \\
 & MVPFormer & .25$\pm$.01 & .51$\pm$.00 & .23$\pm$.02 & .005$\pm$.006 \\
 & NeuroLM & .25$\pm$.00 & .50$\pm$.00 & .13$\pm$.04 & .001$\pm$.002 \\
\bottomrule
\end{tabular}
}
\caption{Performance of BFMs on the SEED-IV.}
\label{tab:app_SEED_IV}
\end{table}

\FloatBarrier
\begin{table}[!h]
\setlength{\tabcolsep}{2.2pt}
{\small
\begin{tabular}{llcccc}
\toprule
Type & Model & Acc & AUROC & MF1 & Kappa \\ \midrule
C & MBrain & .03$\pm$.00 & .50$\pm$.01 & .00$\pm$.00 & .000$\pm$.001 \\
 & Bendr & .03$\pm$.00 & .50$\pm$.01 & \textbf{.02$\pm$.00} & .000$\pm$.002 \\
 & SppEEGNet & .02$\pm$.00 & .50$\pm$.00 & .01$\pm$.00 & -.000$\pm$.001 \\
 & BIOT & .03$\pm$.01 & .49$\pm$.00 & .01$\pm$.00 & .000$\pm$.001 \\
\midrule
G & LaBraM & .04$\pm$.01 & \textbf{.51$\pm$.01} & .01$\pm$.00 & \textbf{.005$\pm$.004} \\
 & CBraMod & .04$\pm$.01 & .51$\pm$.01 & .01$\pm$.00 & .003$\pm$.003 \\
 & CodeBrain & .05$\pm$.00 & .51$\pm$.01 & .00$\pm$.00 & .001$\pm$.001 \\
 & BrainOmni & .02$\pm$.00 & .50$\pm$.01 & .01$\pm$.00 & .001$\pm$.001 \\
 & BFM & .03$\pm$.00 & .50$\pm$.00 & .01$\pm$.00 & .001$\pm$.001 \\
 & NeuroGPT-D & .05$\pm$.00 & .50$\pm$.00 & .00$\pm$.00 & .001$\pm$.004 \\
 & BrainBERT & .03$\pm$.00 & .50$\pm$.01 & .01$\pm$.01 & .001$\pm$.003 \\
 & REVE & .03$\pm$.00 & .50$\pm$.00 & .02$\pm$.00 & .000$\pm$.001 \\
 & Brant & .04$\pm$.01 & .50$\pm$.00 & .00$\pm$.00 & .000$\pm$.000 \\
 & NeuroGPT-E & \textbf{.05$\pm$.00} & .50$\pm$.01 & .00$\pm$.00 & -.000$\pm$.004 \\
 & BrainWave & .03$\pm$.01 & .50$\pm$.01 & .00$\pm$.00 & .002$\pm$.003 \\
\midrule
O & EEGPT & .03$\pm$.01 & .50$\pm$.00 & .01$\pm$.00 & .001$\pm$.003 \\
 & NeuroLM & .04$\pm$.01 & .50$\pm$.00 & .00$\pm$.00 & -.001$\pm$.001 \\
 & MVPFormer & .02$\pm$.00 & .50$\pm$.01 & .02$\pm$.00 & .001$\pm$.002 \\
\bottomrule
\end{tabular}
}
\caption{Performance of BFMs on the Chisco-I.}
\label{tab:app_Chisco_I}
\end{table}

\FloatBarrier
\begin{table}[!h]
\setlength{\tabcolsep}{2.2pt}
{\small
\begin{tabular}{llcccc}
\toprule
Type & Model & Acc & AUROC & MF1 & Kappa \\ \midrule
C & BIOT & .04$\pm$.00 & .50$\pm$.01 & .01$\pm$.00 & -.000$\pm$.003 \\
 & MBrain & .04$\pm$.00 & .50$\pm$.00 & .00$\pm$.00 & .000$\pm$.000 \\
 & SppEEGNet & .02$\pm$.00 & .50$\pm$.00 & .01$\pm$.00 & .000$\pm$.003 \\
 & Bendr & .03$\pm$.00 & .50$\pm$.00 & \textbf{.02$\pm$.00} & .001$\pm$.003 \\
\midrule
G & BrainWave & .03$\pm$.01 & \textbf{.51$\pm$.02} & .00$\pm$.00 & .001$\pm$.003 \\
 & BrainOmni & .03$\pm$.01 & .51$\pm$.01 & .01$\pm$.00 & .002$\pm$.005 \\
 & LaBraM & .03$\pm$.01 & .51$\pm$.01 & .01$\pm$.00 & .002$\pm$.004 \\
 & NeuroGPT-D & \textbf{.05$\pm$.00} & .51$\pm$.01 & .00$\pm$.00 & .002$\pm$.003 \\
 & CBraMod & .03$\pm$.01 & .51$\pm$.01 & .01$\pm$.00 & -.000$\pm$.005 \\
 & NeuroGPT-E & .05$\pm$.00 & .50$\pm$.01 & .00$\pm$.00 & .001$\pm$.003 \\
 & BFM & .03$\pm$.01 & .50$\pm$.00 & .01$\pm$.00 & .001$\pm$.002 \\
 & REVE & .03$\pm$.00 & .50$\pm$.01 & .02$\pm$.00 & .000$\pm$.003 \\
 & Brant & .04$\pm$.01 & .50$\pm$.00 & .00$\pm$.00 & .000$\pm$.000 \\
 & BrainBERT & .03$\pm$.01 & .50$\pm$.00 & .00$\pm$.00 & .001$\pm$.001 \\
 & CodeBrain & .03$\pm$.01 & .50$\pm$.01 & .01$\pm$.00 & -.003$\pm$.001 \\
\midrule
O & EEGPT & .03$\pm$.01 & .50$\pm$.01 & .01$\pm$.01 & \textbf{.002$\pm$.002} \\
 & NeuroLM & .03$\pm$.02 & .50$\pm$.00 & .00$\pm$.00 & -.000$\pm$.000 \\
 & MVPFormer & .02$\pm$.00 & .50$\pm$.00 & .02$\pm$.00 & .001$\pm$.002 \\
\bottomrule
\end{tabular}
}
\caption{Performance of BFMs on the Chisco-R.}
\label{tab:app_Chisco_R}
\end{table}

\FloatBarrier
\begin{table}[!h]
\setlength{\tabcolsep}{2.2pt}
{\small
\begin{tabular}{llcccc}
\toprule
Type & Model & AUROC & Acc & F1 & F2 \\ \midrule
C & SppEEGNet & .51$\pm$.02 & .50$\pm$.01 & .44$\pm$.07 & .42$\pm$.09 \\
 & MBrain & .50$\pm$.01 & .50$\pm$.01 & \textbf{.63$\pm$.07} & \textbf{.76$\pm$.15} \\
 & BIOT & .50$\pm$.02 & .50$\pm$.01 & .34$\pm$.32 & .38$\pm$.39 \\
\midrule
G & LaBraM & \textbf{.53$\pm$.01} & .50$\pm$.00 & .53$\pm$.30 & .66$\pm$.37 \\
 & CodeBrain & .52$\pm$.01 & \textbf{.52$\pm$.01} & .29$\pm$.23 & .28$\pm$.32 \\
 & BrainBERT & .52$\pm$.01 & .50$\pm$.00 & .00$\pm$.00 & .00$\pm$.00 \\
 & CBraMod & .50$\pm$.01 & .50$\pm$.00 & .24$\pm$.28 & .25$\pm$.35 \\
 & BFM & .50$\pm$.01 & .50$\pm$.00 & .51$\pm$.05 & .52$\pm$.09 \\
 & BrainWave & .49$\pm$.01 & .50$\pm$.00 & .53$\pm$.30 & .67$\pm$.37 \\
\midrule
O & MVPFormer & .51$\pm$.00 & .49$\pm$.00 & .53$\pm$.05 & .57$\pm$.08 \\
\bottomrule
\end{tabular}
}
\caption{Performance of BFMs on the Cogitate.}
\label{tab:app_Cogitate}
\end{table}

\subsection{Frozen Encoder Fine-tuning}
\label{app:frozen}
\begin{figure*}[!ht]
  \centering
  \includegraphics[width = 0.95\textwidth]{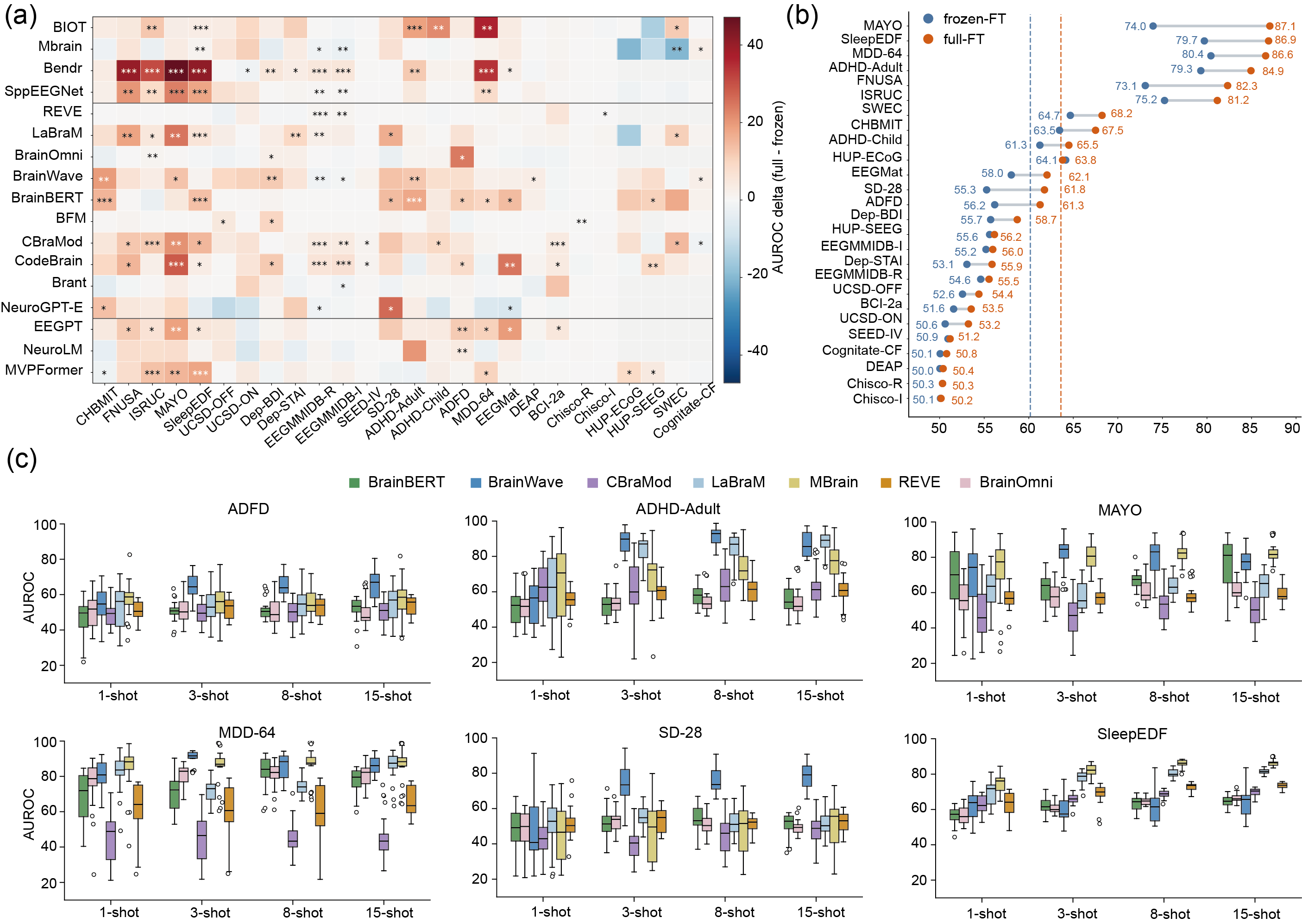}
  \caption{
   Effect of frozen-FT and full-FT.
    (a) Validation AUROC gain of full-FT over frozen-FT for each model-dataset pair, computed as average AUROC$_{\mathrm{full}}-$average AUROC$_{\mathrm{frozen}}$ (* $p<0.05$, ** $p<0.01$, and *** $p<0.001$).
    (b) Dataset-level average validation AUROC across all models under frozen-FT and full-FT.
    (c) Prototype-based few-shot evaluation.
}
  \label{fig:freeze}
\end{figure*}

In frozen-FT, the pretrained encoder is fixed and only the downstream classification head is optimized. For each model-dataset pair, we use the classifier-head learning rate selected in the corresponding full-FT setting, avoiding an arbitrary fixed head LR while keeping the two settings comparable. 
We additionally conduct a head-LR sensitivity analysis to examine whether the shared setting unfairly disadvantages frozen-FT. We evaluate representative model-dataset pairs with large full-FT gains, covering Bendr, EEGPT, SppEEGNet, and NeuroGPT on MDD-64, MAYO, ADHD-Adult, and ISRUC-G1. For most pairs, varying the head LR produces only minor changes in frozen-FT performance and therefore cannot explain the advantage of full-FT. EEGPT on MAYO is the only clearly LR-sensitive case. Nevertheless, the head LR used in the main experiment ($5\times10^{-4}$) yields a validation AUROC of 73.20, close to the best searched result of 75.13 at $10^{-3}$ and still far below the full-FT result of 91.76. These results indicate that the shared head-LR setting does not materially disadvantage frozen-FT or alter its performance ordering relative to full-FT.
Figure~\ref{fig:freeze}(a) reports the difference in mean AUROC between full-FT and frozen-FT for each model-dataset pair, with significance indicated by asterisks.

\subsection{Additional Prototype-based Transfer Analysis}
\label{app:proto}
Figure~\ref{fig:freeze}(c) provides the detailed results underlying the pair-level heterogeneity summarized in the main text, showing the AUROC distributions of seven models on six datasets under the 1-, 3-, 8-, and 15-shot settings, with each box summarizing all fold-level results from five repeated support-set samplings per fold.
To assess whether the global shot-performance trend is consistent across model-dataset combinations, we conducted pair-level analyses. The results show clear pair-level heterogeneity. Among the 42 model-dataset pairs with all four shot settings available, 22 exhibit a significant positive 1/3/8/15-shot trend after FDR correction; for the 15-shot versus 1-shot comparison, 21 of 42 pairs show a significant improvement for 15-shot. Thus, although the aggregate results indicate a positive shot-performance relationship, statistically reliable gains are limited to a subset of model-dataset pairs. This heterogeneity is also visible in the pair-level patterns. Several models on SleepEDF show significant positive trends together with relatively large 15-shot gains, while BrainWave shows particularly large improvements on ADHD-Adult, SZ-28, and ADFD. In contrast, other pairs show only modest or non-significant gains, including ADFD/BrainOmni, ADFD/MBrain, SZ-28/REVE, ADHD-Adult/CBraMod, and MDD-64/CBraMod.
\subsection{Cross-modal representation geometry}
This section provides the full protocol for the cross-modal representation analysis in \textbf{EEG and iEEG modalities}. The analysis examines whether pretrained BFMs encode the two modalities in shared principal subspaces, whether this sharing is directionally asymmetric, and whether modality-specific residual structure is associated with frozen-transfer performance. The iEEG pool was fixed across replicates and comprised HUP-ECoG, HUP-SEEG, MAYO, FNUSA, SWEC, and Cogitate-CF. We used three EEG-cohort draws: ADHD-Adult, ADHD-Child, BCI-2a, CHBMIT, EEGMMIDB-I, and ISRUC-G1 (seed 2026); BCI-2a, CHBMIT, Chisco-R, DEAP, EEGMMIDB-R, and MDD-64 (seed 2030); and ADFD, Chisco-I, DEAP, EEGMMIDB-I, ISRUC, and SleepEDF (seed 2089).

For each model-dataset pair, we load the original pretrained checkpoint, remove the prediction head, and extract sample-level representations from matched 2-s epochs. Each cohort contributes 2,000 label-balanced samples. Representations are centered separately within each cohort before pooling, reducing cohort-level mean shifts associated with task and acquisition differences. The EEG and iEEG pools therefore contain 12,000 samples each. For the mixed pool, we sample 1,000 epochs from each of the 12 cohorts, yielding the same total sample size.

\begin{table*}[t]
\centering
\caption{Cross-modal geometry of pretrained representations after dataset-wise
centering. $C_{\mathrm{E}\rightarrow\mathrm{I}}$ denotes the fraction of iEEG
variance explained by the EEG $99\%$ PCA subspace, and
$C_{\mathrm{I}\rightarrow\mathrm{E}}$ is defined symmetrically.}
\label{tab:cross_modal_geometry}
\small
\begin{tabular}{lrrrrrr}
\toprule
Model &
$D_{99}^{\mathrm{EEG}}$ &
$D_{99}^{\mathrm{iEEG}}$ &
$D_{99}^{\mathrm{mix}}$ &
$C_{\mathrm{E}\rightarrow\mathrm{I}}$ &
$C_{\mathrm{I}\rightarrow\mathrm{E}}$ &
$\bar C$ \\
\midrule
BrainWave  & $22.0 \pm 5.2$   & $11.0 \pm 0.0$   & $22.0 \pm 2.6$   & $0.940 \pm 0.033$ & $0.866 \pm 0.007$ & $0.903 \pm 0.017$ \\
BrainBERT  & $184.0 \pm 5.6$  & $173.0 \pm 0.0$  & $186.0 \pm 2.6$  & $0.986 \pm 0.001$ & $0.981 \pm 0.006$ & $0.984 \pm 0.003$ \\
BIOT       & $49.7 \pm 5.5$   & $47.0 \pm 0.0$   & $58.7 \pm 3.1$   & $0.970 \pm 0.005$ & $0.969 \pm 0.003$ & $0.970 \pm 0.002$ \\
LaBraM     & $128.0 \pm 4.0$  & $117.0 \pm 0.0$  & $130.3 \pm 1.5$  & $0.976 \pm 0.004$ & $0.969 \pm 0.001$ & $0.973 \pm 0.002$ \\
CBraMod    & $64.7 \pm 11.2$  & $83.0 \pm 0.0$   & $80.0 \pm 6.0$   & $0.969 \pm 0.006$ & $0.983 \pm 0.005$ & $0.976 \pm 0.002$ \\
SppEEGNet  & $37.7 \pm 0.6$   & $37.0 \pm 0.0$   & $38.0 \pm 0.0$   & $0.987 \pm 0.002$ & $0.985 \pm 0.001$ & $0.986 \pm 0.001$ \\
Mbrain     & $351.3 \pm 38.7$ & $495.0 \pm 0.0$  & $508.3 \pm 6.7$  & $0.805 \pm 0.041$ & $0.885 \pm 0.017$ & $0.845 \pm 0.028$ \\
CodeBrain  & $93.7 \pm 3.8$   & $93.0 \pm 0.0$   & $95.3 \pm 1.5$   & $0.983 \pm 0.004$ & $0.985 \pm 0.001$ & $0.984 \pm 0.001$ \\
MVPFormer  & $452.3 \pm 35.5$ & $316.0 \pm 0.0$  & $414.0 \pm 21.5$ & $0.987 \pm 0.006$ & $0.952 \pm 0.011$ & $0.970 \pm 0.008$ \\
BFM        & $129.0 \pm 30.5$ & $94.0 \pm 0.0$   & $126.0 \pm 14.0$ & $0.979 \pm 0.005$ & $0.928 \pm 0.042$ & $0.953 \pm 0.022$ \\
\bottomrule
\end{tabular}
\end{table*}

Let $\mathbf{X}_M\in\mathbb{R}^{n_M\times d}$ denote the pooled centered representations for
$M\in\{\mathrm{EEG},\mathrm{iEEG},\mathrm{mix}\}$, where $d$ is the hidden dimension. We compute the covariance matrix
\begin{equation}
\boldsymbol{\Sigma}_M
=
\frac{1}{n_M-1}\mathbf{X}_M^\top\mathbf{X}_M
\end{equation}
and its eigendecomposition
\begin{equation}
\boldsymbol{\Sigma}_M
=
\mathbf{V}_M\boldsymbol{\Lambda}_M\mathbf{V}_M^\top,
\end{equation}
with eigenvalues $\lambda_{M,1}\geq\cdots\geq\lambda_{M,D}$. For
$\alpha=0.99$, we define the effective PCA dimensionality as
\begin{equation}
D_{\alpha}^{M}
=
\min\left\{
k:
\frac{\sum_{i=1}^{k}\lambda_{M,i}}
{\sum_{i=1}^{d}\lambda_{M,i}}
\geq \alpha
\right\}.
\end{equation}
Let
$\mathbf{U}_M\in\mathbb{R}^{d\times D_{99}^{M}}$ denote the leading eigenvectors spanning the corresponding $99\%$-variance subspace.

To obtain an out-of-sample estimate of subspace coverage, we split each pooled modality matrix into complementary interleaved halves,
$\mathbf{X}_M^{(a)}$ and $\mathbf{X}_M^{(b)}$. PCA subspaces are fitted independently on each half. For source modality $S$ and target modality $T$, directional coverage is evaluated on the opposite target half and averaged over the two split directions:
\begin{equation}
C_{S\rightarrow T}
=
\frac{1}{2}
\Bigg(
\frac{
\left\|
\mathbf{X}_T^{(b)}\mathbf{U}_S^{(a)}
\right\|_F^2
}{
\left\|
\mathbf{X}_T^{(b)}
\right\|_F^2
}+
\frac{
\left\|
\mathbf{X}_T^{(a)}\mathbf{U}_S^{(b)}
\right\|_F^2
}{
\left\|
\mathbf{X}_T^{(a)}
\right\|_F^2
}
\Bigg).
\end{equation}
Here, $\mathbf{U}_S^{(h)}$ is the orthonormal $99\%$-variance PCA basis fitted on source split $h$. Thus, $C_{S\rightarrow T}$ measures the fraction of target variance retained after projection onto the source subspace. We compute both
$C_{\mathrm{EEG}\rightarrow\mathrm{iEEG}}$ and
$C_{\mathrm{iEEG}\rightarrow\mathrm{EEG}}$ to characterize directional asymmetry.

Within-modality coverage is computed identically using complementary halves of the same modality:
\begin{equation}
C_{M\rightarrow M}
=
\frac{1}{2}
\left[
\frac{
\left\|
\mathbf{X}_M^{(b)}\mathbf{U}_M^{(a)}
\right\|_F^2
}{
\left\|
\mathbf{X}_M^{(b)}
\right\|_F^2
}
+
\frac{
\left\|
\mathbf{X}_M^{(a)}\mathbf{U}_M^{(b)}
\right\|_F^2
}{
\left\|
\mathbf{X}_M^{(a)}
\right\|_F^2
}
\right].
\end{equation}
We define the modality-specific residual gap as the mean loss in coverage when replacing a within-modality subspace with the opposite-modality subspace:
\begin{equation}
\begin{aligned}
G
&=
\frac{1}{2}
\Big[
\big(
C_{\mathrm{EEG}\rightarrow\mathrm{EEG}}
-
C_{\mathrm{iEEG}\rightarrow\mathrm{EEG}}
\big)
\\
&\quad+
\big(
C_{\mathrm{iEEG}\rightarrow\mathrm{iEEG}}
-
C_{\mathrm{EEG}\rightarrow\mathrm{iEEG}}
\big)
\Big].
\end{aligned}
\end{equation}
Larger $G$ indicates that within-modality subspaces retain substantially more variance than cross-modal subspaces, corresponding to stronger modality-specific residual structure. We report the mean and standard deviation across the three EEG-cohort draws and relate $G$ to macro-averaged frozen-transfer AUROC across the cohorts included in each draw.

\subsection{Additional Spatial Analysis Results}
\label{app:exp4}
This subsection provides detailed results supporting the analysis discussed in \textbf{Spatial Sensitivity}. We report the full quantitative impact of channel-permutation perturbations across datasets and models. We permute channels in the training and validation splits to probe whether BFMs encode dataset-specific spatial structure. For models using channel identifiers, signal channels and their corresponding identifiers are jointly permuted to preserve signal–identifier correspondence. All other settings are kept identical, so performance changes reflect sensitivity to the disrupted spatial organization, including channel-wise heterogeneity and, for order-dependent models, induced topology. \Cref{tab:exp4-1,tab:exp4-2,tab:exp4-3,tab:exp4-4,tab:exp4-5} report all metrics before and after permutation. We focus on AUROC in the main analysis, and these results support that some BFMs internalize dataset-specific spatial structure during training.
\begin{table}[!h]
\setlength{\tabcolsep}{1.8pt}
\resizebox{1.05\columnwidth}{!}{
\begin{tabular}{lcccccccc}
\toprule
           & \multicolumn{2}{c}{AUROC} & \multicolumn{2}{c}{Acc} & \multicolumn{2}{c}{F1} & \multicolumn{2}{c}{F2} \\ \midrule
Model      & shuffle     & origin      & shuffle     & origin     & shuffle     & origin      & shuffle    & origin    \\ \midrule
BrainBERT  &  .54±.04 &  .55±.03 &  .55±.03 &  .56±.03 &  .58±.04 &  .56±.07 &  .58±.05 &  .53±.09 \\
BrainWave  &  .68±.04 &  .78±.04 &  .66±.05 &  .71±.04 &  .68±.12 &  .74±.06 &  .68±.19 &  .74±.12 \\
BrainOmni  &  .71±.09 &  .71±.09 &  .62±.07 &  .62±.07 &  .63±.09 &  .63±.09 &  .61±.15 &  .61±.14 \\
CBraMod    &  .60±.06 &  .64±.05 &  .57±.03 &  .64±.06 &  .72±.03 &  .73±.03 &  .85±.03 &  .81±.05 \\
EEGPT      &  .63±.11 &  .67±.13 &  .60±.10 &  .64±.10 &  .57±.28 &  .68±.10 &  .59±.32 &  .68±.12 \\
MBrain     &  .69±.11 &  .69±.11 &  .63±.09 &  .63±.08 &  .68±.07 &  .66±.06 &  .70±.05 &  .66±.04 \\
NeuroGPT-E &  .55±.02 &  .55±.02 &  .56±.03 &  .56±.03 &  .72±.02 &  .72±.02 &  .86±.01 &  .86±.01 \\
REVE       &  .71±.09 &  .79±.07 &  .55±.11 &  .71±.04 &  .36±.31 &  .74±.06 &  .32±.30 &  .74±.08 \\ \bottomrule
\end{tabular}
}
\caption{Performance under channel-permutation setting. We report AUROC, Accuracy, F1 and F2 for models trained on original data and shuffled data on ADHD-Child.}
\label{tab:exp4-1}
\end{table}

\begin{table}[h]
\setlength{\tabcolsep}{1.8pt}
\resizebox{1.05\columnwidth}{!}{
\begin{tabular}{lcccccccc}
\toprule
           & \multicolumn{2}{c}{AUROC} & \multicolumn{2}{c}{Acc} & \multicolumn{2}{c}{F1} & \multicolumn{2}{c}{F2} \\ \midrule
Model      & shuffle     & origin      & shuffle     & origin     & shuffle     & origin      & shuffle    & origin    \\ \midrule
BrainBERT  &  .62±.05 &  .59±.05 &  .64±.01 &  .58±.05 &  .66±.02 &  .60±.11 &  .64±.04 &  .64±.15 \\
BrainWave  &  .62±.12 &  .74±.05 &  .57±.09 &  .68±.05 &  .49±.26 &  .68±.10 &  .47±.31 &  .73±.10 \\
BrainOmni  &  .80±.07 &  .80±.07 &  .71±.04 &  .71±.04 &  .73±.05 &  .73±.05 &  .72±.09 &  .72±.09 \\
CBraMod    &  .51±.02 &  .55±.06 &  .52±.07 &  .59±.04 &  .32±.26 &  .41±.20 &  .28±.25 &  .37±.20 \\
EEGPT      &  .79±.04 &  .81±.05 &  .72±.04 &  .72±.05 &  .72±.05 &  .73±.05 &  .68±.07 &  .70±.07 \\
MBrain     &  .52±.13 &  .53±.12 &  .51±.09 &  .50±.10 &  .54±.12 &  .53±.12 &  .54±.15 &  .52±.14 \\
NeuroGPT-E &  .49±.05 &  .51±.03 &  .55±.01 &  .55±.01 &  .71±.01 &  .71±.01 &  .85±.01 &  .85±.01 \\
REVE       &  .79±.08 &  .84±.07 &  .68±.07 &  .77±.08 &  .74±.05 &  .79±.07 &  .79±.07 &  .79±.09 \\ \bottomrule
\end{tabular}
}
\caption{Channel-permutation performance on ADFD.}
\label{tab:exp4-2}
\end{table}

\begin{table}[h]
\setlength{\tabcolsep}{1.8pt}
\resizebox{1.05\columnwidth}{!}{
\begin{tabular}{lcccccccc}
\toprule
           & \multicolumn{2}{c}{AUROC} & \multicolumn{2}{c}{Acc} & \multicolumn{2}{c}{F1} & \multicolumn{2}{c}{F2} \\ \midrule
Model      & shuffle      & origin     & shuffle     & origin     & shuffle     & origin      & shuffle    & origin    \\ \midrule
BrainBERT  &  .79±.04 &  .78±.06 &  .80±.04 &  .75±.10 &  .59±.08 &  .58±.08 &  .59±.11 &  .61±.09 \\
BrainWave  &  .59±.14 &  .90±.03 &  .63±.09 &  .80±.12 &  .30±.11 &  .71±.09 &  .32±.15 &  .78±.04 \\
BrainOmni  &  .74±.13 &  .74±.13 &  .67±.19 &  .67±.19 &  .51±.12 &  .51±.12 &  .56±.12 &  .56±.12 \\
CBraMod    &  .67±.12 &  .69±.04 &  .67±.11 &  .70±.08 &  .42±.16 &  .46±.08 &  .48±.23 &  .49±.14 \\
EEGPT      &  .69±.13 &  .71±.09 &  .73±.10 &  .67±.11 &  .47±.16 &  .46±.06 &  .48±.19 &  .51±.15 \\
MBrain     &  .70±.08 &  .75±.08 &  .77±.03 &  .70±.11 &  .22±.22 &  .54±.10 &  .17±.18 &  .59±.09 \\
NeuroGPT-E &  .65±.07 &  .78±.12 &  .70±.06 &  .75±.07 &  .26±.10 &  .51±.22 &  .24±.14 &  .63±.16 \\
REVE       &  .78±.12 &  .84±.10 &  .59±.25 &  .78±.12 &  .53±.18 &  .63±.14 &  .65±.12 &  .66±.13 \\ \bottomrule
\end{tabular}
}
\caption{Channel-permutation performance on CHBMIT.}
\label{tab:exp4-3}
\end{table}

\begin{table}[!h]
\setlength{\tabcolsep}{1.8pt}
\resizebox{1.05\columnwidth}{!}{
\begin{tabular}{lcccccccc}
\toprule
    & \multicolumn{2}{c}{AUROC} & \multicolumn{2}{c}{Acc} & \multicolumn{2}{c}{F1} & \multicolumn{2}{c}{F2} \\ \midrule
Model      & shuffle     & origin      & shuffle     & origin     & shuffle      & origin      & shuffle    & origin    \\ \midrule
BrainBERT  &  .50±.09 &  .51±.05 &  .49±.05 &  .51±.06 &  .57±.04 &  .51±.11 &  .64±.06 &  .54±.17 \\
BrainWave  &  .50±.11 &  .69±.10 &  .49±.12 &  .53±.08 &  .40±.28 &  .60±.06 &  .43±.31 &  .69±.14 \\
BrainOmni  &  .51±.19 &  .64±.09 &  .46±.15 &  .57±.06 &  .44±.21 &  .54±.16 &  .46±.26 &  .57±.26 \\
CBraMod    &  .59±.08 &  .60±.12 &  .51±.07 &  .49±.08 &  .21±.29 &  .18±.08 &  .22±.35 &  .13±.07 \\
EEGPT      &  .63±.17 &  .51±.21 &  .55±.14 &  .48±.17 &  .49±.16 &  .49±.15 &  .46±.16 &  .49±.14 \\
MBrain     &  .43±.16 &  .49±.15 &  .46±.06 &  .46±.14 &  .49±.21 &  .37±.17 &  .56±.28 &  .34±.18 \\
NeuroGPT-E &  .55±.25 &  .56±.10 &  .53±.10 &  .53±.07 &  .51±.25 &  .58±.10 &  .57±.34 &  .64±.13 \\
REVE       &  .58±.12 &  .63±.19 &  .58±.06 &  .56±.13 &  .53±.17 &  .45±.29 &  .53±.25 &  .46±.33 \\ \bottomrule
\end{tabular}
}
\caption{Channel-permutation setting performance on UCSD-ON.}
\label{tab:exp4-4}
\end{table}

\begin{table}[!h]
\setlength{\tabcolsep}{1.8pt}
\resizebox{1.05\columnwidth}{!}{
\begin{tabular}{lcccccccc}
\toprule
           & \multicolumn{2}{c}{AUROC} & \multicolumn{2}{c}{Acc} & \multicolumn{2}{c}{F1} & \multicolumn{2}{c}{F2} \\ \midrule
Model      & shuffle     & origin      & shuffle     & origin     & shuffle      & origin      & shuffle    & origin    \\ \midrule
BrainBERT  &  .52±.09 &  .51±.06 &  .53±.06 &  .54±.06 &  .60±.10 &  .49±.17 &  .68±.15 &  .49±.20 \\
BrainWave  &  .49±.19 &  .58±.06 &  .56±.10 &  .53±.10 &  .43±.29 &  .62±.12 &  .44±.34 &  .71±.19 \\
BrainOmni  &  .62±.07 &  .64±.09 &  .56±.04 &  .57±.06 &  .57±.10 &  .54±.16 &  .61±.21 &  .57±.26 \\
CBraMod    &  .60±.05 &  .63±.05 &  .56±.04 &  .58±.04 &  .32±.17 &  .46±.09 &  .27±.19 &  .40±.12 \\
EEGPT      &  .62±.15 &  .61±.13 &  .60±.12 &  .55±.08 &  .55±.17 &  .56±.10 &  .53±.20 &  .58±.17 \\
MBrain     &  .48±.09 &  .50±.04 &  .44±.06 &  .44±.04 &  .35±.22 &  .32±.23 &  .38±.31 &  .34±.29 \\
NeuroGPT-E &  .56±.07 &  .56±.06 &  .52±.05 &  .54±.05 &  .57±.09 &  .57±.12 &  .63±.17 &  .62±.22 \\
REVE       &  .63±.11 &  .62±.19 &  .59±.13 &  .58±.17 &  .61±.13 &  .58±.19 &  .63±.16 &  .60±.24 \\ \bottomrule
\end{tabular}
}
\caption{Channel-permutation setting performance on UCSD-OFF.}
\label{tab:exp4-5}
\end{table}

\subsection{Additional Frequency Analysis Results}
\label{app:exp3}
This subsection provides detailed results supporting the frequency-band analysis described in \textbf{Frequency-Domain Associations}. Our goal is not to isolate spectral information acquired during pretraining, but to characterize the frequency structure retained after downstream adaptation. We report the full quantitative outcomes of the band-wise PSD prediction experiments. \Cref{tab:exp3-1,tab:exp3-2,tab:exp3-3} report the band-wise prediction performance for all evaluated models across the six canonical frequency bands. For each band $b$, we list the Pearson correlation between the ground-truth PSD and the predicted PSD, as well as the normalized relative strength $r_b^{n}$ used in the main analysis.
\FloatBarrier
\begin{table}[!h]
\setlength{\tabcolsep}{3pt}
\resizebox{1\columnwidth}{!}{
\begin{tabular}{lcccccclcccccc}
\toprule
           & \multicolumn{6}{c}{ADFD}                          &  & \multicolumn{6}{c}{Dep-BDI}                       \\ \cmidrule{1-7} \cmidrule{9-14} 
Model      & $r_d^n$ & $r_t^n$ & $r_a^n$ & $r_b^n$ & $r_{gl}^n$ & $r_{gh}^n$ &  & $r_d^n$ & $r_t^n$ & $r_a^n$ & $r_b^n$ & $r_{gl}^n$ & $r_{gh}^n$ \\ \cmidrule{1-7} \cmidrule{9-14} 
REVE       & .85  & .00  & .71  & .11  & .59       & 1.0       &  & .99  & .00  & 1.0  & .92  & .78       & .65       \\
EEGPT      & 1.0  & .32  & .36  & .20  & .92       & .00       &  & .80  & .67  & 1.0  & .67  & .13       & .00       \\
BrainOmni  & .95  & .00  & .82  & .02  & .28       & 1.0       &  & .83  & .39  & 1.0  & .81  & .35       & .00       \\
BrainWave  & .63  & .35  & .77  & .00  & .14       & 1.0       &  & 1.0  & .47  & .75  & .23  & .00       & .28       \\
LaBraM     & .08  & .00  & 1.0  & .26  & .28       & .10       &  & .77  & .36  & 1.0  & .55  & .15       & .00       \\
BrainBERT  & .69  & .08  & .16  & .00  & .68       & 1.0       &  & .70  & .44  & 1.0  & .00  & .27       & .87       \\
BFM        & .89  & .00  & .33  & .09  & .27       & 1.0       &  & .89  & 1.0  & .99  & .76  & 1.0       & .71       \\
BIOT       & .87  & .00  & .62  & .31  & .37       & 1.0       &  & .80  & .38  & 1.0  & .75  & .17       & .00       \\
CBraMod    & .84  & .00  & .67  & .55  & .64       & 1.0       &  & .91  & .00  & 1.0  & .86  & .47       & .05       \\
MBrain     & .87  & .00  & .42  & .15  & .40       & 1.0       &  & .98  & .00  & 1.0  & .87  & .87       & .61       \\
SppEEGNet  & .26  & .10  & .00  & .02  & .27       & 1.0       &  & 1.0  & .55  & .94  & .27  & .00       & .08       \\
NeuroLM    & .66  & .89  & .51  & .08  & .00       & 1.0       &  & 1.0  & .36  & .35  & .00  & .20       & .05       \\
NeuroGPT-E & .85  & .13  & .00  & .01  & .06       & 1.0       &  & .52  & .39  & 1.0  & .00  & .06       & .10       \\
Bendr      & .74  & .28  & .00  & .10  & .07       & 1.0       &  & 1.0  & .28  & .79  & .44  & .00       & .03       \\
Brant      & .53  & .33  & .17  & .08  & .00       & 1.0       &  & .00  & .68  & .33  & .43  & .93       & 1.0       \\
NeuroGPT-D & .95  & .23  & .25  & .10  & .00       & 1.0       &  & 1.0  & .30  & .86  & .62  & .11       & .00       \\ \bottomrule
\end{tabular}
}
\caption{Band-wise PSD predictability across BFMs on ADFD and Dep-BDI, reporting normalized band predictability $r_b^n$ for delta (d), theta (t), alpha (a), beta (b), low gamma (gl), and high gamma (gh).}
\label{tab:exp3-1}
\end{table}

\begin{table}[!h]
\setlength{\tabcolsep}{2pt}
{\small
\begin{tabular}{lcccccclcccccc}
\toprule
           & \multicolumn{6}{c}{SZ-28}                         &  & \multicolumn{6}{c}{MDD-64}                        \\ \cmidrule{1-7} \cmidrule{9-14} 
Model      & $r_d^n$ & $r_t^n$ & $r_a^n$ & $r_b^n$ & $r_{gl}^n$ & $r_{gh}^n$ &  & $r_d^n$ & $r_t^n$ & $r_a^n$ & $r_b^n$ & $r_{gl}^n$ & $r_{gh}^n$ \\ \cmidrule{1-7} \cmidrule{9-14} 
BrainWave  & .95  & .64  & 1.0  & .75  & .00       & .22       &  & .89  & 1.0  & .89  & .61  & .00       & .02       \\
REVE       & 1.0  & .00  & .97  & .66  & .67       & .22       &  & .96  & 1.0  & .99  & 1.0  & .93       & .00       \\
NeuroGPT-E & 1.0  & .87  & .06  & .00  & .41       & .90       &  & .25  & 1.0  & .00  & .62  & .60       & .08       \\
BrainBERT  & .82  & .15  & 1.0  & .00  & .13       & .29       &  & .75  & 1.0  & .73  & .70  & .80       & .00       \\
BrainOmni  & .86  & .46  & 1.0  & .71  & .27       & .00       &  & .66  & .84  & 1.0  & .99  & .74       & .00       \\
BFM        & .92  & .00  & 1.0  & .84  & .90       & .84       &  & .86  & 1.0  & .97  & .97  & .97       & .00       \\
MBrain     & 1.0  & .00  & 1.0  & .94  & .83       & .69       &  & .94  & .96  & .96  & 1.0  & .92       & .00       \\
EEGPT      & .91  & .66  & 1.0  & .97  & .38       & .00       &  & .89  & .97  & 1.0  & .89  & .74       & .00       \\
BIOT       & .68  & .00  & 1.0  & .58  & .65       & .19       &  & .93  & .97  & 1.0  & .99  & .86       & .00       \\
NeuroGPT-D & 1.0  & .00  & .76  & .67  & .36       & .14       &  & .92  & 1.0  & .98  & .71  & .67       & .00       \\
CBraMod    & 1.0  & .25  & .90  & .89  & .58       & .00       &  & .95  & .96  & .98  & 1.0  & .90       & .00       \\
LaBraM     & .87  & .44  & 1.0  & .70  & .34       & .00       &  & .59  & .86  & .59  & 1.0  & .76       & .00       \\
Brant      & .64  & .00  & 1.0  & .71  & .61       & .54       &  & .84  & .87  & .89  & 1.0  & .93       & .00       \\
Bendr      & 1.0  & .34  & .53  & .51  & .01       & .00       &  & .73  & 1.0  & .48  & .49  & .46       & .00       \\
SppEEGNet  & 1.0  & .16  & .68  & .29  & .00       & .02       &  & 1.0  & .84  & .78  & .63  & .51       & .00       \\
NeuroLM    & 1.0  & .31  & .81  & .31  & .00       & .24       &  & .59  & .86  & .59  & 1.0  & .76       & .00       \\ \bottomrule
\end{tabular}
}
\caption{Band-wise PSD predictability on SZ-28 and MDD-64.}
\label{tab:exp3-2}
\end{table}

\begin{table}[!h]
\setlength{\tabcolsep}{2pt}
{\small
\begin{tabular}{lcccccclcccccc}
\toprule
           & \multicolumn{6}{c}{ADHD-Adult}                    &  & \multicolumn{6}{c}{ADHD-Child}                    \\ \cmidrule{1-7} \cmidrule{9-14} 
Model      & $r_d^n$ & $r_t^n$ & $r_a^n$ & $r_b^n$ & $r_{gl}^n$ & $r_{gh}^n$ &  & $r_d^n$ & $r_t^n$ & $r_a^n$ & $r_b^n$ & $r_{gl}^n$ & $r_{gh}^n$ \\ \cmidrule{1-7} \cmidrule{9-14} 
EEGPT      & .66  & .00  & .59  & 1.0  & .18       & .08       &  & .95  & .82  & 1.0  & 1.0  & .56       & .00       \\
BrainOmni  & .65  & .00  & .67  & 1.0  & .32       & .14       &  & 1.0  & .86  & .99  & .84  & .39       & .00       \\
BrainWave  & .32  & .00  & .68  & 1.0  & .80       & .82       &  & .83  & .59  & .94  & 1.0  & .34       & .00       \\
BIOT       & .40  & .00  & .40  & 1.0  & .58       & .17       &  & .98  & .89  & 1.0  & .78  & .32       & .00       \\
REVE       & .71  & .00  & .25  & 1.0  & .53       & .30       &  & .96  & .82  & .98  & 1.0  & .62       & .00       \\
MBrain     & .66  & .00  & .55  & 1.0  & .53       & .16       &  & 1.0  & .87  & .98  & 1.0  & .60       & .00       \\
Brant      & .71  & .00  & .33  & 1.0  & .32       & .08       &  & .89  & .58  & .68  & 1.0  & .59       & .00       \\
CBraMod    & .65  & .19  & .61  & 1.0  & .52       & .00       &  & .99  & .86  & .94  & 1.0  & .79       & .00       \\
LaBraM     & .76  & .39  & .71  & 1.0  & .08       & .00       &  & 1.0  & .86  & .96  & .92  & .41       & .00       \\
NeuroGPT-E & .39  & .74  & .00  & 1.0  & .40       & .15       &  & .84  & 1.0  & .75  & .37  & .00       & .44       \\
NeuroGPT-D & .95  & .00  & .71  & 1.0  & .16       & .03       &  & 1.0  & .70  & .83  & .77  & .34       & .00       \\
BFM        & .61  & .00  & .45  & 1.0  & .70       & .62       &  & .96  & .83  & .94  & 1.0  & .67       & .00       \\
NeuroLM    & .41  & .00  & .55  & 1.0  & .29       & .19       &  & 1.0  & .82  & .97  & .88  & .27       & .00       \\
BrainBERT  & 1.0  & .26  & .91  & .84  & .13       & .00       &  & 1.0  & .64  & .82  & .66  & .60       & .00       \\
SppEEGNet  & 1.0  & .06  & .72  & .61  & .06       & .00       &  & 1.0  & .80  & .70  & .64  & .31       & .00       \\
Bendr      & 1.0  & .16  & .73  & .78  & .03       & .00       &  & 1.0  & .62  & .83  & .57  & .20       & .00       \\ \bottomrule
\end{tabular}
}
\caption{Band-wise PSD predictability across BFMs on ADHD.}
\label{tab:exp3-3}
\end{table}

\subsection{Codebook Analysis Results}
\label{app:exp5}
This subsection provides the complete results supporting the \textbf{Codebook Representation} in the main text. We first report codebook token usage and class geometry across all evaluated settings. We then present a detailed analysis of LaBraM on ADFD, CHBMIT, SZ-28, and SleepEDF, comparing standard fine-tuning without the codebook with codebook-enabled fine-tuning under identical protocols in terms of AUROC, accuracy, F1, and F2.

\begin{table}[]
\renewcommand{\arraystretch}{1}
\setlength{\tabcolsep}{1.8pt}
\centering
\resizebox{1.02\columnwidth}{!}{
\begin{tabular}{lcccccccccccc}
\toprule
              & \multicolumn{4}{c}{ADFD}    & \multicolumn{4}{c}{CHBMIT}  & \multicolumn{4}{c}{SZ-28}  \\ \midrule
Model         & Cov. & Ent & Inter & DR    & Cov. & Ent & Inter & DR    & Cov. & Ent & Inter & DR   \\ \midrule
BFM           & .430 & .868 & .998  & .002  & .635 & .838 & .999  & .001  & .547 & .853 & .999  & .001 \\
BrainOmni$_1$ & .997 & \textbf{.947} & .996  & .007  & .988 & .948 & .999  & .001  & .964 & .941 & .999  & .002 \\
BrainOmni$_2$ & .998 & .891 & .928  & .077  & .998 & .950 & .966  & .037  & .998 & \textbf{.982} & .994  & .007 \\
BrainOmni$_3$ & .998 & .929 & .927  & .076  & .998 & .970 & .959  & .043  & .998 & .973 & .954  & .049 \\
BrainOmni$_4$ & .998 & .946 & .947  & .055  & .998 & \textbf{.988} & .930  & .072  & .998 & .972 & .909  & .094 \\
LaBraM$_{no}$    & .353 & .755 & .995  & .006  & .369 & .721 & .973  & .038  & .171 & .607 & .986  & .022 \\
LaBraM        & .140 & .694 & \textbf{.267}  & \textbf{1.155} & .093 & .518 & \textbf{-.954} & \textbf{6.755} & .043 & .586 & \textbf{.800}  & \textbf{.294} \\ \bottomrule
\end{tabular}
}
\caption{Codebook utilization and class geometry under different settings. We report codebook coverage (Cov), normalized token entropy (Ent), mean inter-class similarity (Inter), and the inter-/intra-class distance ratio (DR) for BFM, LaBraM with a frozen or updated codebook, and the four RVQ levels of BrainOmni.}
\label{tab:codebook}
\end{table}

\begin{table}[!h]
\setlength{\tabcolsep}{1.8pt}
\resizebox{1.05\columnwidth}{!}{
\begin{tabular}{lcccccccc}
\toprule
           & \multicolumn{2}{c}{AUROC} & \multicolumn{2}{c}{Acc} & \multicolumn{2}{c}{F1} & \multicolumn{2}{c}{F2} \\ \midrule
Dataset      & CB     & origin     & CB        & origin    & CB       & origin      & CB     & origin    \\ \midrule
ADFD  &  .65±.06 &  .77±.10 &  .62±.05 &  .72±.08 &  .68±.03 &  .75±.06 &  .70±.03 &  .75±.06 \\
CHBMIT  &  .77±.06 &  .78±.07 &  .75±.04 &  .75±.06 &  .60±.10 &  .54±.24 &  .55±.07 &  .53±.16 \\
SZ-28    &  .47±.12 &  .54±.07 &  .50±.09 &  .54±.04 &  .56±.12 &  .59±.07 &  .60±.15 &  .59±.13 \\ \bottomrule
\end{tabular}
}
\caption{LaBraM codebook analysis on ADFD, CHBMIT, and SZ-28. We compare standard fine-tuning without codebook (Origin) and codebook-enabled fine-tuning (CB) under identical protocols, reporting AUROC, accuracy, F1, and F2.}
\label{tab:exp5} 
\end{table}

\begin{table}[!h]
\setlength{\tabcolsep}{2pt}
{\small
\begin{tabular}{cccccccc}
\toprule
            \multicolumn{2}{c}{AUROC} & \multicolumn{2}{c}{Acc} & \multicolumn{2}{c}{MF1} & \multicolumn{2}{c}{Kappa} \\ \midrule
 CB     & origin     & CB        & origin    & CB       & origin      & CB     & origin    \\ \midrule
 .86±.02 &  .93±.01 &  .66±.04 &  .76±.04 &  .61±.03 &  .68±.03 &  .54±.05 &  .67±.05 \\ \bottomrule
\end{tabular}
}
\caption{LaBraM codebook analysis on SleepEDF, reporting Accuracy, AUROC (OvR), macro-F1 (MF1), and Cohen’s $\kappa$.}
\label{tab:exp5} 
\end{table}
\FloatBarrier
\section{SSL Paradigms}
\label{app:ssl}
EEG and iEEG are naturally modeled as multivariate time series. We denote a recording as $X \in \mathbb {R} ^ {C \times N} $, where $C$ is the number of channels and $N=f_s\times t$ is the number of samples over duration $t$ at sampling rate $f_s $. 
For electrode-wise representations, let $\epsilon \subseteq \{1,\ldots,C\}$ be a subset of electrodes with $E_e = |\epsilon|$. The corresponding multivariate time series is
$x^{e} \in \mathbb{R}^{E_e \times N}$.
We categorize pretraining pipelines into patch-level and sequence-level formulations.
In patch-level modeling, the raw signal $X$ is partitioned along the temporal dimension into $k$ non-overlapping windows of length $w = \frac{N}{k}$.
Objectives are defined at the patch granularity, and patches are typically treated as exchangeable tokens. 
Sequence-level modeling preserves the temporal structure and treats the input as an ordered sequence. 

SSL derives training signals from the data, enabling pretraining without manual annotations. 
Most SSL methods can be viewed as optimizing an objective constructed from transformations and prediction modules applied to the input signal. We form two sets from each input, a source set $\mathcal{X}_{src}$ and a target set $\mathcal{X}_{tgt}$. An encoder $\mathcal{F}(\cdot)$ maps the input to a $d$-dimensional latent representation, followed by an optional module $\mathcal{G}(\cdot)$ that serves either as a decoder for reconstruction or as a prediction or alignment head. 
Under this formulation, SSL methods can be expressed with a unified objective:

{\fontsize{9}{4}\selectfont
\begin{equation}
\begin{aligned}
\label{total}
\mathcal{L}_{\mathrm{sum}} = \mathcal{L}(
\mathcal{G}(\mathcal{Q}(\mathcal{F}(\mathcal{T}_{d,s}(\mathcal{X}_{src}),\mathcal{M}))),\mathcal{F}(\mathcal{T}_{d,s}(\mathcal{X}_{tgt})),\mathcal{N})
\end{aligned}
\end{equation}}

Here, $\mathcal{T}_{d,s}=\Phi_d \circ a_s$ denotes a deterministic transform composed of domain mapping $\Phi_d$ and augmentation $a_s$. The tokenizer $\mathcal{Q}(\cdot)$ discretizes continuous latent representations. And $\mathcal{M}$ denotes a masking operator, while $\mathcal{N}$ is the negative set when contrastive objectives are employed. 
Based on this formulation, we categorize BFM pretraining strategies into three SSL paradigms.

\subsection{Contrastive-based Method}
Contrastive-based SSL defines learning objectives by comparing representations of paired inputs and contrasting them against a set of negatives. This paradigm is expressed as:

{\fontsize{9}{4}\selectfont
\begin{equation}
\begin{aligned}
\label{contrast}
\mathcal{L}_C=\mathcal{L}(\mathcal{G}(z_1), z_2,\mathcal{N})
\end{aligned}
\end{equation}}

where $z_1=\mathcal{F}_1(\mathcal{T}_{d,s}(\mathcal{X}_{src},\mathcal{M}))$ and $z_2=\mathcal{F}_2(\mathcal{T}_{d,s}(\mathcal{X}_{tgt}))$, $\mathcal{F}_1(\cdot)$ produces a query representation from the causal context while $\mathcal{F}_2(\cdot)$ generates the target embedding from a future segment. The projection head $\mathcal{G}(\cdot)$ maps representations into a latent space. 
The loss contrasts the positive pair $(z_1, z_2)$ with negatives $\mathcal{N}$, promoting alignment of matched pairs and separation from mismatches.
\subsubsection{Augmentation contrast}
Augmentation-based contrastive learning constructs positive pairs by applying transformation operators $a_s$ to the same input signal, with representation learning driven by enforcing invariance across augmented views. Early BFM studies primarily focused on expanding the effective view space induced by $a_s$. SeqCLR \cite{mohsenvand2020contrastive} adapts SimCLR ~\cite{chen2020simple} to multichannel EEG by recombining channels to expand the augmentation-induced view space and enforce consistency across views. SppEEGNet ~\cite{li2022spp} further applies a suite of EEG-specific augmentations and forms positive pairs within sliding windows on a 2D signal representation, which helps reduce cross-dataset sampling-rate mismatch. BIOT ~\cite{yang2023biot} introduces channel-level augmentations tailored to biosignals. Nevertheless, contrastive objectives can suffer from false negatives, especially in sleep staging, where adjacent segments may share labels. SSLAPP ~\cite{lee2022self} addresses this by adversarially generating high-quality positives and performing attention-guided augmentation in latent space, mitigating semantic collisions in pair construction.

As view construction strategies matured, research attention gradually shifted toward the design of the encoder $\mathcal{F}(\cdot)$ to better capture EEG dynamics. Early approaches predominantly relied on CNN-based architectures ~\cite{mohsenvand2020contrastive, li2022spp, kumar2022muleeg}, which primarily model local temporal patterns. Recognizing the intrinsically non-stationary and multi-scale nature of neural signals, later works incorporate temporal–spectral duality into representation learning. TF-C ~\cite{zhang2022self} enforces consistency between time-domain and frequency-domain embeddings of the same signal, explicitly coupling complementary signal views. More recently, transformer-based encoders have been introduced to support cross-dataset and cross-subject learning at scale. BIOT ~\cite{yang2023biot} employs a biosignal transformer to unify heterogeneous modalities, while LEAD ~\cite{wei2025multi} further regularizes representations through subject-level consistency, encouraging invariant embeddings across samples from the same individual.

\subsubsection{Contrastive Predictive Coding}
Contrastive-based SSL defines learning objectives by comparing representations of paired inputs and contrasting them against a set of negatives. 
This paradigm can be expressed as:

{\fontsize{9}{4}\selectfont
\begin{equation}
\begin{aligned}
\label{contrast}
\mathcal{L}_C=\mathcal{L}(\mathcal{G}(z_1), z_2,\mathcal{N})
\end{aligned}
\end{equation}}

where $z_1=\mathcal{F}_1(\mathcal{T}_{d,s}(\mathcal{X}_{src},\mathcal{M}))$ and $z_2=\mathcal{F}_2(\mathcal{T}_{d,s}(\mathcal{X}_{tgt}))$, $\mathcal{F}_1(\cdot)$ produces a query representation from the causal context while $\mathcal{F}_2(\cdot)$ generates the target embedding from a future segment. The projection head $\mathcal{G}(\cdot)$ maps representations into a latent space. 
The loss contrasts the positive pair $(z_1, z_2)$ with negatives $\mathcal{N}$, promoting alignment of matched pairs and separation from mismatches.
Inspired by BERT and wav2vec 2.0~\cite{baevski2020wav2vec}, Bendr~\cite{kostas2021bendr} encodes EEG segments into a unified sequence of learned vector representations. Maeeg~\cite{chien2022maeeg} and GEFM~\cite{wang2024graph} have similar architecture to Bendr with different PT mindset. Although Bendr can extract features well, it ignores spatial information. To address this issue, several models have been proposed. GEFM adds a graph structure to Bendr and proposes a sequence length adjustment mechanism before GNN to make EEG signal lengths consistent for fixed-length node features in GNN. Zhu et al.~~\cite{zhu2023eeg2vec} incorporated spatial information through channel-mixing augmentation, effectively enhancing dataset diversity and improving the performance of the contrastive EEG2Vec framework. Another line of work extends CPC to better exploit multichannel structure and continual data streams. MBrain ~\cite{cai2023mbrain} augments CPC with spatial awareness by aggregating multichannel semantics to predict single-channel local representations, encouraging implicit spatio-temporal dependencies across channels. BrainUICL ~\cite{zhoubrainuicl} further combines CPC with replay-based continual learning, and the contrastive model is incrementally updated via joint training with replayed samples.

\begin{figure*}[!ht]
  \centering
  \includegraphics[width = 0.95\textwidth]{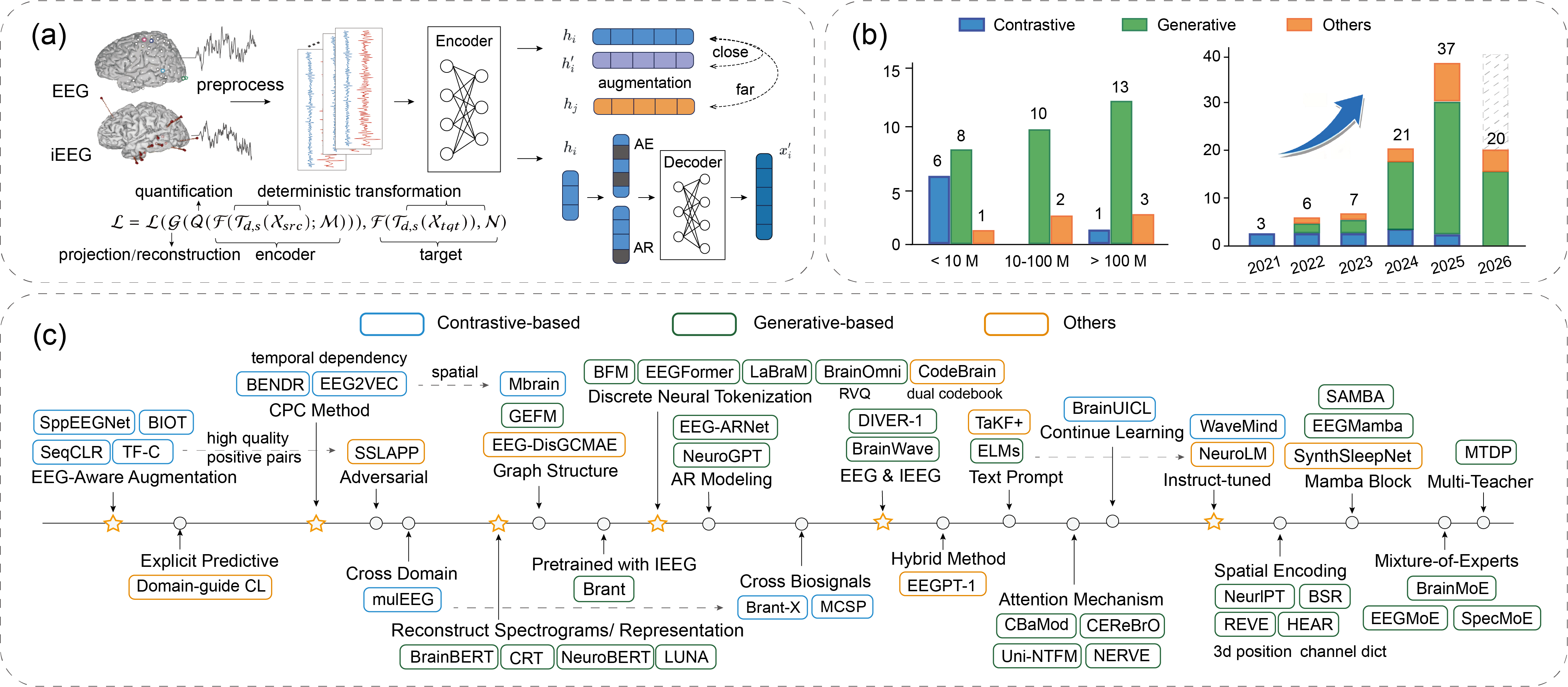}
  \caption{\textbf{Overview of BFMs.} (a) \textmd{A unified pretraining pipeline. EEG/iEEG recordings are preprocessed, encoded into latent representations, and optimized under different SSL paradigms.} (b) \textmd{Model scale statistics. Parameter-size buckets by training paradigm are shown for the reported subset, using each model’s maximum parameter count, alongside yearly model counts under the same grouping.} (c) \textmd{Timeline-style family tree of representative BFMs from 2021 to 2026.6, organized by paradigm and annotated with major methodological shifts.}}
  \label{fig:background}
\end{figure*}

\subsubsection{Cross-modal Contrast}
Inspired by multimodal contrastive learning, BFMs extend cross-modal alignment to EEG, where “modality” arises from the inherent heterogeneity of brain recordings. At the intra-signal level, different views of the same EEG (e.g., raw waveforms vs. time–frequency representations) can serve as distinct modalities. MuLEEG ~\cite{kumar2022muleeg} follows this paradigm by contrasting multi-view raw signals with spectrograms. Beyond view-level heterogeneity, MCSP ~\cite{wei2025multi} mines complementary information across modalities and models latent interactions within each domain, particularly spatial structure via graphs. For truly multi-sensor settings, SleepFM ~\cite{thapa2024sleepfm} jointly embeds three biosignal modalities using both pairwise contrast and a leave-one-out objective, aligning each modality to the average of the others. Brant-X ~\cite{zhang2024brant} further targets EEG–EXG coupling by aligning representations at both patch and sequence levels, capturing correlations at fine- and coarse-grained semantic scales. Moving to heterogeneous cross-modal alignment beyond biosignals, \citep{ferrante2024towards} align neural recordings with visual stimuli, demonstrating that visual content can be decoded from neural data and that images can be mapped into neural representation spaces.

\subsection{Generative-based Methods}
Generative-based SSL formulates pretraining objectives by reconstructing or predicting structured targets from transformed inputs, avoiding explicit pair construction and negative sampling~~\cite{zhang2023self}. 
This paradigm has become increasingly prevalent in recent BFMs. 
Given a transformed input $\mathcal{T}_{d}(x)$, an encoder $\mathcal{F}_{\theta_1}(\cdot)$ produces a continuous latent representation $z_i = \mathcal{F}_{\theta_1}(\mathcal{T}_{d}(x_i))$ where $x_i\in\mathcal{X}_{src}$. 
A learnable codebook $\mathcal{Q}_{\theta_3}(\cdot)$ discretizes $z_i$ into a latent embedding, which is then processed by a decoder $\mathcal{G}_{\theta_2}(\cdot)$ to reconstruct a target view or predict future content.
The objective is defined as:

{\fontsize{9}{4}\selectfont
\begin{equation}
\begin{aligned}
\label{eqn-generative}
\mathcal{L}_G = \mathcal{L}(
\mathcal{G}_{\theta_2}(\mathcal{Q}_{\theta_3}(\mathcal{F}_{\theta_1}(\mathcal{T}_d({x}_{i}),\mathcal{M}))),\mathcal{F}_{\theta_1}(\mathcal{T}_{d}(x_{i})))
\end{aligned}
\end{equation}}

By minimizing the reconstruction or prediction loss, the model learns latent representations that retain information necessary to recover the underlying signal structure.
\subsubsection{Autoregressive}
The success of autoregressive (AR) language models in modeling long-range dependencies via next-token prediction has motivated the adoption in BFMs. Using a decoder-only transformer architecture, GPT-style models ~\cite{radford2019language} have been adapted to EEG pretraining in several BFMs ~\cite{yue2024eegpt, wang2024eegpt, lloyd2024stress}. Early efforts such as Neuro-GPT ~\cite{cui2024neurogpt} segment continuous EEG into fixed-length chunks and treat each chunk as a token, enabling a GPT model $\mathcal{G}_{\theta_2}(\cdot)$ to learn spatio-temporal structure by predicting masked segments. Building on this framework, Stress-GPT ~\cite{lloyd2024stress} fine-tunes the pretrained model for stress-related tasks, demonstrating the transferability of AR-pretrained representations.

Subsequent works refine the tokenization and prediction strategy to better reflect EEG-specific structure. EEGPT-2 ~\cite{yue2024eegpt} adopts an electrode-wise autoregressive formulation, treating each electrode signal $x_i^e$ as a basic token and modeling temporal dependencies through an Electrode Temporal Encoder. EEG2Rep ~\cite{mohammadi2024eeg2rep} shifts target prediction into latent space via context-driven masking and introduces a semantic subsequence preserving mechanism to provide more informative masked inputs. Inspired by large language models, NeuroLM ~\cite{jiang2024neurolm} generalizes autoregressive pretraining to multichannel EEG, explicitly modeling inter-channel dependencies. More recently, LBLM ~\cite{zhou2025pretraining} unifies temporal and spectral autoregression to capture spectro-temporal dynamics, while ECHO ~\cite{liu2025echo} constructs discrete, prompt-like context supports encoding hierarchical signal–task–label relations, improving in-context learning for downstream EEG tasks.

\subsubsection{Autoencoder}
Autoencoder is the most common method in BFMs pretraining. To learn informative representations $\mathcal{M}(x)$, AE-based BFMs rely on carefully designed disturbance mechanisms that force reconstruction from partial observations. Masked Autoencoders (MAE) ~\cite{he2022masked} randomly mask input regions and reconstruct missing content from visible context, encouraging the model to capture global structure. Inspired by both Bendr ~\cite{kostas2021bendr} and MAE, MAEEG ~\cite{chien2022maeeg} and GEFM ~\cite{wang2024graph} adopt masked reconstruction for EEG pretraining. UniEEG ~\cite{jinunieeg} further introduces Masked Signal Modeling (MSM) with electrode-wise masking, while EEGPT-1 ~\cite{wang2024eegpt} embeds local spatio-temporal patches as tokens and proposes a dual SSL objective combining mask-based reconstruction with spatio-temporal alignment.

Beyond masking strategies, modern AE-based BFMs primarily differ in the design of the encoder $\mathcal{F}_{\theta_1}(\cdot)$, which must model long-range temporal dependencies while capturing cross-channel interactions. Early models such as BrainBERT ~\cite{wang2023brainbert} largely encode electrodes independently, limiting spatial coupling. Subsequent works introduce stronger spatial inductive biases: EEG2TEXT ~\cite{liu2024eeg2text} employs multi-view attention to reflect region-wise processing, while CBraMod ~\cite{wang2024cbramod} adopts a criss-cross encoder that alternates spatial–temporal attention with asymmetric conditional positional encoding. To handle heterogeneous montages, LUNA ~\cite{doner2025luna} unifies variable electrode layouts into a fixed latent space via learned queries, and REVE ~\cite{ouahidi2025reve} injects anatomical priors through 4D positional embeddings derived from electrode coordinates and time indices. Beyond attention-based designs, Mamba-style encoders enable linear-time sequence modeling for long recordings. SynthSleepNet ~\cite{lee2025toward} introduces a Mamba-based Temporal Context Module for inter-epoch dependencies, while SAMBA ~\cite{hong2025samba} proposes a Multi-head Differential Mamba to suppress background noise while aggregating contextual information.

However, the inherent randomness, non-stationarity, and nonlinearity of neurophysiological signals make direct amplitude reconstruction in the spatiotemporal domain suboptimal ~\cite{wu2022neuro2vec}. As a result, recent BFMs increasingly perform masking and reconstruction in transformed spaces $\mathcal{T}_d(\cdot)$ or low-dimensional latent variables $z_i$. Neuro2vec ~\cite{wu2022neuro2vec} pioneers denoising in the Fourier domain, while CRT ~\cite{zhang2023self} performs cross-domain dropping–reconstruction to align time- and frequency-domain representations. Building on Fourier-domain MAE, Neuro-BERT ~\cite{wu2024neuro} introduces Fourier Inversion Prediction (FIP) as a pretraining objective. In parallel, spectrogram-based reconstruction has gained traction due to its rich time–frequency semantics, supporting both signal understanding and downstream learning ~\cite{wang2023brainbert, ogg2024self, yuan2024brainwave}. 

Most AE-based BFMs are developed for non-invasive EEG, while intracranial EEG (iEEG) remains comparatively underexplored. To bridge this gap, BrainBERT ~\cite{wang2023brainbert}, Brant ~\cite{zhang2023brant}, and BrainWave ~\cite{yuan2024brainwave} extend AE-based pretraining to SEEG recordings. BrainBERT leverages SEEG data from subjects watching videos, Brant adopts a dual-encoder MAE to jointly capture temporal dependencies and spatial correlations, and BrainWave scales pretraining to over 16,000 subjects, setting new benchmarks for diagnosis tasks. While these approaches substantially improve performance, they often increase model size. To address this, CEReBrO ~\cite{dimofte2025cerebro} proposes a compact encoder that represents brain oscillations through alternating attention, offering a parameter-efficient alternative for AE-based BFMs.

\subsubsection{Codebook}
The codebook mechanism, originally introduced by Vector-Quantized Variational Autoencoders (VQ-VAE) ~\cite{van2017neural}, offers a key advantage over standard VAEs by discretizing continuous representations into a finite set of tokens. Recently, this paradigm has been adapted to EEG and iEEG modeling, where encoder outputs are quantized into discrete codebook entries via a quantization operator $\mathcal{Q}_{\theta_3}(\cdot)$ prior to decoding. Such discretization enables token-based representations of neural signals, bridging continuous biosignals with sequence modeling frameworks. The codebook is optimized using the vector-quantization objective: $\mathcal{L}_{vq}=||sg[\mathcal{Q}_{\theta_3}(z_i)]-z_i||_2^2+\beta||sg[z_i]-\mathcal{Q}_{\theta_3}(z_i)||_2^2$ 
where $sg[\cdot]$ denotes the stop-gradient operator. Compared to conventional masked reconstruction, vector-quantization provides a discrete interface for neural sequence modeling, although its downstream benefits depend on the tokenizer design and adaptation setting.

A central challenge of this paradigm lies in learning an effective quantization operator $\mathcal{Q}_{\theta_3}(\cdot)$. EEGFormer ~\cite{chen2024eegformer} first applies codebook-based discretization to EEG pretraining, generating discrete indices that yield transferable and interpretable representations. LaBraM ~\cite{jiang2024large} further introduces a neural codebook that quantizes patch-level embeddings, with the codebook optimized through reconstruction in the Fourier domain. BioSerenity-E1 ~\cite{bettinardi2025bioserenity} adopts a transformer-based VQ-VAE to tokenize EEG spectra and performs masked token prediction, forcing the model to capture complex spatiotemporal dependencies and achieving state-of-the-art diagnostic performance. Beyond reconstruction-based objectives, NeuroLM ~\cite{jiang2024neurolm} extends this line of work by introducing vector-quantized time–frequency prediction and aligning EEG tokens with textual representations via adversarial training. After discretization, it employs multichannel autoregressive modeling, enabling LLMs to predict EEG tokens in a manner analogous to language modeling.

While codebook learning is often instantiated within autoencoder-style frameworks, it should be viewed as a general representation mechanism rather than a method specific to AE-based BFMs. Discrete neural tokens can serve as a unifying interface across reconstruction, contrastive, and autoregressive paradigms, motivating their independent treatment in this appendix.

\subsection{Other Advanced Methods}
Beyond the SSL paradigms discussed above, BFMs explore explicit predictive objectives, hybrid formulations, and post-SSL instruction tuning.
These methods extend the pretraining framework along dimensions of supervision, objective design, and task alignment.
\subsubsection{Explicit predictive}
In the context of EEG, several predictive tasks have been specifically proposed by updating the form of $\mathcal{G}_{\theta_2}(\cdot)$ to better align with neurophysiological characteristics. Domain-guided CL ~\cite{wagh2021domain} introduces Behavioral State Estimation, which predicts the delta-beta ratio to capture arousal-related spectral patterns. The Stopped Band Prediction pretext task~\cite{jo2023channel} focuses on learning frequency-aware representations by asking the model to identify which frequency band has been suppressed in the input signal to capture frequency information. And the Temporal Trend Identification task~\cite{jo2023channel} aims to extract temporal dynamics by classifying pre-defined trends. Although these tasks are neurophysiologically interpretable, their dependence on specific assumptions about signal behavior has still limited their adoption in BFMs.

\subsubsection{Hybrid-based}
Hybrid SSL combines multiple self-supervised objectives within a unified framework to exploit complementary learning signals ~\cite{weng2025self}. In EEG, Domain-guided contrastive learning ~\cite{wagh2021domain} designs SSL pretext tasks grounded in neurophysiological priors, leveraging similarity of brain activity as well as behavioral-state and age-related differences. EEG-DisGCMAE ~\cite{wei2024pre} further introduces a graph-structured hybrid that couples Graph Contrastive pretraining with Graph MAE. LCM ~\cite{chen2025large} follows a different route, using masked generative reconstruction as the primary objective and adding a momentum-updated target encoder with a contrastive term to stabilize representation learning. More broadly, BFMs increasingly pair contrastive objectives with generative modeling to jointly enforce cross-view consistency and preserve signal structure ~\cite{ wang2024eegpt, chen2025large, li2025comet, lee2025toward}. Other hybrids integrate reconstruction with autoregressive prediction to capture both local structure and long-range temporal dynamics ~\cite{jiang2025elastiq, yang2025thd}, and may further incorporate adversarial objectives for domain-aware alignment ~\cite{jiang2024neurolm, lee2022self}.

\subsubsection{Instruction-tuned}
Recent instruction-tuned BFMs bridge EEG representations with language semantics through lightweight alignment modules while keeping LLM backbones mostly frozen. NeuroLM~\cite{jiang2024neurolm} discretizes EEG into temporal–frequency tokens via VQ encoding and aligns EEG and text embeddings through domain-adversarial learning, followed by multichannel autoregressive pretraining and prompt-based multi-task instruction tuning. TaKF+~\cite{jeontakf} emphasizes parameter-efficient transfer by injecting task information through adapters and cross-attention blocks after masked patch pretraining. UniMind~\cite{lu2025unimind} proposes a Neuro-Language Connector that extracts task-relevant temporal–spatial semantics using learnable queries with task-aware routing, and aligns them to the LLM latent space for instruction-conditioned generation. WaveMind~\cite{zeng2025wavemind} combines contrastive multimodal alignment with CLIP and instruction tuning on a large EEG instruction dataset, enabling conversational EEG understanding across tasks.

\end{document}